\documentclass[10pt,twocolumn,letterpaper]{article}

\usepackage{cvpr}              

\usepackage[accsupp]{axessibility} 

\usepackage[dvipsnames]{xcolor}

\usepackage[hidelinks,breaklinks,colorlinks]{hyperref}
\hypersetup{
    colorlinks=true,          
    linkcolor=WildStrawberry, 
    citecolor=ForestGreen,    
    filecolor=magenta,        
    urlcolor=RoyalBlue        
}

\usepackage{url}            
\usepackage{booktabs}       
\usepackage{amsfonts}       
\usepackage{nicefrac}       
\usepackage{microtype}      
\usepackage{xcolor}         
\usepackage{ulem}
\usepackage{graphicx}
\usepackage{subcaption}
\usepackage{amsmath}
\usepackage{amssymb}
\usepackage{mathtools}
\usepackage{amsthm}
\usepackage{bbm}
\usepackage{multirow}
\usepackage{multicol}
\usepackage{mathrsfs}
\usepackage{float}
\usepackage{lipsum}
\usepackage{tikz}
\usepackage{amsfonts}
\usepackage{algorithm}
\usepackage{algpseudocode}
\usepackage{makecell}
\usepackage{colortbl}
\usepackage{mathabx}
\usepackage{contour}
\usepackage{wrapfig}

\newcommand{\todoc}[2]{{\textcolor{#1}{\textbf{#2}}}}
\newcommand{\todored}[1]{{\todoc{red}{\textbf{[#1]}}}}

\newcommand{\todoblue}[1]{\todoc{blue}{\textbf{[#1]}}}

\newcommand{\sy}[1]{\todoblue{Siyuan: #1}}

\newcommand{\xz}[1]{\todored{xz: #1}}

\usepackage{amsmath,amsfonts,bm}

\def\eqref#1{equation~\ref{#1}}

\def\1{\bm{1}}

\DeclareMathAlphabet{\mathsfit}{\encodingdefault}{\sfdefault}{m}{sl}
\SetMathAlphabet{\mathsfit}{bold}{\encodingdefault}{\sfdefault}{bx}{n}

\definecolor{Gray}{gray}{0.91}
\newcolumntype{g}{>{\columncolor{Gray}}c}
\newcolumntype{G}{>{\columncolor{Gray}}r}

\definecolor{myred}{rgb}{1.0, 0.4, 0.4}   
\definecolor{myblue}{rgb}{0.1, 0.1, 1.0}  

\definecolor{radar_red}{rgb}{0.6, 0.0, 0.0}
\definecolor{radar_blue}{rgb}{0.0, 0.4, 0.6}

\definecolor{ov_sem}{rgb}{0.8, 0.88, 0.74}

\newcommand*\circled[1]{\tikz[baseline=(char.base)]{
            \node[shape=circle,draw,inner sep=0.3pt] (char) {#1};}}

\newcommand*\emptycircle[1][1.1ex]{%
    \begin{tikzpicture}
    \draw[thick] (0,0) circle (#1);
    \end{tikzpicture}
}
\newcommand*\halfcircle[1][1.1ex]{%
    \begin{tikzpicture}
    \draw[fill] (0,0)-- (90:#1) arc (90:270:#1) -- cycle ;
    \draw[thick] (0,0) circle (#1);
    \end{tikzpicture}
}
\newcommand*\fullcircle[1][1.1ex]{%
    \begin{tikzpicture}
    \draw[fill] (0,0)-- (0:#1) arc (0:360:#1) -- cycle ;
    \draw[thick] (0,0) circle (#1);
    \end{tikzpicture}
}

\newcommand{\ours}{\mbox{{\textsc{Co-Spy}}}\@}

\newcommand{\ourtest}{\mbox{{\textsc{Co-SpyBench}}}\@}

\def\paperID{11795} 
\def\confName{CVPR}
\def\confYear{2025}

\title{\ours: Combining Semantic and Pixel Features to \\ Detect Synthetic Images by AI}

\author{
Siyuan Cheng$^{\ast}$, Lingjuan Lyu$^{\dagger\star}$, Zhenting Wang$^{\ddagger}$, Xiangyu Zhang, Vikash Sehwag$^{\dagger}$ \\ \vspace{3pt}
Purdue University, $^{\dagger}$Sony AI, $^{\ddagger}$Rutgers University \\
{\tt\small
\{cheng535, xyzhang\}@purdue.edu}\\
{\tt\small
$^{\dagger}$\{Vikash.Sehwag, Lingjuan.Lv\}@sony.com} \\
{\tt\small
$^{\ddagger}$zhenting.wang@rutgers.edu}
}

\begin{document}

\maketitle

\def\thefootnote{$\ast$}\footnotetext{Work done during Siyuan Cheng’s internship at Sony AI.}
\def\thefootnote{$\star$}\footnotetext{Corresponding Author}

\begin{abstract}
    With the rapid advancement of generative AI, it is now possible to synthesize high-quality images in a few seconds. Despite the power of these technologies, they raise significant concerns regarding misuse.
    Current efforts to distinguish between real and AI-generated images may lack generalization, being effective for only certain types of generative models and susceptible to post-processing techniques like JPEG compression.
    To overcome these limitations, we propose a novel framework, \ours{}, that first enhances existing semantic features (e.g., the number of fingers in a hand) and artifact features (e.g., pixel value differences), and then adaptively integrates them to achieve more general and robust synthetic image detection. Additionally, we create \ourtest{}, a comprehensive dataset comprising 5 real image datasets and 22 state-of-the-art generative models, including the latest models like FLUX. We also collect 50k synthetic images in the wild from the Internet to enable evaluation in a more practical setting.
    Our extensive evaluations demonstrate that our detector outperforms existing methods under identical training conditions, achieving an average accuracy improvement of approximately 11\% to 34\%.
    The code is available at \href{https://github.com/Megum1/Co-Spy}{https://github.com/Megum1/Co-Spy}.
\end{abstract}

\section{Introduction} 
\label{sec:intro}
Despite the tremendous benefits brought by generative AI, the misuse of
generation technologies is a growing concern. They could be employed to create fake news, potentially causing false alarms and spreading widespread misinformation online. For example, in the vision domain, it is disturbingly straightforward to generate realistic images that fabricate sensational stories, such as falsely depicting public figures in controversial situations.
These issues become increasingly prominent as the growth of image generation technologies. There is hence an urgent need to develop effective detection methods for AI-generated images that can mitigate misuse and enhance the trustworthiness of these technologies.

Existing detection methods of AI-generated images, or {\em synthetic images} for brevity, 
can be largely classified into two categories: (1) detectors based on {\em semantic features} and (2) detectors based on {\em texture-level artifacts}.
Semantic detectors~\cite{cnndet,univfd,fusing,drct} distinguish synthetic images primarily based on semantic features, 
such as human hand outlines. Typically, these methods collect a large amount of real and synthetic data across a wide range of objects and then train a deep neural network to identify synthetic images. 
In contrast, artifact detectors~\cite{npr,lnp,freqfd,freqnet} focus on identifying texture-level features
arising from the up-sampling processes commonly used in image generation models. Intuitively, these 
features are distinct pixel-level signatures, such as similar values between adjacent pixels.

While existing techniques have shown effectiveness on specific datasets, these datasets often rely on images generated from a limited range of models from several years ago, such as ProGAN~\cite{progan}. Since then, generated models have seen significant advancements in resolution and realism. Moreover, many of these images are in raw formats like PNG, without undergoing lossy compression or transformation, e.g., JPEG compression.
To more accurately assess the effectiveness of existing techniques on the latest models, diverse transformations, and a wider range of generated content, we created a new dataset \ourtest{},
which comprises over one million images generated by 22 of the latest generative models.
Compared to widely-used datasets such as CNNDet~\cite{cnndet}, DiffusionForensics~\cite{dire}, and the state-of-the-art DRCT-2M~\cite{drct}, \ourtest{} includes 10 additional cutting-edge models, e.g., FLUX~\cite{flux}. Furthermore, our dataset is significantly more diverse. For instance, it incorporates caption descriptions from five different datasets, whereas DRCT only utilizes MSCOCO~\cite{mscoco}.
Our image generation process utilizes variable diffusion steps between 10 and 50, along with various guidance scales, whereas existing datasets typically use fixed, default settings (e.g., 50 diffusion steps in DRCT-2M~\cite{drct}).
Additionally, we apply JPEG compression rates randomly selected between 75 and 95, reflecting typical real-world scenarios. This comprehensive approach ensures that \ourtest{} effectively simulates real-world detection challenges.
Moreover, we have collected 50,000 in-the-wild fake images from five online sources, e.g., Lexica~\cite{lexica} and Midjourney~\cite{midjourney}, to further enhance the dataset's applicability for real-world detection tasks. We present the limitations of existing datasets in \autoref{app:limit_dataset} and provide details of our dataset in \autoref{app:detail_data}.

\begin{table}[t]
    \centering
    \scriptsize
    \tabcolsep=2.5pt
    \caption{\textbf{Generalization Comparison of Existing and Our Enhanced Detectors.} \emptycircle indicates minimal capacity, \halfcircle represents moderate performance, and \fullcircle denotes well support.}
    \label{tab:intro_rate}
    \begin{tabular}{lcgcgcg}
    \toprule
    \multirow{2}{*}{\textbf{Generalization}} & \multicolumn{2}{c}{\textbf{Artifact}~\cite{freqfd,npr}} & \multicolumn{2}{c}{\textbf{Semantic}~\cite{cnndet,univfd}} & \multicolumn{2}{c}{\textbf{Fusion}} \\
    \cmidrule(lr){2-3} \cmidrule(lr){4-5} \cmidrule(lr){6-7}
    ~ & Existing & Enhanced & Existing & Enhanced & Simple & Enhanced \\
    \midrule
    Diverse Models & \fullcircle & \fullcircle & \emptycircle & \halfcircle & \halfcircle & \fullcircle \\
    Lossy Formats & \emptycircle & \halfcircle & \fullcircle & \fullcircle & \halfcircle & \fullcircle \\
    Unseen Objects & \fullcircle & \fullcircle & \emptycircle & \halfcircle & \halfcircle & \fullcircle \\
    \bottomrule
    \end{tabular}
\end{table}

Our evaluation on this dataset reveals a substantial degradation in the performance of existing techniques, with the two kinds of techniques showing varying levels of effectiveness across scenarios.
We summarize the strengths and limitations of these two kinds of detectors in \autoref{tab:intro_rate}.
We consider three practical generalization aspects: (1) supporting detection of images produced by \textit{Diverse Models}; (2) supporting \textit{Lossy Formats}; and (3) generalizing to \textit{Unseen Objects} beyond the training data.
Artifact detectors (Column 2) tend to perform well across diverse models and unseen objects but are vulnerable to lossy formats. This is because they primarily focus on identifying specific pixel values and do not rely on the content of image. However, raw pixel values are easily affected by image compression.
In contrast, semantic detectors are robust against lossy formats (Column 4) but perform poorly on new models and unseen objects. This is because semantic features, e.g., synthetic styles and objects, are specific to the training data.

A straightforward idea is to combine the two approaches to achieve better generalization.
However, our evaluation in \autoref{app:limit_simp_comb}
indicates that the success is limited (as conceptually illustrated in Column 6).
The root cause is that these two types of techniques are effective in distinct scenarios, as shown in Columns 2 and 4. Consequently, the direct integration is merely an additive combination rather than a synergistic one, meaning that neither can elevate the other’s performance in specific areas (e.g., ``Unseen Objects''). Our main approach, therefore, is to first improve each technique to maximize its effectiveness across all three aspects and then fuse the two to achieve optimal performance. For instance, in the ``Diverse Models'' aspect (row 2), current semantic detectors lack support in this area, so the integration only partially inherits what the artifact detectors already provide. However, with enhancements, both artifact and semantic detectors can independently support diverse models, enabling a synergy where each strengthens the other (as demonstrated in \autoref{app:eval_ablation}).

In particular, we enhance semantic detection by adopting a more comprehensive backbone feature extractor using the latest OpenCLIP~\cite{openclip} model. Trained on millions of images, the CLIP model provides extensive coverage of objects and concepts, offering rich semantic representations. Additionally, to mitigate overfitting, we introduce feature-space interpolation as a data augmentation technique, efficiently generating new and challenging samples.
Together, these two enhancements improve the detection of unseen generative models and objects (see Column 5 in \autoref{tab:intro_rate}).
Second, we develop a robust artifact extractor leveraging a pre-trained Variational Autoencoder (VAE)~\cite{vae}. The VAE reconstructs each input image, and by computing the difference between the original and reconstructed images, it identifies higher-level artifacts that extend beyond simple pixel-level discrepancies. These enhanced artifact features are able to support lossy formats (see Column 3 in \autoref{tab:intro_rate}).
Moreover, we introduce a unified detection framework that seamlessly integrates our enhanced semantic and artifact features. Unlike simple ensembling methods, our adaptive fusion technique dynamically balances the importance of these two feature types based on the specific characteristics of each input image. This dynamic balancing ensures robust detection performance across a broader spectrum of scenarios, effectively leveraging the strengths of both 
approaches and supporting generalization across all three scenarios (last column in \autoref{tab:intro_rate}).
As demonstrated in \autoref{sec:eval}, our approach outperforms the state-of-the-art semantic detector DRCT~\cite{drct} by 11\% in accuracy and the latest artifact detector NPR~\cite{npr} by 21\%, trained on the same datasets and evaluated on synthetic images generated by 22 models across 5 datasets and various compression conditions.

Our main contributions are summarized in the following: 
\begin{enumerate}
    \item 
    We present \ourtest{}, a novel, high-quality, and diverse benchmark for synthetic image detection. It comprises over one million images, including real images sourced from five established databases and synthetic images generated using the corresponding real image captions under various configurations. 
    The synthetic images are produced by 22 state-of-the-art text-to-image diffusion models, including latest models like the recently released FLUX~\cite{flux}.
    To enhance 
    diversity, \ourtest{} includes synthetic images generated with varied captions, resolutions, and configurations, such as different diffusion steps and guidance scales. This benchmark surpasses existing datasets in both comprehensiveness and diversity, providing a robust foundation for evaluating future detection methods. Additionally, we collect synthetic images from 5 prominent websites to simulate real-world, in-the-wild testing scenarios.
    \item We offer a comprehensive and insightful analysis of existing detectors. We outline their strengths and weaknesses, and explaining the reasons behind each finding.
    \item We propose a novel synthetic image detection framework \ours{} (``{\it
    \underline{\sc Co}mbining \underline{\sc S}emantic and \underline{\sc P}ixel Features to Detect S\underline{\sc y}nthetic Images by AI
    }'')\footnote{Inspired by the iSpy games that search for hidden objects.} that enhances the existing semantic and artifact extraction methods, and strategically integrates the strengths of both features. This comprehensive framework substantially improves detection performance, outperforming state-of-the-art methods.
    \item We conduct a thorough evaluation to demonstrate the superiority of our framework over existing state-of-the-art methods. As detailed in \autoref{sec:eval}, our approach outperforms the state-of-the-art semantic detector DRCT~\cite{drct} by 11\% in accuracy and the latest artifact detector NPR~\cite{npr} by 21\%, trained on the same datasets and evaluated on synthetic images generated by 22 models across 5 datasets and various compression conditions.
\end{enumerate}




\section{Rethinking the Generalization of Existing Detectors} \label{sec:motivation}


In this section, we present a thorough analysis of current synthetic image detectors, offering an in-depth understanding of their strength and weakness.
As mentioned in \autoref{sec:intro}, existing detectors can be broadly categorized into semantic and artifact detectors. 
We assess two latest semantic detectors, Fusing~\cite{fusing} and UnivFD~\cite{univfd}, alongside with two artifact detectors, LNP~\cite{lnp} and NPR~\cite{npr}, using DRCT-2M/SD-v1.4~\cite{drct} for training and \ourtest{} for evaluation. We treat JPEG compression as a common transformation, applying it with a random quality factor between 75 and 95, representing typical compression levels~\cite{jpeg+quality}. We focus on the three aspects in \autoref{tab:intro_rate}.



\smallskip \noindent
\underline{\textbf{Artifacts Detectors Do Not Support Lossy Formats.}}

\begin{figure}[t]
    \centering
    \begin{minipage}[t]{0.235\textwidth}
        \centering
        \includegraphics[width=1\textwidth]{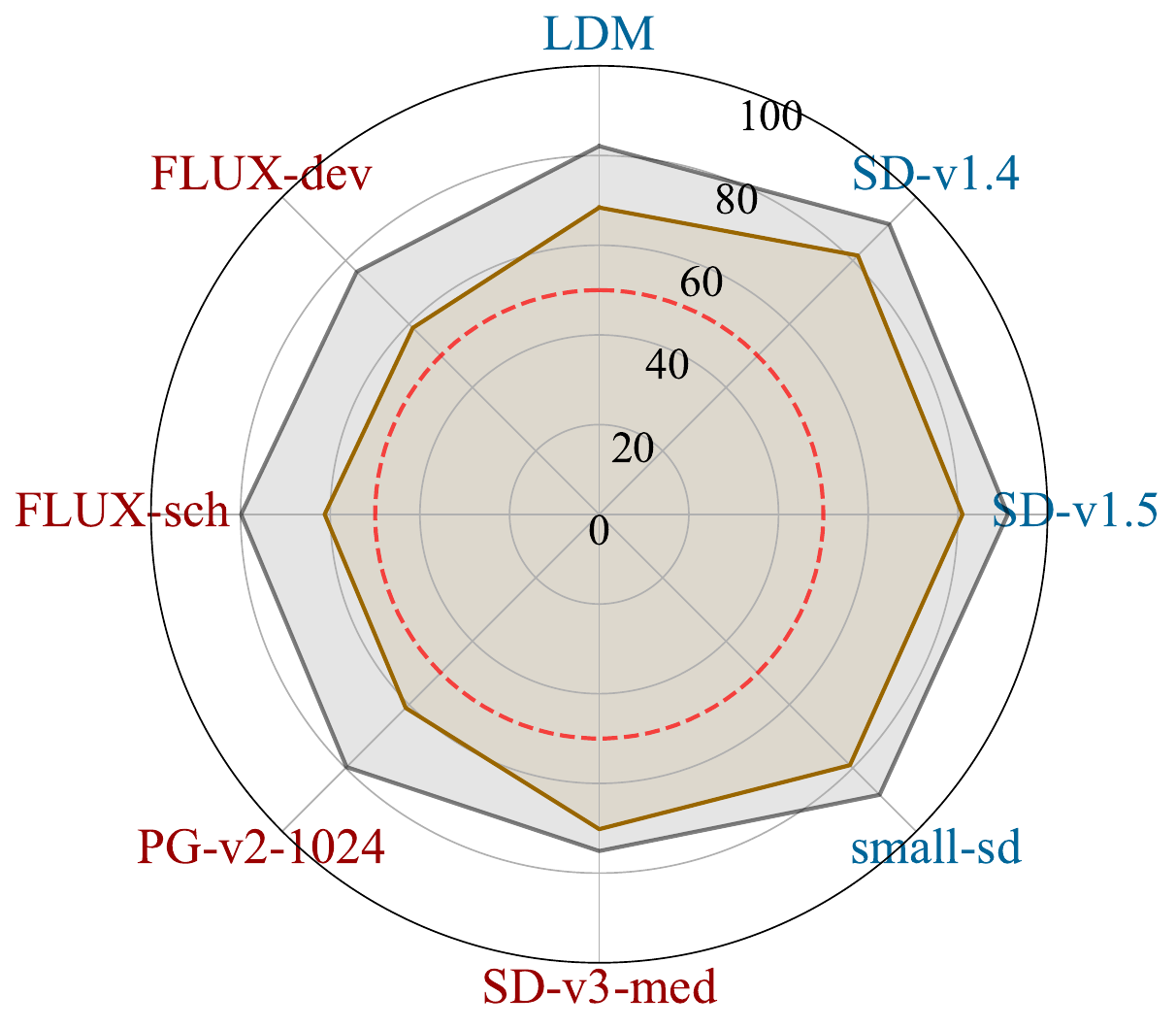}
        \subcaption{LNP~\cite{lnp}
        }
    \end{minipage}
    \hfill
    \begin{minipage}[t]{0.235\textwidth}
        \centering
        \includegraphics[width=1\textwidth]{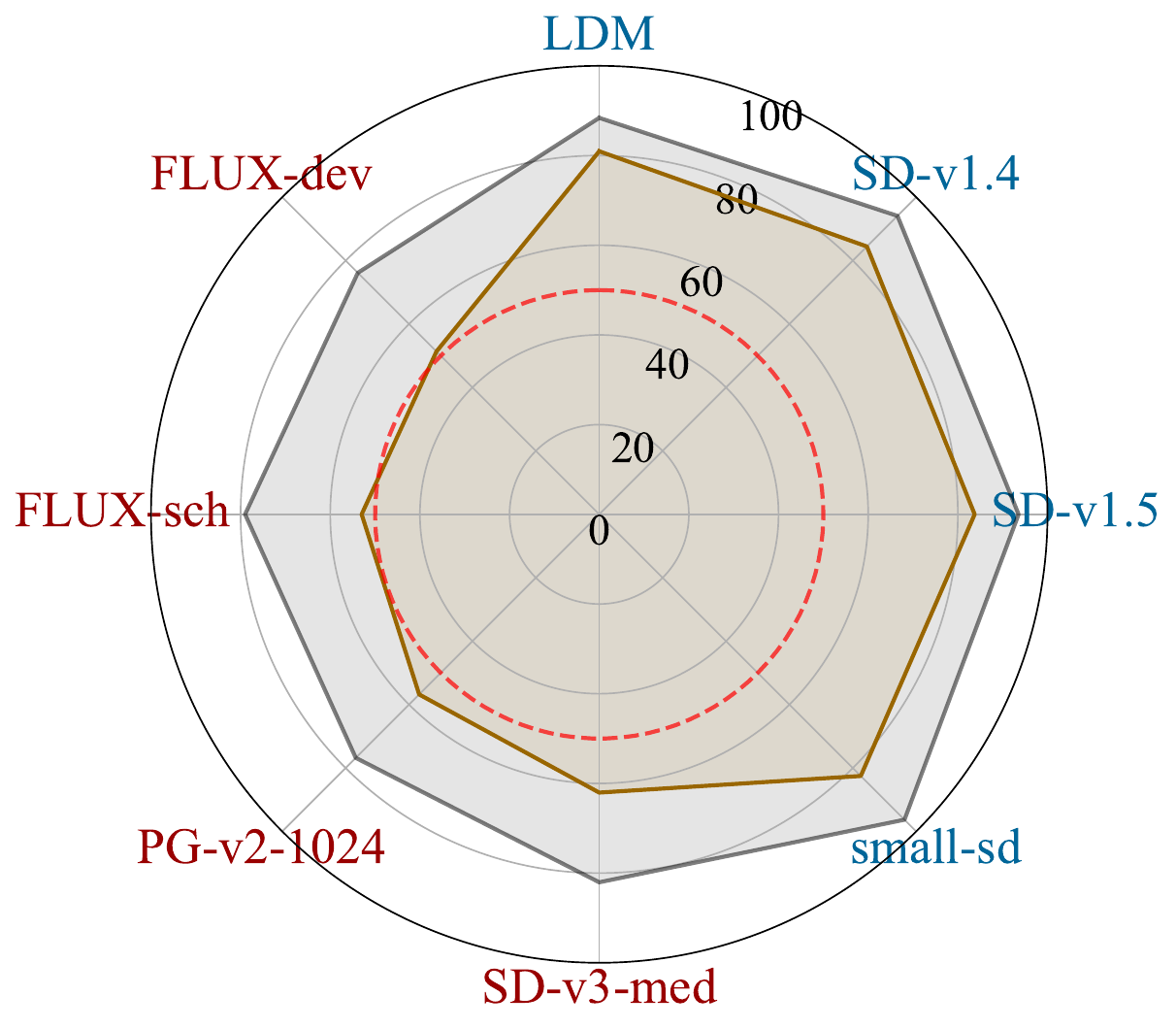}
        \subcaption{NPR~\cite{npr}}
    \end{minipage}
    \caption{\textbf{Artifact Detectors.} Test results \textcolor{gray}{w/o} and \textcolor{brown}{w/} JPEG compression are shown in \textcolor{gray}{gray} and \textcolor{brown}{brown} plots, respectively.
    \textcolor{radar_blue}{Blue} models correspond to older models that generate images closely resembling synthetic training samples. \textcolor{radar_red}{Red} models represent recent models, producing more diverse and distinct images.}
    \label{fig:moti_artifact}
\end{figure}

\noindent
We observe that artifact detectors demonstrate strong generalization to unseen models. However, their performance significantly degrades even with minor JPEG compression. \autoref{fig:moti_artifact} illustrates the performance of the two artifact detectors. In this figure, results for old and similar models as used in the training set, e.g., SD-v1.5~\cite{stablediffusion}, are highlighted in blue while recent and distinct models, e.g., FLUX~\cite{flux}, are shown in red. The gray plots depict test performance without any transformations, whereas the brown plots display performance under the JPEG compression. 
Both detectors achieve high generalization over all models, with an average accuracy of 85\%. However, even slight JPEG compression leads to a substantial accuracy degradation of approximately 17\%. This occurs because artifacts are consistently present across nearly all generative models, primarily due to the necessary up-sampling operations in these models~\cite{freqfd,npr}. However, these artifacts are highly sensitive to input transformations,
which distort or eliminate the subtle cues necessary for detection, thereby impairing performance. More illustrations can be found in \autoref{app:illu_jpeg}.

\begin{figure}[t]
    \begin{minipage}[t]{0.235\textwidth}
        \centering
        \includegraphics[width=1\textwidth]{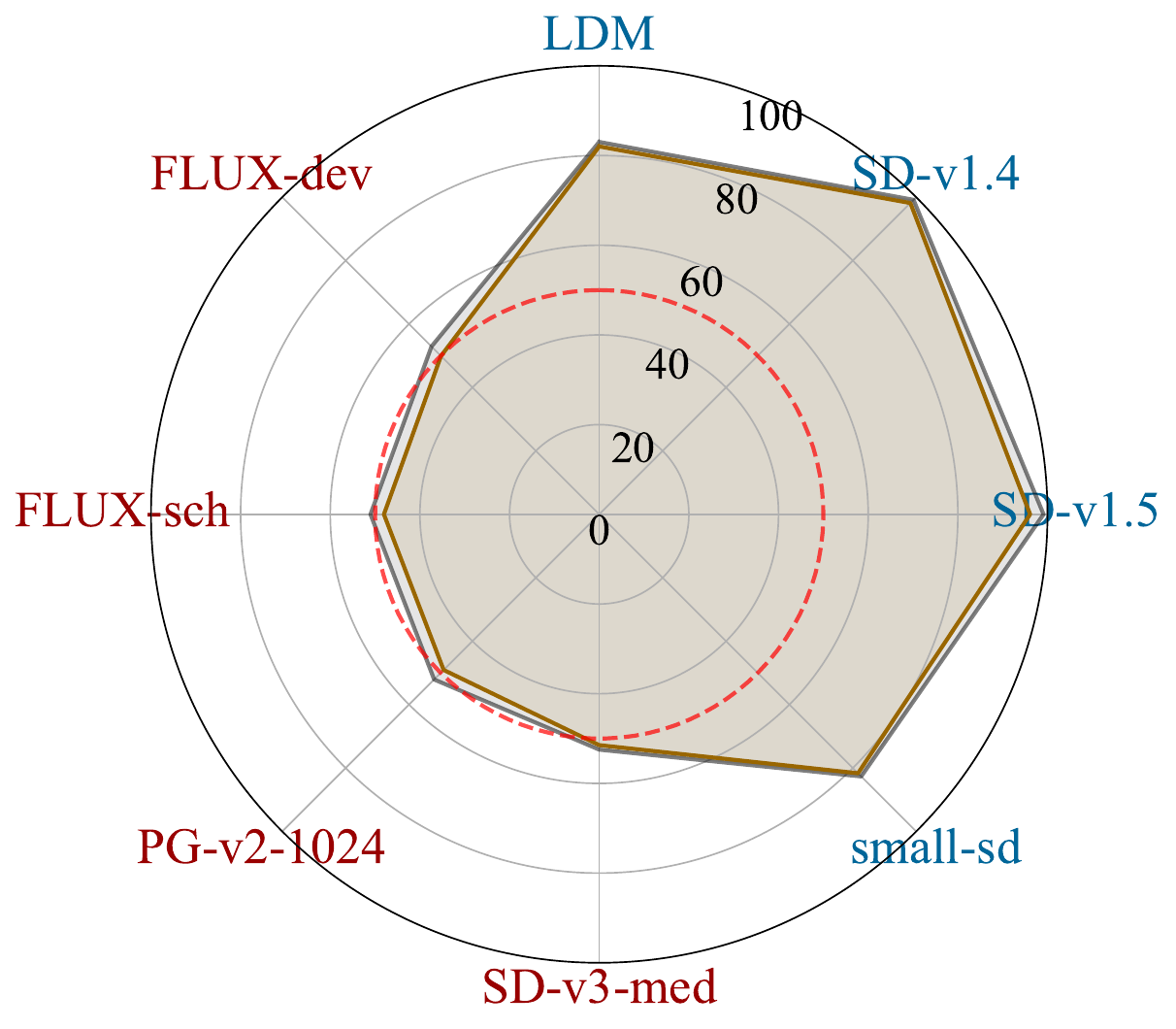}
        \subcaption{Fusing~\cite{fusing}}
    \end{minipage}
    \hfill
    \begin{minipage}[t]{0.235\textwidth}
        \centering
        \includegraphics[width=1\textwidth]{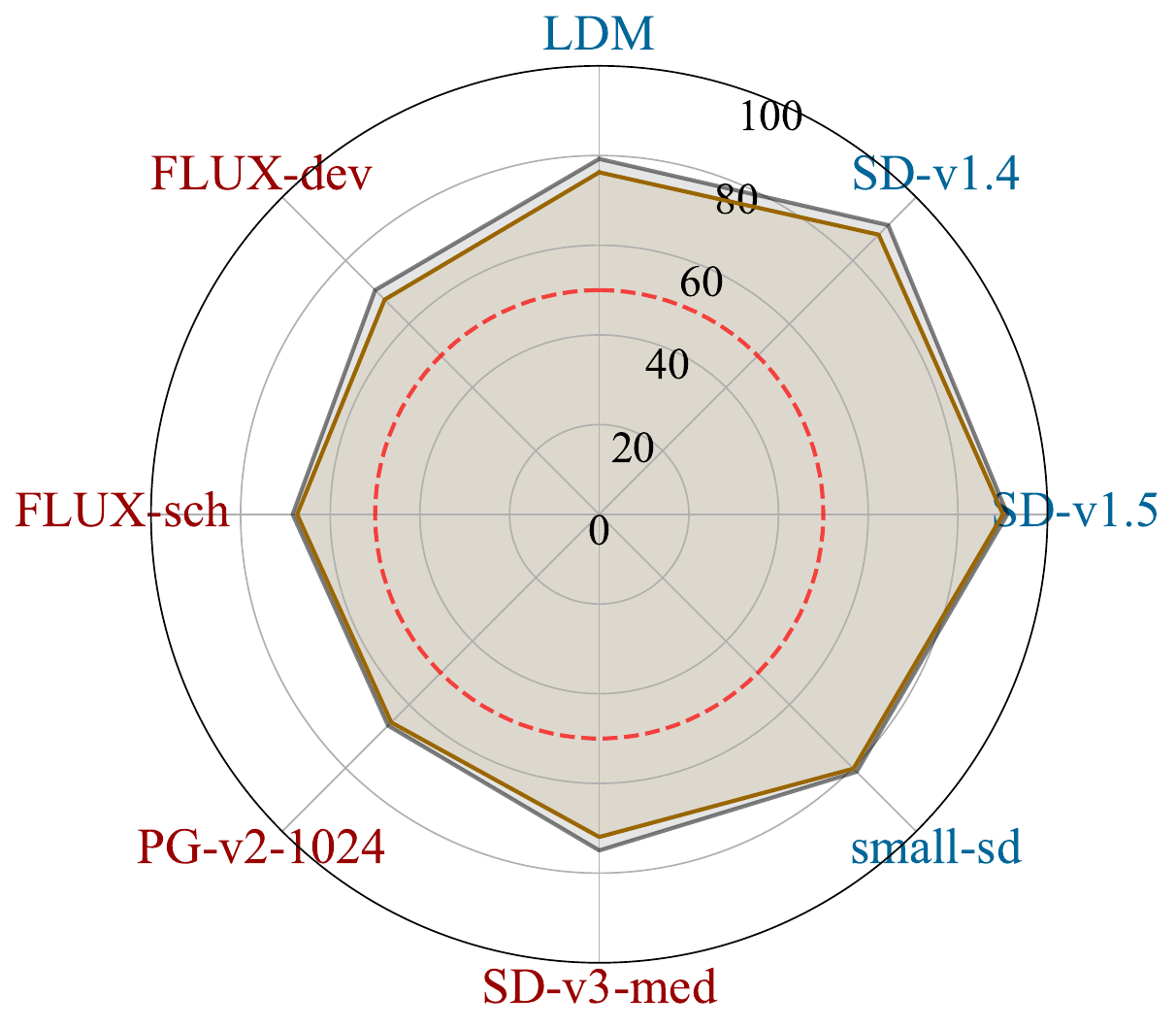}
        \subcaption{DRCT~\cite{drct}}
    \end{minipage}
    \caption{\textbf{Semantic Detectors.} Test results \textcolor{gray}{w/o} and \textcolor{brown}{w/} JPEG compression are shown in \textcolor{gray}{gray} and \textcolor{brown}{brown} plots, respectively.
    \textcolor{radar_blue}{Blue} models correspond to older models that generate images closely resembling synthetic training samples. \textcolor{radar_red}{Red} models represent recent models, producing more diverse and distinct images. }
    \label{fig:moti_input_semantic}
\end{figure}

\smallskip \noindent
\underline{\textbf{Semantic Detectors Do Not Generalize to Unseen Models}} \underline{\bf  and Unseen Contents.}
Semantic detectors, in contrast, perform well on images generated by 
old, similar models, even in the presence of lossy transformation, but exhibit limited ability to generalize to latest models and unseen contents.
\autoref{fig:moti_input_semantic} illustrates the performance of the semantic detectors, which achieve high performance on old models but only around 50\% in latest models which have network structure changes, configuration variations, and use new captions to generate images. Additionally, the accuracy degradation caused by JPEG compression is minor, with a reduction of approximately 2\%.
This is because semantic detectors focus on the meaningful content of images rather than raw pixel values. However, their ability to generalize 
is limited because the features learned from synthetic images from one source differ significantly from those produced by latest diffusion models.

\section{Design} \label{sec:design}
As noted in \autoref{sec:intro}, a direct integration of artifact and semantic detectors results in an additive effect without achieving synergy. Our approach hence first focuses on aligning the two kinds of methods by maximizing their generalization across the three key aspects: models, transformations, and content. We then fuse these enhanced detectors into a unified, cohesive solution. The following sections outline our generalization of each detector type and the process of combining them into an integrated solution.

\begin{figure}[t]
    \centering
    \begin{minipage}[t]{0.47\textwidth}
        \centering
        \includegraphics[width=0.9\textwidth]{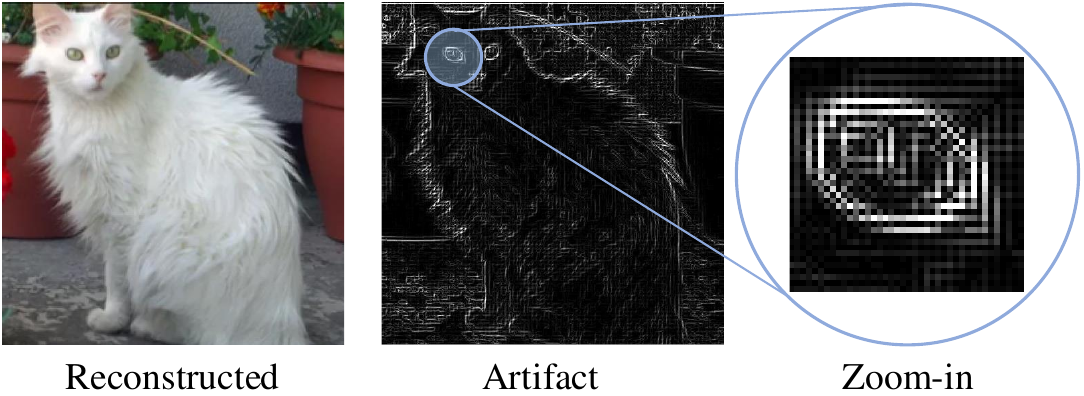}
        \subcaption{Upsampling Artifacts~\cite{npr}}
    \end{minipage}
    \hfill
    \begin{minipage}[t]{0.47\textwidth}
        \centering
        \includegraphics[width=0.9\textwidth]{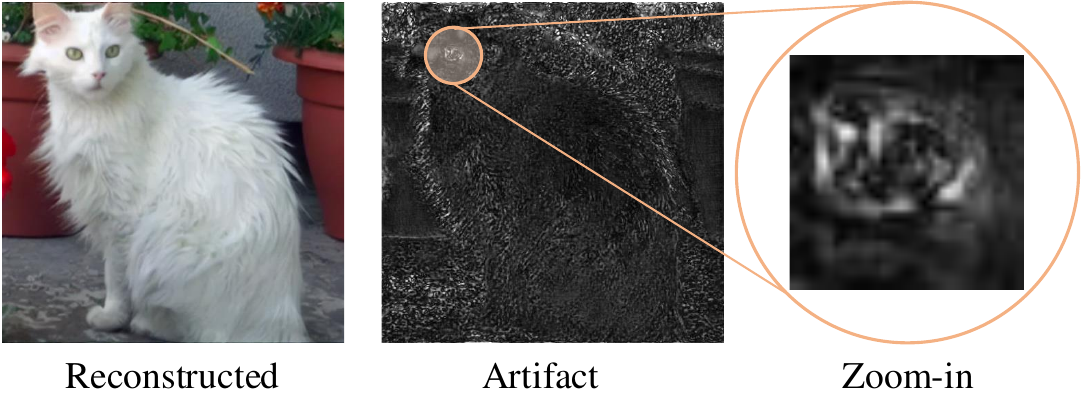}
        \subcaption{VAE Artifacts}
    \end{minipage}
    \caption{\textbf{Enhanced Artifact Extraction.} Sub-figure (a) presents existing artifacts approximated using upsampling, while sub-figure (b) illustrates the artifacts extracted using VAE. Three columns denote reconstructed images, extracted artifact by differencing the original and reconstructed images, and zoom-in areas.}
    \label{fig:design_vae}
\end{figure}


\begin{figure}[t]
    \centering
    \includegraphics[width=0.48\textwidth]{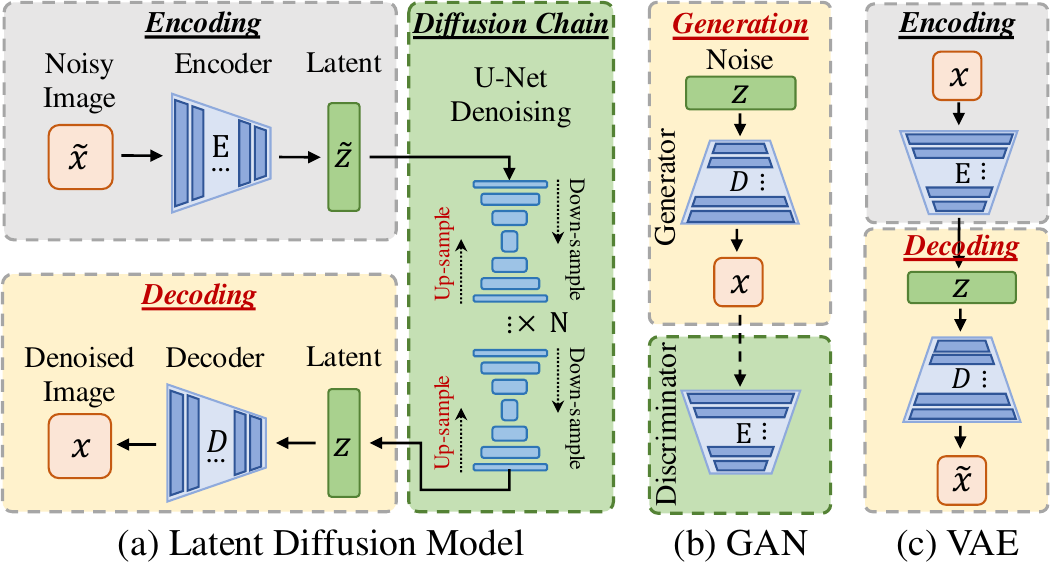}
    \caption{\textbf{
    Architecture Comparison Between Generative Models.}
    Sub-figure (a) presents a typical diffusion model pipeline~\cite{stablediffusion}, sub-figure (b) illustrates a typical GAN pipeline~\cite{gan}, and sub-figure (c) shows our VAE reconstruction. Observe that they all share the similar decoding/generation stage immediately before the output, highlighted in the yellow areas.}
    \label{fig:arch_vae}
\end{figure}

\begin{figure}[t]
    \centering
    \includegraphics[width=0.47\textwidth]{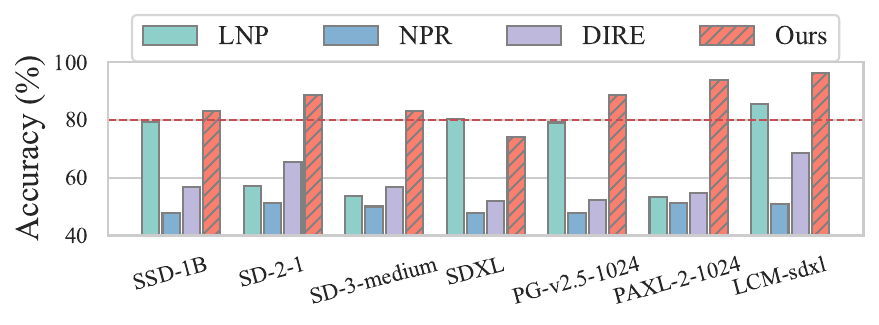}
    \caption{\textbf{Comparison with Latest Artifact Detectors.} Our enhanced detector outperforms the baselines across diverse models.}
    \label{fig:design_art_comp}
\end{figure}

\subsection{Generalizing Artifact Detectors using VAE-Based Artifact Extraction} \label{sec:design_artifact}
As discussed in \autoref{sec:motivation}, texture-level artifacts can generalize to unseen models and objects but do not hold up under lossy transformations like JPEG compression because
these artifacts are low-level and vulnerable to compression.
For example, a popular method~\cite{npr} 
leverages the observation that generative models have extensive down-sampling and up-sampling operations. \autoref{fig:arch_vae}(a) 
shows a typical diffusion pipeline, in which each diffusion step involves down-sampling and up-sampling.
As such, when a synthetic image is passed to a pair of down-sampling and up-sampling operations, the resulted image has less difference from the original image, when compared to a real image that has never undergone such operations. 
Therefore, artifacts can be extracted from such differences and used in classification.
However, these artifacts are very low-level, e.g., in the form of grainy texture, 
as illustrated in \autoref{fig:design_vae}(a).

We observe that besides down-sampling and up-sampling, most generative models also share another common/similar operation -- decoding.  
\autoref{fig:arch_vae}(a-b) show two popular generative models, i.e., the Latent Diffusion Model (LDM)~\cite{stablediffusion} and Generative Adversarial Network (GAN)~\cite{gan}.
LDM encodes a noisy input into a latent space, processes it through a number of diffusion steps, and finally decodes it back to the pixel space. GAN takes a simpler approach, directly mapping a noisy vector to an output image. It also employs a discriminator to guide the generation process. Both have a decoder component $D$. 
Similar to down-sampling and up-sampling, decoding also introduces its own artifacts.
As such, we employ a pre-trained Variational Autoencoder (VAE)~\cite{vae} (i.e., a pair of encoder and decoder)  in \autoref{fig:arch_vae}(c) to capture the 
artifacts, by feeding an image to the VAE and differencing the resulted image with the original one. A downstream classifier is hence trained to distinguish the differences from synthetic images and the differences from real images. 
Compared to the artifacts introduced by down-sampling and up-sampling, the decoding artifacts are at a higher level and hence more robust in the presence of transformations. \autoref{fig:design_vae}(b) 
shows an example of such artifacts. Observe that they are smoother and encode more information of the object, in comparison to the artifacts from the down-sampling and up-sampling operations in \autoref{fig:design_vae}(a).

We formally define our method for extracting these enhanced artifacts in Equation~\ref{equ:vae}.
Let $x$ denote the input, $E$ the VAE encoder, and $D$ the VAE decoder:
\begin{equation} \label{equ:vae}
    \underbrace{\mu, \sigma = E(x)}_{\text{VAE Encoding}} \quad\quad\quad \underbrace{x^{\prime} = E(\mu)}_{\text{VAE Decoding}} \quad\quad\quad \underbrace{\Delta = |x^{\prime} - x|}_{\text{Artifact Extraction}}
\end{equation}
First, the input is encoded into the latent space, resulting in the mean $\mu$ and variance $\sigma$. We then reconstruct the input using only the mean $\mu$, without introducing noise, to generate $x^{\prime}$. The artifact is extracted as the absolute difference between $x^{\prime}$ and $x$. These extracted artifacts are then classified as real or synthetic using a ResNet-50~\cite{resnet}.
\autoref{fig:design_art_comp} compares the performance of our artifact detector, trained on the DRCT/SD-v1.5~\cite{drct} dataset, with LNP~\cite{lnp}, NPR~\cite{npr} (the latest artifact detectors), and DIRE~\cite{dire} (a state-of-the-art detector based on reconstruction artifacts). We applied random JPEG compression (quality 75-95) for training augmentation and evaluation. Our detector shows an average accuracy improvement of 17-37\% over LNP and NPR, and 28\% over DIRE.

\subsection{Generalizing Semantic Detectors} \label{sec:design_semantic}
As discussed earlier, semantic detectors 
do not generalize well for unseen models and objects. This issue is primarily due to overfitting on the training data (more details in \autoref{app:overfit}).
Inspired by UnivFD~\cite{univfd}, we mitigate the problem by  using a recent 
CLIP model~\cite{clip} as the feature extractor.
The idea is to utilize CLIP’s extensive pre-trained knowledge on billions of images. In particular, we use CLIP's vision encoder to generate embeddings for both real and fake images and then use the embeddings to train a downstream classifier. 

\begin{figure*}[t]
    \begin{minipage}[t]{.73\linewidth}
        \centering
        \includegraphics[width=1\linewidth]{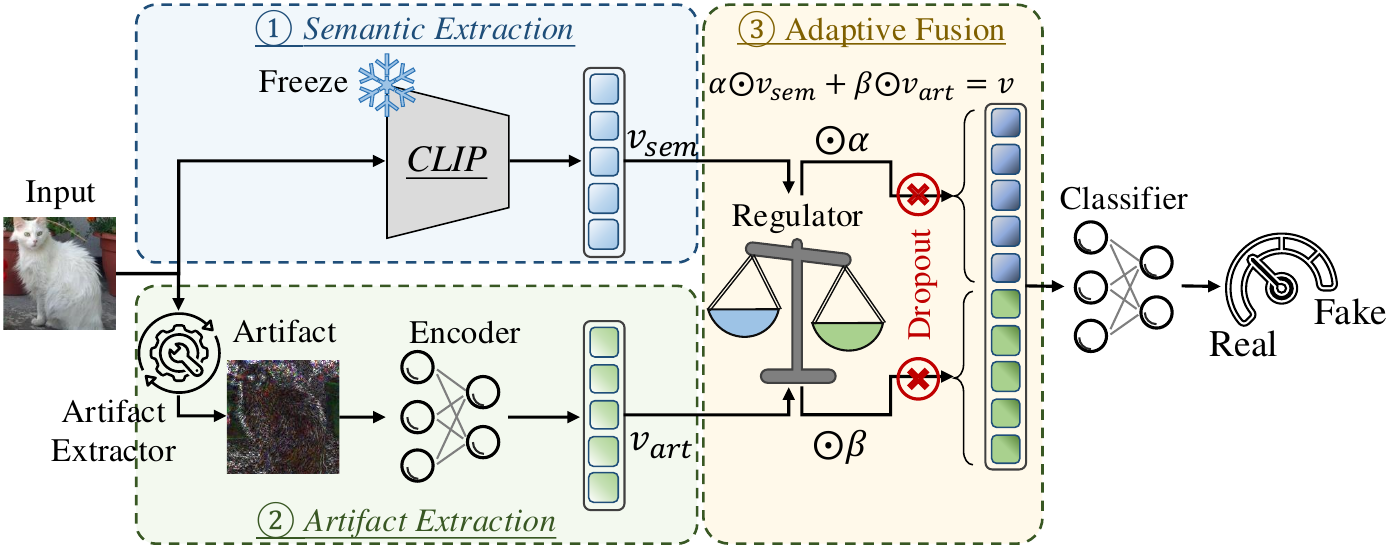}
        \caption{\textbf{Overview of \ours}. The approach consists of three steps: (1) Semantic Extraction, which extracts semantic features using the frozen CLIP model; (2) Artifact Extraction, which derives artifact features through VAE reconstruction and a learnable encoder; and (3) Adaptive Fusion, which dynamically integrates these two types of features. Finally, the fused feature is passed through a binary classifier to distinguish real and synthetic images.}
        \label{fig:overview}
    \end{minipage}
    \hfill
    \begin{minipage}[t]{.25\linewidth}
        \centering
        \includegraphics[width=1\linewidth]{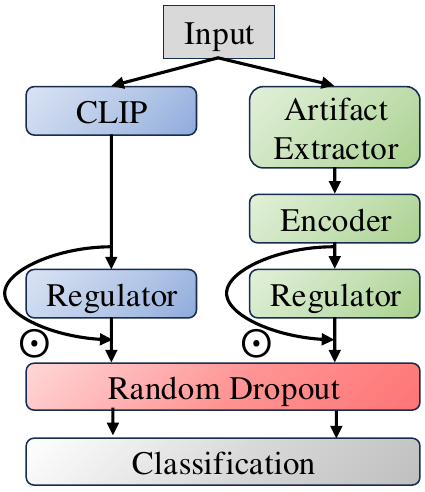}
        \caption{\textbf{Network Architecture of \ours.} Note that in implementation, there are two regulators designed for semantic and artifact features, respectively.}
        \label{fig:arch}
    \end{minipage}
\end{figure*}

In addition, we observe that training with hard-label samples (i.e., samples simply labeled as synthetic and real) does not achieve the optimal result.
We propose to generate additional soft-label samples by feature interpolation~\cite{feat_inter_1,feat_inter_2}, in order to (1) enlarge the training dataset and (2) reduce possible overfitting to hard labels.
In particular, we 
interpolate between the embeddings of real and synthetic images
and assign a continuous synthetic score based on the interpolation coefficient. Let $x_R$ represent real images with a synthetic score of 0, and $x_S$ represent synthetic images with a synthetic score of 1. The interpolated visual embedding and score are defined as:
\begin{equation} \label{equ:feat_inter}
    \underbrace{(1 - \delta) \cdot M(x_R) + \delta \cdot M(x_S)}_{\text{Interpolated Visual Embedding}}
     \Rightarrow \underbrace{(1 - \delta) \cdot 0 + \delta \cdot 1 = \delta}_{\text{Interpolated Score}},
\end{equation}
\noindent where $M$ is the pre-trained CLIP visual encoder and $\delta$ is the interpolation coefficient. 
In our implementation, for each data batch, we randomly select 50\% of the samples to perform feature interpolation using a random $\delta$.
These two enhancements together lead to around 10\% accuracy improvement.
More details (including ablation studies) can be found in \autoref{app:diff_clip}, \autoref{app:overfit} and \autoref{app:ablation_feat_inter}.



\subsection{Combining Semantic and Artifact Features}
While enhancing each type of detector improves their individual performance, our primary goal is to combine them in a way that mutually amplifies their effectiveness. The approach involves training a classifier that leverages features from both detectors, with a focus on avoiding overfitting to any specific feature set.
We hence design an adaptive fusion training strategy by incorporating two regulators that dynamically assess the importance of the two respective sets of features, balancing their weights accordingly. To further reduce overfitting,
we also apply random dropout.
An overview of \ours{} is illustrated in \autoref{fig:overview}, and the network architecture is detailed in \autoref{fig:arch}. The framework comprises three main modules: (1) Semantic feature extraction, (2) Artifact extraction, and (3) Adaptive Fusion.

\smallskip \noindent \textbf{\circled{1} Semantic Extraction} (blue region of \autoref{fig:overview}):
As discussed in \autoref{sec:design_semantic}, we utilize (CLIP) ViT-SO400M-14-384~\cite{ViT-SO400M-14-384} as our semantic feature extractor, which remains frozen during training. Additionally, we apply feature interpolation, as defined in Equation~\ref{equ:feat_inter}, as a data augmentation technique. The resulting semantic feature is denoted as $v_{sem}$.

\smallskip \noindent \textbf{\circled{2} Artifact Extraction} (green region of \autoref{fig:overview}):
Artifacts are extracted using the VAE from SD-v1.5~\cite{stablediffusion}, following Equation~\ref{equ:vae}. A ResNet-50~\cite{resnet} encoder is employed to extract features from the derived artifacts, resulting in $v_{art}$.

\smallskip \noindent \textbf{\circled{3} Adaptive Fusion} (yellow region of \autoref{fig:overview}):
To dynamically combine semantic and artifact features, we design a regulator module that maps each feature to a scaling coefficient before concatenation. Specifically,
\begin{gather} \label{equ:aggr}
    \underbrace{\alpha = R_{sem}(v_{sem})}_{\text{Regulate Semantic Features}}, \quad
    \underbrace{\beta = R_{art}(v_{art})}_{\text{Regulate Artifact Features}},
    \notag \\
    \underbrace{v = \alpha * v_{sem} \bigoplus \beta * v_{art}}_{\text{Adaptive Fusion}},
\end{gather}

\noindent where $R_{sem}$ and $R_{art}$ are the regulators for semantic and artifact features, respectively. The coefficients $\alpha$ and $\beta$ scale the respective features, and the aggregated feature $v$ is obtained by concatenating the weighted semantic and artifact features.
Additionally, we apply random dropout to the features before aggregation by randomly setting $\alpha$ or $\beta$ to 0 during training, preventing overfitting to either feature. Both coefficients are never set to 0 simultaneously to avoid $v$ becoming a zero vector.
Finally, a classification layer (fully connected layer) maps the aggregated feature $v$ to an output score, representing the probability of the input being synthetic.
Our evaluation in \autoref{sec:eval} demonstrates that \ours{} outperforms the best baseline by an average of 11\% accuracy, better than the improvement achieved by the enhanced artifact or semantic detector alone (i.e., 3\% and 7\%, respectively).

\section{Evaluation} \label{sec:eval}
In this section, we conduct an extensive evaluation of \ours. We first introduce the experiment setup and then detail the results and observations.

\begin{table*}[t]
    \centering
    \scriptsize
    \tabcolsep=3.3pt
    \caption{\textbf{Comparison with existing baselines, trained on DRCT~\cite{drct} and evaluated on \ourtest{}.} The results are measured in average precision (AP) and accuracy, with a decision threshold of 0.5. The highest AP scores are highlighted in \textcolor{myred}{red}, and the highest accuracy scores are highlighted in \textcolor{myblue}{blue}. Note that only \ours's results are highlighted if they match the best performance achieved by the baselines.}
    \label{tab:eval_custom}
    \begin{tabular}{lgcgcgcgcgcgcgcgcgcgc}
    \toprule
    \multirow{2}{*}{\textbf{Detector}} & \multicolumn{2}{c}{\textbf{CNNDet}} & \multicolumn{2}{c}{\textbf{FreqFD}} & \multicolumn{2}{c}{\textbf{Fusing}} & \multicolumn{2}{c}{\textbf{LNP}} & \multicolumn{2}{c}{\textbf{UnivFD}} & \multicolumn{2}{c}{\textbf{DIRE}} & \multicolumn{2}{c}{\textbf{FreqNet}} & \multicolumn{2}{c}{\textbf{NPR}} & \multicolumn{2}{c}{\textbf{DRCT}} & \multicolumn{2}{c}{\ours} \\
    \cmidrule(lr){2-3} \cmidrule(lr){4-5} \cmidrule(lr){6-7} \cmidrule(lr){8-9} \cmidrule(lr){10-11} \cmidrule(lr){12-13} \cmidrule(lr){14-15} \cmidrule(lr){16-17} \cmidrule(lr){18-19} \cmidrule(lr){20-21}
    ~ & \cellcolor{white}{AP} & Acc. & \cellcolor{white}{AP} & Acc. & \cellcolor{white}{AP} & Acc. & \cellcolor{white}{AP} & Acc. & \cellcolor{white}{AP} & Acc. & \cellcolor{white}{AP} & Acc. & \cellcolor{white}{AP} & Acc. & \cellcolor{white}{AP} & Acc. & \cellcolor{white}{AP} & Acc. & \cellcolor{white}{AP} & Acc. \\
    \midrule
LDM & 90.57 & 77.56 & 74.88 & 54.17 & 98.18 & 83.03 & 96.04 & 84.87 & 85.76 & 79.07 & 86.46 & 66.25 & 92.37 & 74.92 & 93.45 & 84.34 & 88.28 & 79.25 & \textcolor{myred}{98.91} & \textcolor{myblue}{95.04} \\
SD-v1.4 & 97.00 & 89.95 & 92.40 & 62.91 & \textcolor{myred}{99.95} & \textcolor{myblue}{99.16} & 99.12 & 95.92 & 88.35 & 80.87 & 96.51 & 83.55 & 90.55 & 69.20 & 96.88 & 90.90 & 91.61 & 81.18 & 97.80 & 91.95 \\
SD-v1.5 & 97.03 & 89.75 & 92.30 & 62.56 & \textcolor{myred}{99.96} & \textcolor{myblue}{99.12} & 99.23 & 96.21 & 88.57 & 80.88 & 96.72 & 83.77 & 90.33 & 68.86 & 97.05 & 91.30 & 91.08 & 81.06 & 98.02 & 91.31 \\
SSD-1B & 87.35 & 66.55 & 48.90 & 49.72 & 84.65 & 53.96 & 93.81 & 79.12 & 86.46 & 76.47 & 74.59 & 56.63 & 50.28 & 49.18 & 52.84 & 47.87 & 82.04 & 75.83 & \textcolor{myred}{95.40} & \textcolor{myblue}{83.20} \\
tiny-sd & 87.44 & 66.37 & 80.01 & 52.19 & \textcolor{myred}{98.01} & 77.12 & 95.03 & 81.48 & 84.58 & 76.96 & 87.83 & 63.65 & 88.03 & 63.56 & 95.67 & \textcolor{myblue}{88.42} & 88.06 & 79.99 & 95.99 & 84.80 \\
SegMoE-SD & 91.12 & 74.41 & 80.21 & 51.74 & 97.16 & 73.58 & 96.36 & 86.62 & 89.59 & 83.07 & 88.95 & 65.98 & 88.55 & 64.40 & 97.21 & \textcolor{myblue}{93.79} & 79.29 & 75.12 & \textcolor{myred}{97.39} & 89.49 \\
small-sd & 89.78 & 70.15 & 81.81 & 52.57 & \textcolor{myred}{99.06} & 82.65 & 94.81 & 80.75 & 85.67 & 77.45 & 91.31 & 68.38 & 89.44 & 65.42 & 95.77 & \textcolor{myblue}{89.14} & 90.08 & 81.20 & 96.22 & 85.80 \\
SD-2-1 & 86.78 & 68.14 & 52.95 & 49.93 & 92.64 & 59.32 & 81.26 & 57.19 & 89.00 & 81.74 & 88.11 & 65.25 & 64.40 & 51.62 & 71.62 & 51.31 & 81.60 & 76.12 & \textcolor{myred}{96.89} & \textcolor{myblue}{88.53} \\
SD-3-medium & 79.00 & 60.68 & 57.99 & 49.98 & 81.86 & 52.47 & 75.08 & 53.69 & 87.62 & 78.42 & 76.64 & 56.75 & 57.19 & 49.64 & 71.36 & 50.00 & 79.95 & 74.95 & \textcolor{myred}{95.04} & \textcolor{myblue}{82.91} \\
SDXL-turbo & 96.42 & 88.67 & 92.98 & 61.34 & 95.19 & 59.69 & 95.07 & 83.47 & 90.58 & 84.31 & 90.97 & 72.97 & 87.04 & 66.62 & 94.63 & 83.57 & 90.46 & 80.36 & \textcolor{myred}{99.17} & \textcolor{myblue}{95.39} \\
SD-2 & 85.93 & 65.73 & 50.79 & 49.84 & 89.08 & 55.92 & 76.79 & 54.56 & 83.24 & 73.78 & 83.69 & 60.07 & 59.28 & 50.64 & 72.97 & 51.19 & 80.13 & 75.14 & \textcolor{myred}{94.94} & \textcolor{myblue}{83.67} \\
SDXL & 83.39 & 61.79 & 43.93 & 49.70 & 76.81 & 51.01 & \textcolor{myred}{94.00} & \textcolor{myblue}{80.33} & 72.48 & 63.64 & 64.48 & 51.95 & 47.90 & 48.84 & 46.99 & 47.75 & 80.62 & 75.18 & 91.68 & 74.12 \\
PG-v2.5-1024 & 65.10 & 53.65 & 47.54 & 49.70 & 75.22 & 50.41 & 93.46 & 79.07 & 82.98 & 78.23 & 61.97 & 52.32 & 55.09 & 48.67 & 50.95 & 47.71 & 79.16 & 71.33 & \textcolor{myred}{96.45} & \textcolor{myblue}{88.65} \\
PG-v2-1024 & 83.85 & 63.48 & 48.93 & 49.70 & 85.62 & 52.08 & 74.18 & 52.48 & 83.77 & 78.55 & 76.95 & 56.63 & 53.95 & 48.79 & 63.62 & 48.40 & 70.06 & 66.62 & \textcolor{myred}{96.72} & \textcolor{myblue}{89.14} \\
PG-v2-512 & 77.73 & 57.94 & 55.59 & 49.87 & 74.90 & 51.58 & 59.60 & 49.40 & 69.21 & 58.90 & 71.35 & 53.63 & 45.21 & 49.09 & 65.15 & 49.09 & 83.55 & \textcolor{myblue}{77.61} & \textcolor{myred}{85.02} & 64.86 \\
PG-v2-256 & 81.40 & 63.30 & 57.88 & 50.19 & 75.17 & 51.10 & 72.63 & 54.77 & 72.40 & 62.99 & 81.13 & 60.32 & 49.26 & 49.47 & 60.43 & 49.74 & 78.19 & \textcolor{myblue}{73.55} & \textcolor{myred}{90.22} & 72.92 \\
PAXL-2-1024 & 71.18 & 56.29 & 54.28 & 49.76 & 86.44 & 53.60 & 74.24 & 53.41 & 84.97 & 80.08 & 70.61 & 54.77 & 64.81 & 49.40 & 72.84 & 51.16 & 76.22 & 71.80 & \textcolor{myred}{97.94} & \textcolor{myblue}{93.94} \\
PAXL-2-512 & 83.05 & 65.44 & 80.53 & 52.20 & 95.97 & 68.77 & 91.30 & 74.25 & 85.36 & 80.32 & 81.29 & 62.18 & 82.72 & 57.23 & 94.24 & 81.25 & 79.95 & 75.53 & \textcolor{myred}{98.63} & \textcolor{myblue}{94.96} \\
LCM-sdxl & 93.11 & 81.55 & 81.10 & 52.44 & 97.74 & 70.75 & 95.96 & 85.46 & 81.52 & 78.04 & 89.14 & 68.57 & 85.17 & 62.29 & 70.29 & 50.87 & 91.96 & 81.05 & \textcolor{myred}{98.72} & \textcolor{myblue}{96.20} \\
LCM-sdv1-5 & 97.67 & 92.29 & 94.70 & 68.17 & 98.76 & 81.22 & 97.87 & 90.87 & 83.87 & 79.67 & 93.58 & 79.02 & 93.14 & 76.20 & 98.37 & 93.71 & 87.27 & 79.59 & \textcolor{myred}{99.63} & \textcolor{myblue}{97.14} \\
FLUX.1-sch & 71.72 & 56.04 & 56.01 & 50.02 & 76.74 & 51.03 & 74.75 & 54.38 & 80.14 & 72.89 & 73.67 & 56.27 & 64.35 & 50.39 & 77.91 & 53.14 & 74.38 & 68.39 & \textcolor{myred}{95.52} & \textcolor{myblue}{85.24} \\
FLUX.1-dev & 70.75 & 57.44 & 53.60 & 50.04 & 82.30 & 52.92 & 73.25 & 54.13 & 82.50 & 75.95 & 73.74 & 56.32 & 53.83 & 49.06 & 71.09 & 50.74 & 77.88 & 70.70 & \textcolor{myred}{96.16} & \textcolor{myblue}{86.10} \\
\midrule
\textbf{Average} & 84.88 & 69.42 & 67.24 & 53.12 & 89.15 & 65.02 & 86.54 & 71.93 & 83.57 & 76.47 & 81.80 & 63.42 & 70.59 & 57.43 & 77.79 & 65.70 & 82.81 & 75.98 & \textcolor{myred}{96.02} & \textcolor{myblue}{87.06} \\
    \bottomrule
    \end{tabular}
\end{table*}

\begin{table*}[t]
    \centering
    \scriptsize
    \tabcolsep=3.3pt
    \caption{\textbf{Comparison with existing baselines, trained on DRCT~\cite{drct} and evaluated on \ourtest{}/in-the-wild.} The results are measured in average precision (AP) and accuracy, with a decision threshold of 0.5. The highest AP scores are highlighted in \textcolor{myred}{red}, and the highest accuracy scores are highlighted in \textcolor{myblue}{blue}. Note that only \ours's results are highlighted if they match the best performance achieved by the baselines.}
    \label{tab:eval_wild}
    \begin{tabular}{lgcgcgcgcgcgcgcgcgcgc}
    \toprule
    \multirow{2}{*}{\textbf{Detector}} & \multicolumn{2}{c}{\textbf{CNNDet}} & \multicolumn{2}{c}{\textbf{FreqFD}} & \multicolumn{2}{c}{\textbf{Fusing}} & \multicolumn{2}{c}{\textbf{LNP}} & \multicolumn{2}{c}{\textbf{UnivFD}} & \multicolumn{2}{c}{\textbf{DIRE}} & \multicolumn{2}{c}{\textbf{FreqNet}} & \multicolumn{2}{c}{\textbf{NPR}} & \multicolumn{2}{c}{\textbf{DRCT}} & \multicolumn{2}{c}{\ours} \\
    \cmidrule(lr){2-3} \cmidrule(lr){4-5} \cmidrule(lr){6-7} \cmidrule(lr){8-9} \cmidrule(lr){10-11} \cmidrule(lr){12-13} \cmidrule(lr){14-15} \cmidrule(lr){16-17} \cmidrule(lr){18-19} \cmidrule(lr){20-21}
    ~ & \cellcolor{white}{AP} & Acc. & \cellcolor{white}{AP} & Acc. & \cellcolor{white}{AP} & Acc. & \cellcolor{white}{AP} & Acc. & \cellcolor{white}{AP} & Acc. & \cellcolor{white}{AP} & Acc. & \cellcolor{white}{AP} & Acc. & \cellcolor{white}{AP} & Acc. & \cellcolor{white}{AP} & Acc. & \cellcolor{white}{AP} & Acc. \\
    \midrule
Civitai & 88.08 & 76.85 & 82.61 & 55.95 & 97.22 & 80.40 & 95.83 & \textcolor{myblue}{89.85} & 66.05 & 61.25 & \textcolor{myred}{98.34} & 88.60 & 89.13 & 83.80 & 89.34 & 86.20 & 40.94 & 43.30 & 95.82 & 88.95 \\
DALL-E 3 & 53.83 & 51.10 & 50.23 & 49.68 & 71.33 & 51.50 & 74.61 & 65.18 & 71.14 & 62.98 & 70.33 & 54.33 & 35.70 & 45.52 & 37.12 & 42.98 & 70.33 & 62.62 & \textcolor{myred}{83.94} & \textcolor{myblue}{77.03} \\
instavibe.ai & 59.01 & 53.60 & 42.76 & 50.12 & 65.02 & 50.90 & 31.00 & 44.42 & \textcolor{myred}{77.97} & 64.35 & 52.89 & 50.82 & 34.47 & 45.67 & 31.59 & 42.15 & 68.27 & 59.80 & 75.21 & \textcolor{myblue}{67.73} \\
Lexica & 59.45 & 53.08 & 40.72 & 49.43 & 67.43 & 51.38 & 33.59 & 45.00 & 79.58 & 64.71 & 73.24 & 53.52 & 37.21 & 45.95 & 42.92 & 43.38 & \textcolor{myred}{86.70} & 65.65 & 85.62 & \textcolor{myblue}{78.42} \\
Midjourney-v6 & 44.73 & 48.62 & 39.77 & 49.58 & 55.94 & 49.98 & 37.43 & 45.80 & 67.13 & 60.93 & 61.69 & 51.90 & 35.03 & 45.42 & 35.26 & 42.23 & 60.34 & 57.53 & \textcolor{myred}{82.65} & \textcolor{myblue}{74.92} \\
\midrule
\textbf{Average} & 61.02 & 56.65 & 51.22 & 50.95 & 71.39 & 56.83 & 54.49 & 58.05 & 72.37 & 62.84 & 71.30 & 59.84 & 46.31 & 53.27 & 47.24 & 51.39 & 65.32 & 57.78 & \textcolor{myred}{84.65} & \textcolor{myblue}{77.41} \\
    \bottomrule
    \end{tabular}
\end{table*}

\subsection{Experiment Setup} \label{sec:exp_setup}
\smallskip \noindent
\textbf{Dataset and Pre-processing.}
We consider two well-established training sets, i.e., CNNDet/ProGAN~\cite{cnndet} and DRCT-2M/SD-v1.4~\cite{drct}. CNNDet/ProGAN training set consists of 20 distinct categories with 18,000 synthetic images generated by ProGAN each, along with an equal number of real images from LSUN~\cite{lsun}, altogether 720k images.
DRCT-2M/SD-v1.4 leverages SD-v1.4~\cite{stablediffusion} to produce synthetic images using captions from MSCOCO-2017~\cite{mscoco} and includes an equal number of real images from MSCOCO, altogether 236k images.
In the following sections, we use CNNDet and DRCT to denote the two training sets for simplicity.
We employ four test sets for a comprehensive assessment of the detection performance: (1) AIGCDetectBenchmark~\cite{aigcdetect} including synthetic images from 16 different generative models (half GANs and half diffusion models) with equal number of real images from the source training set of each generative model; (2) GenImage~\cite{genimage} consisting of 8 types of synthetic images with real ones; (3) \ourtest{}; and (4) \ourtest{}/in-the-wild, introduced in \autoref{app:detail_data}.
Following the existing settings~\cite{cnndet,univfd,drct}, all images are resized to 224 $\times$ 224 for both training and test phases, to ensure a fair comparison. In addition, we randomly apply JPEG compression with quality ranging from 75 to 95 with 50\% possibility to the inputs. Our intention is to mimic the practical scenarios where users typically compress their images before uploading, and to avoid the detectors to overfit on the format differences~\cite{fakeorjpeg}, e.g., most real images are saved in JPEG while PNG for synthetic images.
More details of used generative models can be found in \autoref{app:details_models}.

\smallskip \noindent
\textbf{Evaluation Metrics.} We use AP (Average Precision) and Acc. (Accuracy) with threshold 0.5, as two main metrics to evaluate the detection performance. We also consider F1, ROC-AUC scores and TPRs (True Positive Rates) at low FPRs (False Positive Rates) in \autoref{app:more_metrics}.

\smallskip \noindent
\textbf{Baseline Methods and Implementation.} We consider ten state-of-the-art detection methods to compare with, including CNNDet~\cite{cnndet}, FreqFD~\cite{freqfd}, Fusing~\cite{fusing}, LNP~\cite{lnp}, LGrad~\cite{lgrad}, UnivFD~\cite{univfd}, DIRE~\cite{dire}, FreqNet~\cite{freqnet}, NPR~\cite{npr}, and DRCT~\cite{drct}. We follow the official implementation to conduct the experiments. For DRCT, we adopt CLIP as its backbone model and use SD-v1.4 to perform the reconstruction, not introducing extra knowledge. As LGrad is typically designed to detect GAN-based synthetic images and DRCT targets diffusion model–generated images, we don't force them on unintended tasks. More details about the baseline detectors can be found in \autoref{app:detail_baseline}.

\begin{figure*}[t]
    \begin{minipage}[t]{0.135\linewidth}
        \centering
        \includegraphics[width=1\linewidth]{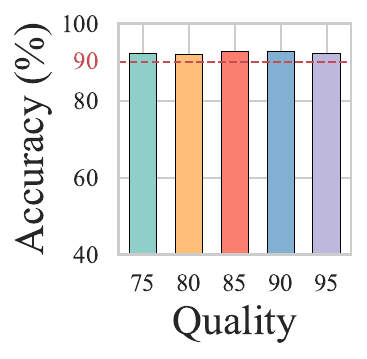}
        \subcaption{JPEG}
    \end{minipage}
    \hfill
    \begin{minipage}[t]{0.135\linewidth}
        \centering
        \includegraphics[width=1\linewidth]{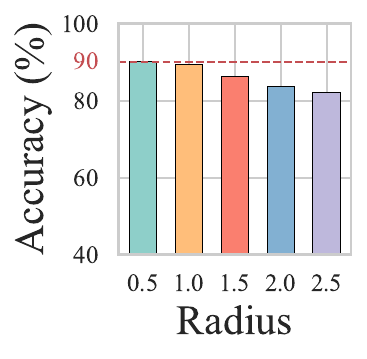}
        \subcaption{Blur}
    \end{minipage}
    \hfill
    \begin{minipage}[t]{0.135\linewidth}
        \centering
        \includegraphics[width=1\linewidth]{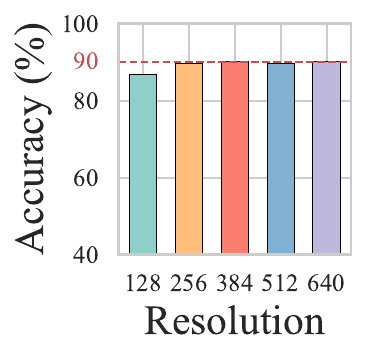}
        \subcaption{Resize}
    \end{minipage}
    \hfill
    \begin{minipage}[t]{0.135\linewidth}
        \centering
        \includegraphics[width=1\linewidth]{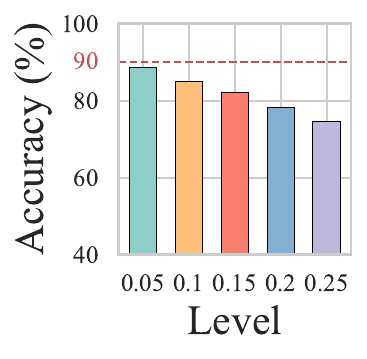}
        \subcaption{Noise}
    \end{minipage}
    \hfill
    \begin{minipage}[t]{0.135\linewidth}
        \centering
        \includegraphics[width=1\linewidth]{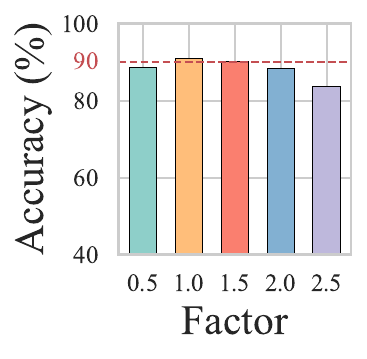}
        \subcaption{Brightness}
    \end{minipage}
    \hfill
    \begin{minipage}[t]{0.135\linewidth}
        \centering
        \includegraphics[width=1\linewidth]{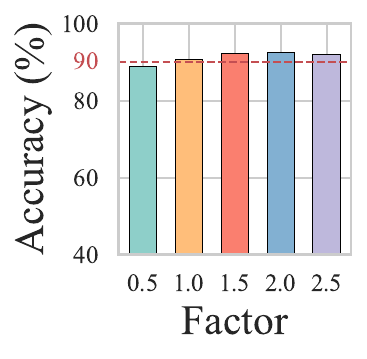}
        \subcaption{Saturation}
    \end{minipage}
    \hfill
    \begin{minipage}[t]{0.135\linewidth}
        \centering
        \includegraphics[width=1\linewidth]{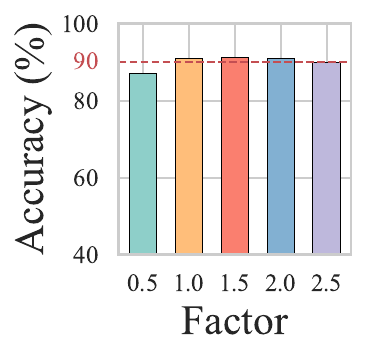}
        \subcaption{Contrast}
    \end{minipage}
    \caption{\textbf{Robustness evaluation of \ours{} on \ourtest/SD-v1.5 with an equal number of real images.} Each subfigure denotes robustness evaluation against different transformation functions. In each subfigure, the x-axis denotes the varying parameters and the y-axis denotes the accuracy.}
    \vspace{-5pt}
    \label{fig:eval_robust}
\end{figure*}

\subsection{Effectiveness and Comparison with Baselines} \label{sec:eval_main}
In this section, we compare the detection performance of \ours{} with existing baselines, primarily focusing on out-of-distribution detection, i.e., test set is highly different from the training set.

\smallskip \noindent
\textbf{Results on DRCT training set.}
Recently, most synthetic images available in the wild are generated by diffusion models, particularly text-to-image diffusion models. We train \ours{} using the DRCT training set, which comprises synthetic images produced by SD-v1.4 and real images from MSCOCO. We then compare its performance against baseline detectors trained on the same dataset. To simulate real-world conditions, we apply JPEG compression to the input images.
We evaluate \ours{} and the baseline detectors on \ourtest{}, with the results presented in \autoref{tab:eval_custom}. We observe that \ours{} consistently outperforms the baselines in nearly all scenarios, achieving an average AP of 96.02\% and an accuracy of 87.06\%. This exceeds the best baseline results by approximately 7\% in AP and 11\% in accuracy. However, for some older models, such as SD-v1.4 and SD-v1.5, \ours{} performs slightly worse than Fusing. This may be because Fusing overfits to the specific generation styles of SD-v1.4, given that the synthetic images in the training set are generated by SD-v1.4. In contrast, Fusing struggles to detect newer synthetic images, like those from FLUX, whereas \ours{} significantly outperforms it by over 30\% in accuracy. This improvement is attributed to \ours{}’s comprehensive approach, which integrates both semantic and artifact-based clues.
We also provide evaluation results on the GenImage~\cite{genimage} test set. As shown in \autoref{app:eval_genimage}, \ours{} outperforms baselines.

Moreover, we conduct experiments on \ourtest{} /in-the-wild containing synthetic images from the Internet and an equal number of real images from CC3M~\cite{cc3m}.
For each source, we randomly select 2,000 real and 2,000 synthetic images for evaluation. The results, shown in \autoref{tab:eval_wild}, demonstrate that \ours{} also outperforms existing baselines, achieving 14\% higher accuracy than the best baseline.
These findings highlight effectiveness of \ours{} for real-world deployment in synthetic image detection.

We also compare \ours{} with the latest baseline AIDE~\cite{aide}. Results in \autoref{app:aide} show \ours{} outperforms it, particularly in diverse testbeds with lossy formats.

\smallskip \noindent
\textbf{Results on CNNDet training set.}
We conduct experiments on the CNNDet training set and evaluate the performance of the converged detectors on the AIGCDetectBenchmark, which comprises synthetic images generated by 16 different generative models. The results presented in \autoref{app:progan} shows that \ours{} achieves an average accuracy improvement of 8\% over the best baseline, UnivFD.

\subsection{Robustness against Various Transformation} \label{sec:eval_robust}
We further evaluate the robustness of \ours{} against various input transformations that may occur in real-world scenarios. Specifically, we apply seven types of transformation functions, each with diverse parameters. These include JPEG compression with quality levels ranging from 75 to 95, random blurring using Gaussian radius between 0.5 and 2.5, random resizing to resolutions from 128 to 640 pixels, and the addition of random Gaussian noise with levels varying from 0.05 to 0.25. Additionally, we adjust brightness, saturation, and contrast with factors between 0.5 and 2.5. A demonstration of these transformation functions is presented in \autoref{app:illu_robust}.
We evaluate \ours{}, trained on the DRCT training set, on \ourtest/SD-v1.5 using an equal number of real images. The results are illustrated in \autoref{fig:eval_robust}, where \ours{} generally maintains robust performance across most transformation functions. However, its performance slightly decreases under high-strength random blurring and noise, which is expected since severe blurring or noise can distort the artifacts and also disrupt the semantic content of the inputs. Despite these challenges, \ours{} still achieves over 75\% accuracy even in such extreme cases. These findings demonstrate the effectiveness of \ours{} in handling various post-processing transformations.

\subsection{Ablation Study} \label{sec:eval_ablation}
We conduct an ablation study of \ours{} to investigate the integration methods for semantic and artifact features. The results presented in \autoref{app:eval_ablation} demonstrate that our adaptive fusion approach outperforms six straightforward methods, such as simply concatenating the two feature vectors.

\section{Related Work} \label{sec:background}

\textbf{Synthetic Image Generation.}
Synthetic image generation has rapidly advanced through deep learning over the past decade. Early models like VAE~\cite{vae} and GANs~\cite{gan,progan} enabled automated creation but lacked realism and control. Since 2020, diffusion models such as DDPM~\cite{ddpm}, LDM~\cite{stablediffusion} and Stable Diffusion~\cite{stablediffusion} have achieved higher fidelity and enhanced text-based control. Today, diffusion models are capable of generating high-quality images in diverse styles by applications like Midjourney~\cite{midjourney} and DALL-E~\cite{dalle3}.

\noindent \textbf{Image Watermarking.}
Image watermarking~\cite{watermark_0,watermark_1,watermark_2,watermark_3} prevents misuse by embedding signatures in generated images. Traditional methods add invisible noise that can be disrupted by cropping or blurring. Recent techniques~\cite{treering} embed watermarks in the latent space for greater resilience, though they may impact generative performance.

\noindent \textbf{Fake Image Detection.}
This paper focuses on detecting fake images without using watermarking~\cite{fatformer,cnndet,univfd,drct}. Early methods like CNNDet~\cite{cnndet} employed binary classifiers (e.g., ResNet-50) trained on real and ProGAN-generated images. Subsequent approaches~\cite{fusing,drct,univfd,fatformer,c2p,sanity} enhance detection based on semantic features, by incorporating multi-modal analysis~\cite{chameleon}, pre-trained feature extractors~\cite{clip}, and contrastive learning~\cite{contrast}. Additionally, techniques like FreqFD~\cite{freqfd}, LGrad~\cite{lgrad}, FreqNet~\cite{freqnet} and NPR~\cite{npr} identify unique artifacts in generated images, improving the generalization ability across various models and objects.
It is important to note that our focus is on detecting images generated by noise-to-image and text-to-image models. This is distinct from other research~\cite{det_edit_0,det_edit_1} that targets \textit{image editing}, which primarily identifies edited or modified image regions.
Our efforts can be complementary to these studies.

\section{Conclusion} \label{sec:conclude}
We propose a novel synthetic image detection framework that integrates enhanced semantic and artifact features, achieving greater generalization and robustness.
Our evaluations show an average accuracy improvement of 11\% over baselines.
These results underscore the need for improved detection reliability in real-world scenarios, supporting broader efforts to counter misinformation and protect digital content integrity.

\section*{Acknowledgements}
We thank the anonymous reviewers for their constructive comments. This research was supported by Sony AI, IARPA TrojAI W911NF-19-S0012, NSF 1901242 and 1910300, ONR N000141712045, N000141410468 and N000141712947. Any opinions, findings, and conclusions in this paper are those of the authors only and do not necessarily reflect the views of our sponsors.

{
    \small
    \bibliographystyle{ieeenat_fullname}
    \bibliography{reference}
}

\clearpage
\newpage
\onecolumn
\appendix

\section*{Appendix}

\section{Details of Baseline Detectors} \label{app:detail_baseline}
In this section, we introduce the baseline methods utilized in our experiments.
\begin{enumerate}[label=\textbullet]
    \item \textbf{CNNDet}~\cite{cnndet}: This pioneering work collects a well-established dataset comprising real images from LSUN~\cite{lsun} and fake images generated by ProGAN~\cite{progan}. It trains a ResNet-50~\cite{resnet} on this dataset, incorporating random JPEG compression and Gaussian blurring as data augmentation techniques. The study demonstrates that fake images generated by early generative models, such as GANs, are relatively easy to detect.
    \item \textbf{FreqFD}~\cite{freqfd}: This paper analyzes the differences between real and fake images in the frequency domain, highlighting that up-sampling operations typically introduce texture-level artifacts in the spatial domain. Consequently, it trains a classifier on images after applying the DCT, focusing on detection in the frequency domain.
    \item \textbf{Fusing}~\cite{fusing}: This method performs detection on inputs at multiple scales and fuses the features from these different scales to make a final decision. It generally improves detection performance compared to CNNDet.
    \item \textbf{LNP}~\cite{lnp}: This approach observes that the noise patterns of real images exhibit consistent characteristics in the frequency domain, whereas fake images differ significantly. It trains a detector to distinguish image authenticity based on these noise patterns.
    \item \textbf{LGrad}~\cite{lgrad}: This method notes that the gradients of fake and real images are distinguishable when approximated using a pre-trained StyleGAN model. It trains a classifier on these gradient features to differentiate between real and fake images.
    \item \textbf{UnivFD}~\cite{univfd}: This approach leverages a pre-trained CLIP model as a feature extractor and performs classification based on the extracted embeddings.
    \item \textbf{DIRE}~\cite{dire}: This method observes that fake images generated by diffusion models can be easily reconstructed using a pre-trained diffusion model, such as ADM~\cite{adm}. It trains a classifier on the reconstruction errors to distinguish between real and fake images.
    \item \textbf{FreqNet}~\cite{freqnet}: This approach enhances frequency artifacts by designing a sophisticated network that conducts an in-depth extraction of the frequency footprints of fake images.
    \item \textbf{NPR}~\cite{npr}: This method extracts up-sampling artifacts by up-sampling the inputs and comparing nearby patch differences. This simulation effectively reveals generative footprints.
    \item \textbf{DRCT}~\cite{drct}: This approach leverages a pre-trained diffusion model, such as SD-v1.4~\cite{stablediffusion}, to reconstruct both real and fake images. It then applies contrastive loss to train the classifier to distinguish reconstructed real images (serving as hard fake samples) from the original real images.
\end{enumerate}

\section{Details of Used Models} \label{app:details_models}

In this section, we introduce the details of the image generation models used in our experiments. 
\begin{enumerate}[label=\textbullet]
\item \textbf{Generative Adversarial Network (GAN)~\cite{gan}.} GAN is a representative class of generative models.
In a GAN, two neural networks (i.e., generator and discriminator) compete in a zero-sum game, where the gain of one network comes at the loss of the other.
Multiple GAN models are involved in our experiments, including
ProGAN~\cite{karras2017progressive},
StyleGAN~\cite{karras2019style},
StyleGAN2~\cite{Karras2019stylegan2},
BigGAN~\cite{brock2018large},
CycleGAN~\cite{zhu2017unpaired},
StarGAN~\cite{choi2018stargan},
GauGAN~\cite{park2019semantic},
and whichfaceisreal (WFIR)~\cite{wfir}.


\item \textbf{Ablated Diffusion Model (ADM)~\cite{dhariwal2021diffusion}.}
This is a relatively early diffusion model developed by OpenAI. It is capable of achieving conditional generation by leveraging gradients from a classifier.
It is open-sourced with MIT license.

\item \textbf{Glide~\cite{nichol2022glide}.} This is a 3.5 billion-parameter diffusion model developed by OpenAI, which uses a text encoder to condition on natural language descriptions with classifier-free guidance. This model is with MIT license.

\item \textbf{VQ-Diffusion (VQDM)~\cite{gu2022vector}.} This model is built on a vector quantized variational autoencoder (VQ-VAE), with its latent space modeled using a conditional version of the Denoising Diffusion Probabilistic Model. This model is open-sourced with MIT license.

\item \textbf{wukong~\cite{wukong}.} This is a text-to-image diffusion model trained on Chinese text description and image pairs.

\item \textbf{Latent Diffusion Model (LDM)~\cite{stablediffusion}.}
This model is the first to implement the diffusion process within the latent space of pretrained autoencoders. It strikes a near-perfect balance between reducing complexity and preserving details, significantly enhancing visual quality while operating under constrained computational resources.  This model is with MIT license.

\item \textbf{Stable Diffusion (SD)~\cite{stablediffusion}.} Stable Diffusion is a series of models based on the LDM architecture with fixed, pretrained CLIP encoders. There are multiple Stable Diffusion models involved in our experiments, i.e., SD-v1.4, SD-v1.5, SD-2, SD-2-1, SDXL, SDXL-turbo, and SD-3-medium.
These models are with creativemlopenrail-m license.

\item \textbf{tiny-sd and small-sd~\cite{distill-sd}.}
These models are distilled from a fine-tuned 
SD-v1.5 model (SG161222/Realistic\_Vision\_V4.0).
Compared to the original model, the distilled models offer up to 100\% faster inference times and reduce VRAM usage by up to 30\%.
These models are with creativemlopenrail-m license.

\item \textbf{Segmind Stable Diffusion 1B (SSD-1B)~\cite{gupta2024progressive}.}
The SSD-1B is a distilled version of Stable Diffusion XL (SDXL), reduced by 50\% in size, delivering a 60\% increase in speed while still preserving high-quality text-to-image generation performance. This model is with apache-2.0 license.

\item \textbf{Segmind Mixture of Diffusion Experts (SegMoE-SD)~\cite{segmoe}.} This is a mixture of expert model combined by 4 SD-v1.5 models using the SegMoE merging framework. Comparing to the single model, this mixture of expert model has better adherence and better image quality. This model is with apache-2.0 license.

\item \textbf{Playground (PG)~\cite{li2024playground}.} Playground is a model family trained from scratch by the research team at Playground. These models are based on LDM architecture with two fixed, pre-trained text encoders (OpenCLIP-ViT/G and CLIP-ViT/L). Four models in this model family are used in our experiments: PG-v2-256, PG-v2-512, PG-v2-1024, and PG-v2.5-1025.
These models are with playground-v2-community and playground-v2dot5-community licenses.

\item \textbf{PixArt-XL (PAXL)~\cite{chen2023pixartalpha}.} These models reduces the training cost by decomposing training into three stages and using an efficient diffusion transformer. Two PAXL models with different generation resolutions (i.e., PAXL-2-512 and PAXL-2-1024) are used in our experiments.
These models are with openrail++ license.

\item \textbf{Latent Consistency Model (LCM)~\cite{luo2023latent}.}
The LCMs are distilled from pre-trained classifier-free guided diffusion models. These distilled models can directly predict the solution of the corresponding ODE in latent space, significantly reducing the need for multiple iterations. Specifically, two LCM models are involved in this paper: LCM-sdv1-5 and LCM-sdxl, which are distilled from SD-v1.5 and SDXL, respectively. These models are with openrail++ license.

\item \textbf{FLUX~\cite{flux}.} FLUX is a set of state-of-the-art text-to-image models developed by the Black Forest Lab. These models excel in prompt adherence, visual quality, image detail, and output diversity. In our experiments, we utilize FLUX.1-sch and FLUX.1-dev. The weights for these two models are open-sourced under the Apache-2.0 license and the FLUX-1-dev-non-commercial-license, respectively.


\item \textbf{DALL-E~\cite{dalle3}.} DALL-E is a series of closed-source text-to-image AI systems built by OpenAI. Both DALL-E 2 and DALL-E 3 are included in our experiments.



\item \textbf{Midjourney~\cite{midjourney}.}
Midjourney is a series of closed-source text-to-image models developed by Midjourney, Inc. In our experiments, we used version Midjourney-v6.

\item \textbf{Other In-the-Wild Sources.} 
We also incorporate additional in the wild sources to generate the images in \ourtest{}/in-the-wild. In addition to DALL-E 3 and Midjourney-v6, we use Civitai, instavibe.ai, and Lexica. These website platforms generate images based on models like Stable Diffusion~\cite{stablediffusion}, FLUX~\cite{flux}, and Lexica Aperture~\cite{lexica}, respectively.

\end{enumerate}

\section{Illustrations of JPEG Compression's Impact on Texture-level Artifacts} \label{app:illu_jpeg}

\begin{figure}[t]
    \centering
    \includegraphics[width=1\linewidth]{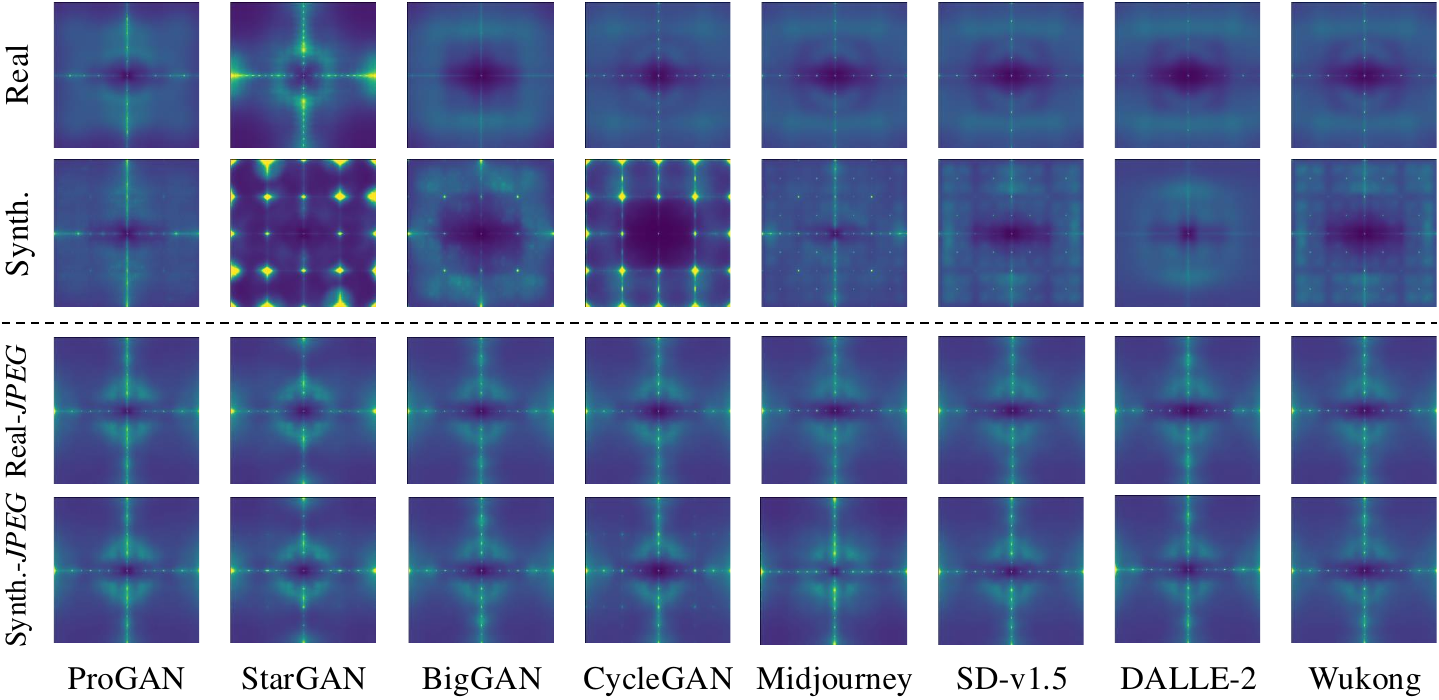}
    \caption{Frequency analysis of the impact of JPEG compression on texture-level artifacts}
    \label{fig:ana_freq}
\end{figure}

To further investigate how JPEG compression affects texture-level artifacts and why artifact detectors struggle with lossy formats (as outlined in \autoref{sec:motivation}), we conduct a frequency domain analysis. We compute the average frequency energy for 500 real images and 500 synthetic images. To ensure the analysis captures only the core content of the images, we first apply denoising with a pre-trained model~\cite{intriguing} before performing the Fourier transform.
\autoref{fig:ana_freq} presents the frequency representations of real images, synthetic images, and their JPEG-compressed versions in separate rows. In the absence of JPEG compression, synthetic images display abnormal patterns in the high-frequency regions (non-central areas) compared to real images. However, JPEG compression significantly reduces these differences between real and synthetic images, suggesting that compression diminishes the artifacts critical for detection.

\begin{figure}[t]
    \centering
    \begin{minipage}[t]{0.325\textwidth}
        \centering
        \includegraphics[height=0.63\textwidth]{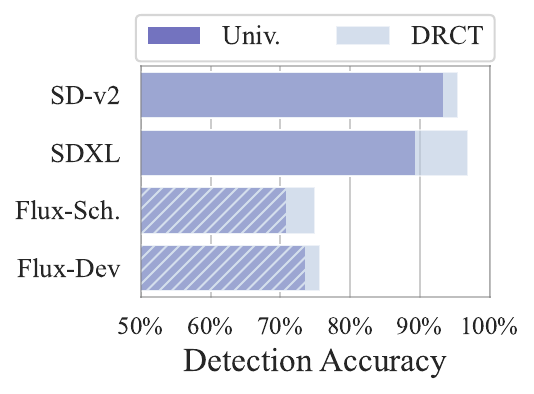}
        \subcaption{Advanced model}
        \label{fig:moti_flux}
    \end{minipage}
    \hfill
    \begin{minipage}[t]{0.325\textwidth}
        \centering
        \includegraphics[height=0.63\textwidth]{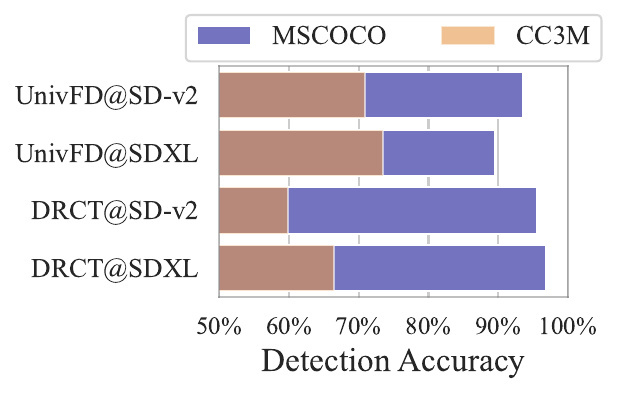}
        \subcaption{Diverse captions}
        \label{fig:moti_caption}
    \end{minipage}
    \hfill
    \begin{minipage}[t]{0.325\textwidth}
        \centering
        \includegraphics[height=0.63\textwidth]{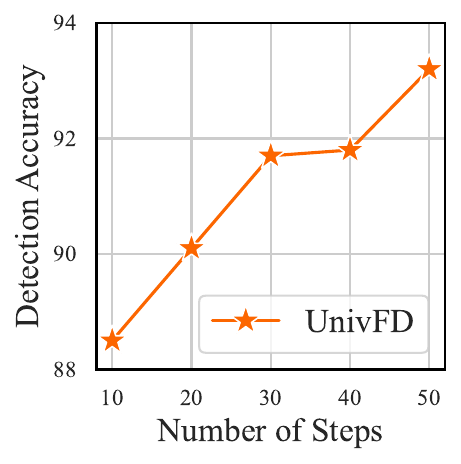}
        \subcaption{Number of steps}
        \label{fig:moti_steps}
    \end{minipage}
    \caption{Detection performance varies significantly across different types of fake image generation.}
    \label{fig:moti_data_limit}
\end{figure}

\section{Limitations of Existing Test Datasets} \label{app:limit_dataset}
To investigate the limitation of existing test datasets, we evaluate two latest detectors, UnivFD~\cite{univfd} and DRCT~\cite{drct}, both trained on the DRCT-2M/SD-v1.5 dataset (which includes real images from MSCOCO~\cite{mscoco} and fake images generated by SD-v1.5 using MSCOCO captions) across various test scenarios.

\smallskip \noindent
\underline{\textbf{Lack of Evaluation on Latest Models.}}
Synthetic images produced by the latest generative models tend to exhibit higher visual quality, making them more challenging to detect. To illustrate this issue, we evaluate the two detectors on synthetic images generated by SD-v2, SDXL, FLUX.1-schnell, and FLUX.1-dev~\cite{flux}. As shown in \autoref{fig:moti_flux}, both detectors perform well on SD-v2 and SDXL, which are included in DRCT-2M. However, their performance significantly degrades when applied to FLUX models, which are not covered by the existing dataset.

\smallskip \noindent
\underline{\textbf{Lack of Evaluation on Diverse Objects.}}
Synthetic images generated by text-to-image models can vary significantly based on the diversity of input captions, as different captions prompt the generation of various objects. Achieving high performance on a limited set of similar captions does not guarantee effective detection across a wider range of objects. To explore this, we evaluate two detectors on synthetic images generated by SD-v2 and SDXL using captions from MSCOCO (included in the training set) and CC3M~\cite{cc3m}. The results, shown in \autoref{fig:moti_caption}, reveal that while the detectors perform well on images generated from MSCOCO captions, their accuracy declines significantly on images generated from CC3M captions. This indicates that existing test sets do not sufficiently represent the diversity of image objects.

\smallskip \noindent
\underline{\textbf{Lack of Evaluation on Various Generation Parameters.}}
Additionally, existing synthetic image datasets often fix certain generation parameters, which can artificially inflate detection performance. For instance, DRCT typically uses 50 inference steps for all models. However, our observations indicate that the number of inference steps impacts detection performance. In \autoref{fig:moti_steps}, we evaluate UnivFD on SD-v2 synthetic images generated with varying numbers of inference steps. The results reveal that images generated with more inference steps are easier to detect, with a 5\% accuracy difference between images generated with 10 steps versus 50 steps.

\begin{figure}[t]
    \centering
    \includegraphics[width=0.8\linewidth]{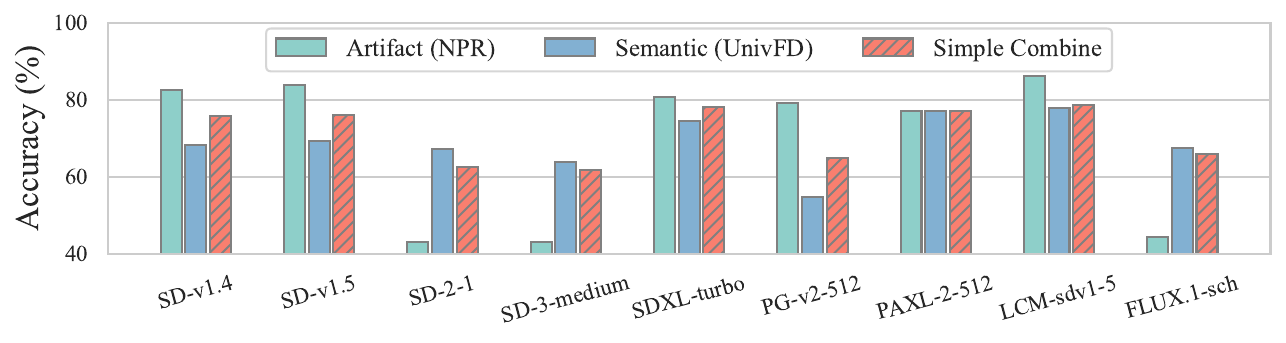}
    \caption{Limitation of simple combination of two types of detectors.}
    \label{fig:limit_simp_comb}
\end{figure}

\section{Limitation of Simply Combining Two Types of Detectors}  \label{app:limit_simp_comb}
We evaluate the effectiveness of directly combining two types of existing detectors to create a new one. In this experiment, we use the state-of-the-art artifact detector, NPR~\cite{npr}, and the semantic detector, UnivFD~\cite{univfd}. To combine the two detectors, we concatenate the artifact and semantic feature vectors before each downstream classifier and then retrain the classifier. The results, presented in \autoref{fig:limit_simp_comb}, leverage the DRCT-2M/SD-v1.4 dataset for training and test on synthetic images generated by various models with random JPEG compression. As shown, the simple combination merely averages the performance of the two detectors rather than enhancing it. This outcome arises because the two types of detectors are effective in distinct scenarios (as discussed in \autoref{sec:intro}), and direct combination fails to create a synergistic effect, yielding only an aggregate rather than a complementary result.

\section{Discussion of Overfitting Problem in Semantic Detector Training}  \label{app:overfit}
As discussed in \autoref{sec:motivation}, semantic detectors often struggle with generalization due to the overfitting problem. For example, a ResNet-50~\cite{resnet} trained on the CNNDet dataset~\cite{cnndet} with data augmentation may achieve perfect performance on unseen ProGAN images but fail to detect samples from BigGAN. This happens because the training data is limited to ProGAN-generated images, leading to overfitting on this specific model. UnivFD~\cite{univfd} mitigates this issue by using a pre-trained CLIP model (ViT-L-14-224~\cite{clip}) as a feature extractor without modifying its weights. This approach leverages the extensive training of CLIP on billions of images, enabling it to capture the semantic meaning of inputs through its text-image self-supervised learning.

To further explore the overfitting problem, we fine-tune a pre-trained CLIP model on the DRCT-2M/SD-v1.5 dataset (comprising over 300,000 real images from MSCOCO and fake images generated by SD-v1.5) using the OpenCLIP~\cite{openclip} training pipeline. We then train a synthetic image detector on these fine-tuned features and compare its performance with the original CLIP model. The results, shown in \autoref{fig:design_feat_inter}, reveal that fine-tuning generally degrades performance on unseen models due to overfitting. However, performance on SD-v2 improves, likely due to its similarity to SD-v1.5.
This finding suggests that fine-tuning CLIP can be risky. Instead, we propose a better data augmentation during training can help. For example, DRCT~\cite{drct} uses a pre-trained diffusion model to reconstruct real images as hard synthetic samples. However, this approach introduces significant computational overhead and may unfairly inflate the dataset size.
Our solution using feature interpolating, introduced in \autoref{sec:design_semantic}, improves accuracy by approximately 3\% over the original UnivFD (presented in \autoref{fig:design_feat_inter}).

\section{Comparison of Using Different CLIPs as Backbone Models}  \label{app:diff_clip}
\begin{figure}[t]
    \centering
    \begin{minipage}[t]{0.48\textwidth}
        \centering
        \includegraphics[width=1\textwidth]{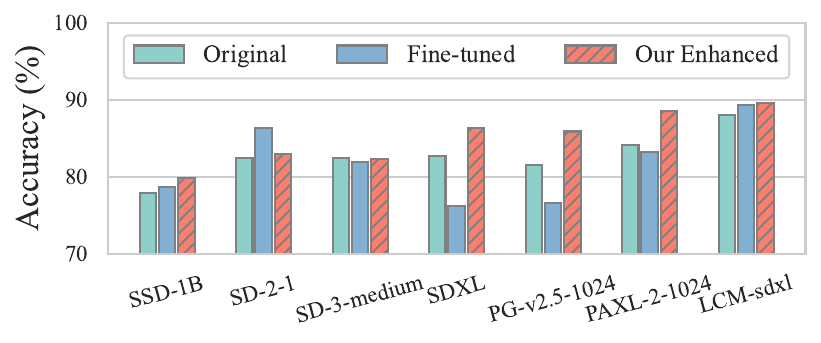}
        \caption{Enhanced training for semantic detectors using CLIP}
        \label{fig:design_feat_inter}
    \end{minipage}
    \hfill
    \begin{minipage}[t]{0.48\textwidth}
        \centering
        \includegraphics[width=1\textwidth]{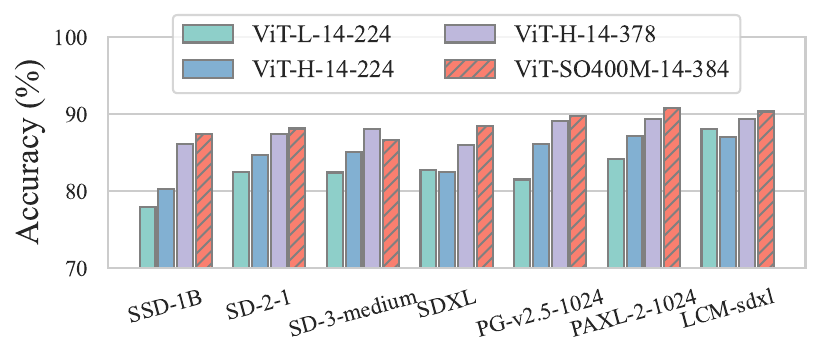}
        \caption{Comparison between latest CLIP models}
        \label{fig:design_clip_model}
    \end{minipage}
\end{figure}

We explore using more advanced CLIP models to better handle the latest generative models. We evaluate three top-performing CLIP models from OpenCLIP~\cite{openclip}: ViT-H-14-224~\cite{ViT-H-14-224}, ViT-H-14-378~\cite{ViT-H-14-378}, and ViT-SO400M-14-384~\cite{ViT-SO400M-14-384}. As shown in \autoref{fig:design_clip_model}, ViT-SO400M-14-384 outperforms the others, potentially due to its use of Sigmoid loss and higher resolution, which enable more effective and robust semantic understanding.
Based on these findings, we propose to use ViT-SO400M-14-384 combined with feature interpolation to enhance the generalization capabilities of semantic detectors, as introduced in \autoref{sec:design_semantic}.

\begin{figure}[t]
    \begin{minipage}[t]{0.24\linewidth}
        \centering
        \includegraphics[width=0.7\linewidth]{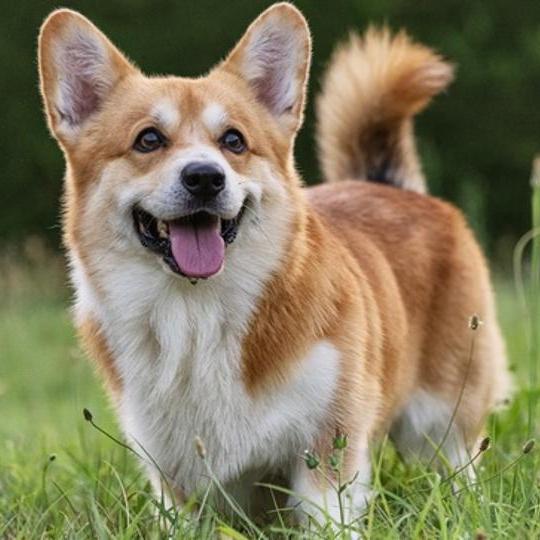}
        \subcaption{Original}
    \end{minipage}
    \hfill
    \begin{minipage}[t]{0.24\linewidth}
        \centering
        \includegraphics[width=0.7\linewidth]{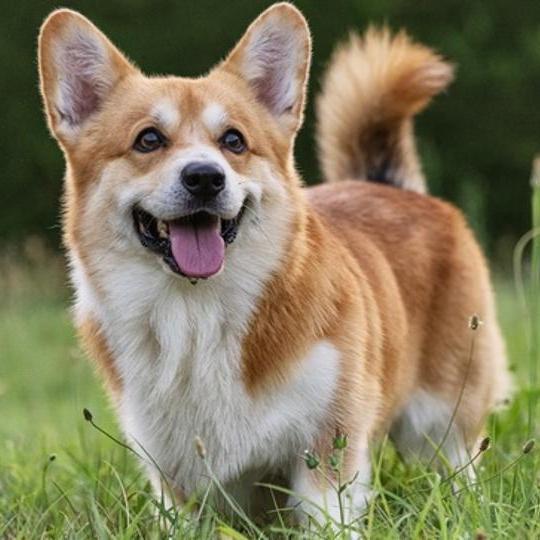}
        \subcaption{JPEG (75)}
    \end{minipage}
    \hfill
    \begin{minipage}[t]{0.24\linewidth}
        \centering
        \includegraphics[width=0.7\linewidth]{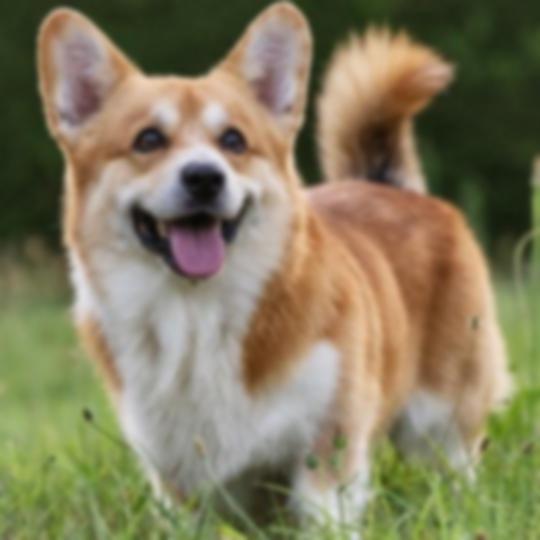}
        \subcaption{Blur (2.5)}
    \end{minipage}
    \hfill
    \begin{minipage}[t]{0.24\linewidth}
        \centering
        \includegraphics[width=0.7\linewidth]{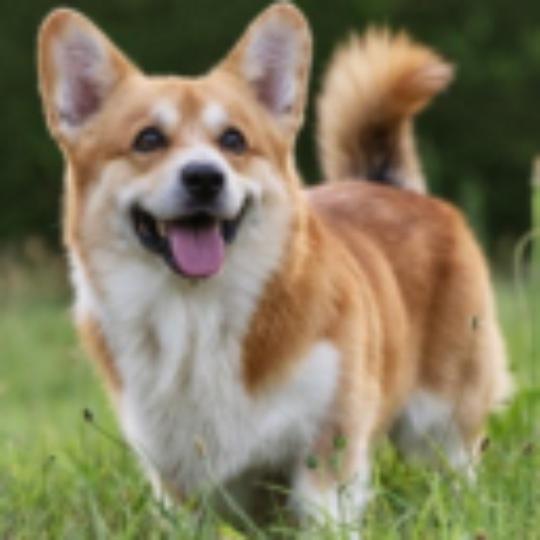}
        \subcaption{Resize (128)}
    \end{minipage}
    \hfill
    \begin{minipage}[t]{0.24\linewidth}
        \centering
        \includegraphics[width=0.7\linewidth]{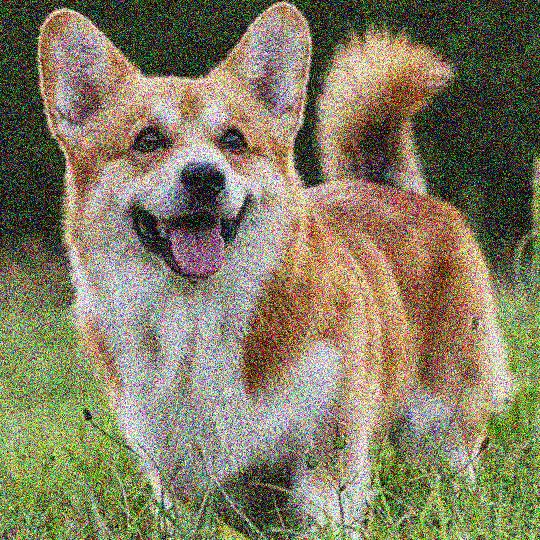}
        \subcaption{Noise (0.25)}
    \end{minipage}
    \hfill
    \begin{minipage}[t]{0.24\linewidth}
        \centering
        \includegraphics[width=0.7\linewidth]{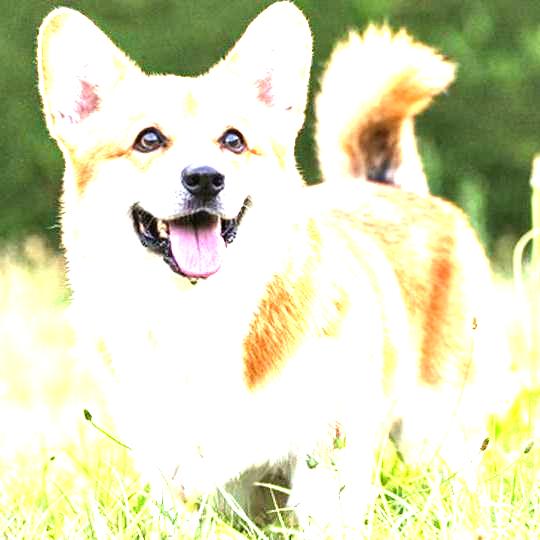}
        \subcaption{Brightness (2.5)}
    \end{minipage}
    \hfill
    \begin{minipage}[t]{0.24\linewidth}
        \centering
        \includegraphics[width=0.7\linewidth]{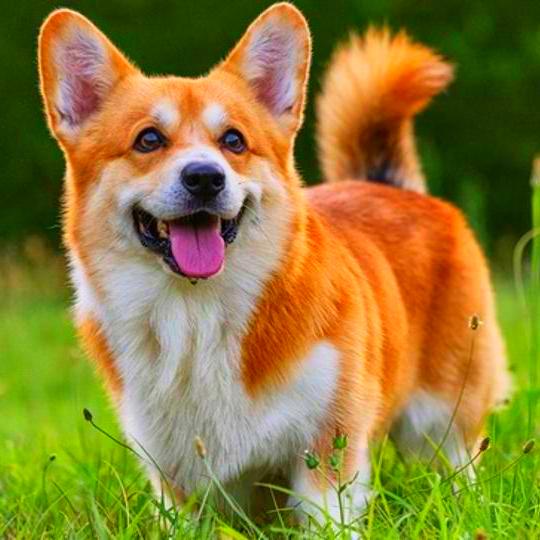}
        \subcaption{Saturation (2.5)}
    \end{minipage}
    \hfill
    \begin{minipage}[t]{0.24\linewidth}
        \centering
        \includegraphics[width=0.7\linewidth]{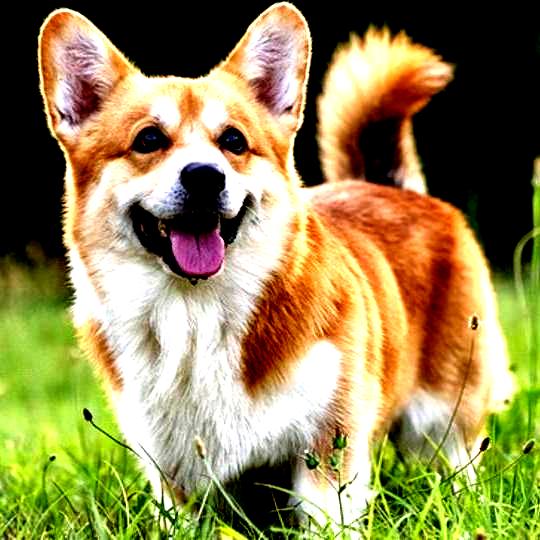}
        \subcaption{Contrast (2.5)}
    \end{minipage}
    \caption{\textbf{Demonstration of various post-processing functions.} Sub-figure (a) shows the original image and the subsequent figures (b)-(h) illustrate the effect of different functions, with the parameter value presented in the parentheses.}
    \label{fig:eval_demo_robust}
\end{figure}

\begin{table}[h]
    \begin{minipage}[c]{0.45\linewidth}
        \centering
        \scriptsize
        \caption{\textbf{Comparison with AIDE~\cite{aide}.} \ours{} outperforms AIDE on Chemeleon dataset and \ourtest{}, and demonstrates greater resilience to lossy formats, e.g., JPEG.}
        \label{tab:reb_baseline}
        \begin{tabular}{lgcgcgc}
        \toprule
        \multirow{2.5}{*}{\textbf{Acc.}} & \multicolumn{2}{c}{AIGCDetect} & \multicolumn{2}{c}{Chameleon} & \multicolumn{2}{c}{\ours} \\
        \cmidrule(lr){2-3} \cmidrule(lr){4-5} \cmidrule(lr){6-7}
        ~ & \cellcolor{white}{Raw} & JPEG & \cellcolor{white}{Raw} & JPEG & \cellcolor{white}{Raw} & JPEG \\
        \midrule
        AIDE  & 92.77 & 73.08 & 61.93 & 55.24 & 85.15 & 74.61 \\
        \ours & 87.75 & 79.76 & 67.63 & 63.19 & 91.45 & 87.06 \\
        \bottomrule
        \end{tabular}
    \end{minipage}
    \hfill
    \begin{minipage}[c]{0.45\linewidth}
        \centering
        \scriptsize
        \caption{\textbf{Comparison of different backbones.} Empirical results show that CLIP achieves the best semantic feature extraction among the evaluated backbone models.}
        \label{tab:reb_backbone}
        \begin{tabular}{lcccc}
        \toprule
        \textbf{Acc.} & CLIP-ViT & ResNet-50 & ConvNeXT & EVA02 \\
        \midrule
        SD-v1.5      & 90.91 & 79.66 & 75.05 & 71.95 \\
        PG-v2.5-1024 & 93.05 & 71.32 & 72.11 & 70.65 \\
        PAXL-2-1024  & 92.37 & 81.95 & 74.85 & 74.73 \\
        \bottomrule
        \end{tabular}
    \end{minipage}
\end{table}

\section{Comparison with Another SOTA Baseline AIDE} \label{app:aide}
We compare \ours{} with AIDE~\cite{aide} by training on the same DRCT/SD-v1.5. We then evaluate on AIGCDetect~\cite{aigcdetect}, Chameleon~\cite{aide}, and \ourtest{} (also testing with JPEG compression). Observe in \autoref{tab:reb_baseline} that while AIDE slightly outperforms \ours{} on AIGCDetect w/o JPEG, it performs worse on the others, especially w/ JPEG. We attribute this to AIDE’s reliance on pixel-level artifacts, which is vulnerable to lossy formats. By contrast, \ours{} fuses enhanced artifact and semantic features, being more robust and generalized.

\section{Backbone Selection: Why Choose CLIP?} \label{app:why_clip}
In the default setting of \ours{}, we use a pre-trained CLIP~\cite{openclip} as the backbone model to extract semantic features. In the study, we evaluate various pre-trained backbones, i.e., CLIP-ViT (default choice), ResNet-50~\cite{resnet}, ConvNeXT~\cite{convnet}, EVA02~\cite{eva02}, by training on DRCT/SD-v1.5 dataset.
The result are shown in ~\autoref{tab:reb_backbone}, CLIP achieves the best performance, due to its large-scale vision-language pretraining. Hence, we choose CLIP-ViT for semantic feature extraction.

\begin{table}[h]
    \centering
    \scriptsize
    \caption{\textbf{Evaluation on Pixel-Space Diffusion Models.} The performance of \ours{}, particularly its artifact detector, slightly degrades on pixel-space diffusion models since they do not use a VAE-based architecture. However, due to their lower generation quality, \ours{} can still effectively detect them based on semantic features.}
    \label{tab:rep_pixel_diffusion}
    \begin{tabular}{lgcgcgc}
    \toprule
    \multirow{2.5}{*}{\textbf{Method}} & \multicolumn{2}{c}{\textbf{Semantic}} & \multicolumn{2}{c}{\textbf{Artifact}} & \multicolumn{2}{c}{\ours} \\
    \cmidrule(lr){2-3} \cmidrule(lr){4-5} \cmidrule(lr){6-7}
    ~ & \cellcolor{white}{AP} & Acc. & \cellcolor{white}{AP} & Acc. & \cellcolor{white}{AP} & Acc. \\
    \midrule
ADM   & 83.96 & 80.20 & 74.31 & 69.96 & 84.48 & 80.30 \\
iDDPM & 84.07 & 81.30 & 75.13 & 70.10 & 87.34 & 84.10 \\
PNDM  & 82.90 & 80.65 & 77.94 & 71.90 & 86.20 & 82.70 \\
    \bottomrule
    \end{tabular}
\end{table}

\section{Evaluation on Pixel-space Diffusion Models} \label{app:pixel_diffusion}
In the main experiment, we primarily focus on evaluating latent diffusion models~\cite{stablediffusion}. In this study, we evaluate 3 pixel-based diffusion models trained on LSUN-bedroom (1,000 synthetic images with 1,000 real ones).
Observe from the \autoref{tab:rep_pixel_diffusion} that artifact-based detection shows a lower accuracy than stable diffusion (over 80\% accuracy), as pixel-space diffusion do not rely on a latent decoding stage (see \autoref{fig:arch_vae}) and thus exhibit fewer up-sampling artifacts. Consequently, the VAE-based artifact detector is less effective (around 70\%). However, pixel-space diffusion are largely outdated and produce lower-quality outputs, enabling \ours{} to detect their semantic inconsistencies (about 85\%).

\section{Detection Performance using CNNDet Training Set} \label{app:progan}
We conduct experiments on the CNNDet training set and evaluate the performance of the converged detectors on the AIGCDetectBenchmark, which comprises synthetic images generated by 16 different generative models. To simulate real-world scenarios, we assess the detectors on images subjected to random JPEG compression. The results are presented in \autoref{tab:robust_progan}, where \ours{} achieves an average accuracy improvement of 8\% over the best baseline, UnivFD.
Although the performance of \ours{} on GAN-based images is slightly lower than that of UnivFD, likely due to the impact of JPEG compression on the artifact detector, \ours{} demonstrates superior performance on the more challenging task of detecting diffusion-generated fake images. This improved generalization is attributed to the adaptive fusion mechanism in \ours{}, which dynamically integrates both semantic and artifact features, enabling more comprehensive decision-making.

In addition, we do not apply any transformation to the input images during training and assess the detection performance on the raw inputs (same as the evaluation setup in most baselines~\cite{npr,lnp}). The results are shown in \autoref{tab:raw_progan}, where each raw shows the test result on different generative models and the last raw presents the averaged result. Observe that \ours{} achieves the best performance with 96.72\% AP and 87.75\% accuracy in average, outperforming the best AP (87.13\% of UnivFD) for over 9\% and the best accuracy (80.69\% of NPR) for over 7\%.
This can be attributed to the enhanced artifact and semantic feature extraction in \ours{} and its comprehensive decision based on both features.
Notably, \ours{} achieves slightly lower but comparable high performance on GAN-generated synthetic images. This is because the baseline detectors tend to overfit on the training data, whose synthetic samples are also generated by GANs. Therefore, they may focus on low-level features typically spread on GAN-generated images but not generalizing to others.
On the other hand, \ours{} makes comprehensive decisions, and hence it generalizes to diffusion-generated images.

\begin{table*}[t]
    \centering
    \scriptsize
    \tabcolsep=3.3pt
    \caption{\textbf{Comparison with existing baselines, trained on CNNDet~\citep{cnndet} and evaluated on AIGCDetectBenchmark~\citep{aigcdetect}.} Note that all images undergo random JPEG compression to simulate real-world scenarios. The results are measured in average precision (AP) and accuracy, with a decision threshold of 0.5. The highest AP scores are highlighted in \textcolor{myred}{red}, and the highest accuracy scores are highlighted in \textcolor{myblue}{blue}. Note that only \ours's results are highlighted if they match the best performance achieved by the baselines.}
    \label{tab:robust_progan}
    \begin{tabular}{lgcgcgcgcgcgcgcgcgcgc}
    \toprule
    \multirow{2}{*}{\textbf{Detector}} & \multicolumn{2}{c}{\textbf{CNNDet}} & \multicolumn{2}{c}{\textbf{FreqFD}} & \multicolumn{2}{c}{\textbf{Fusing}} & \multicolumn{2}{c}{\textbf{LNP}} & \multicolumn{2}{c}{\textbf{LGrad}} & \multicolumn{2}{c}{\textbf{UnivFD}} & \multicolumn{2}{c}{\textbf{DIRE}} & \multicolumn{2}{c}{\textbf{FreqNet}} & \multicolumn{2}{c}{\textbf{NPR}} & \multicolumn{2}{c}{\ours} \\
    \cmidrule(lr){2-3} \cmidrule(lr){4-5} \cmidrule(lr){6-7} \cmidrule(lr){8-9} \cmidrule(lr){10-11} \cmidrule(lr){12-13} \cmidrule(lr){14-15} \cmidrule(lr){16-17} \cmidrule(lr){18-19} \cmidrule(lr){20-21}
    ~ & \cellcolor{white}{AP} & Acc. & \cellcolor{white}{AP} & Acc. & \cellcolor{white}{AP} & Acc. & \cellcolor{white}{AP} & Acc. & \cellcolor{white}{AP} & Acc. & \cellcolor{white}{AP} & Acc. & \cellcolor{white}{AP} & Acc. & \cellcolor{white}{AP} & Acc. & \cellcolor{white}{AP} & Acc. & \cellcolor{white}{AP} & Acc. \\
    \midrule
ProGAN & \textcolor{myred}{100.0} & \textcolor{myblue}{100.0} & 88.50 & 75.48 & 100.0 & 99.98 & 91.04 & 80.95 & 80.19 & 68.84 & 99.95 & 99.01 & 90.54 & 85.80 & 87.19 & 74.95 & 79.81 & 75.11 & 98.45 & 98.99 \\
StyleGAN & \textcolor{myred}{98.79} & 67.68 & 81.06 & 68.93 & 98.74 & 76.84 & 87.35 & 77.10 & 79.98 & 64.11 & 95.71 & 74.79 & 85.18 & 71.90 & 83.04 & 70.26 & 77.22 & 73.85 & 92.32 & \textcolor{myblue}{81.38} \\
BigGAN & 90.01 & 58.23 & 62.32 & 59.48 & 94.88 & 73.88 & 82.18 & 72.47 & 67.20 & 63.25 & \textcolor{myred}{96.80} & \textcolor{myblue}{86.67} & 74.52 & 65.10 & 76.31 & 69.55 & 70.66 & 67.50 & 95.24 & 84.25 \\
CycleGAN & 97.75 & 81.64 & 73.83 & 64.38 & 98.27 & 88.68 & 88.73 & 79.03 & 79.74 & 71.65 & \textcolor{myred}{98.91} & \textcolor{myblue}{93.68} & 71.50 & 63.50 & 85.30 & 72.86 & 78.80 & 71.99 & 91.05 & 90.58 \\
StarGAN & 96.86 & 82.07 & 86.28 & 74.11 & 98.49 & 88.74 & 91.13 & 78.94 & 82.13 & 71.89 & \textcolor{myred}{98.79} & \textcolor{myblue}{94.67} & 94.42 & 82.00 & 85.69 & 68.46 & 74.74 & 74.29 & 94.65 & 86.39 \\
GauGAN & 98.94 & 79.84 & 64.72 & 59.18 & 98.75 & 83.83 & 67.39 & 62.94 & 67.12 & 61.21 & \textcolor{myred}{99.74} & \textcolor{myblue}{97.50} & 80.90 & 72.90 & 78.01 & 71.87 & 66.24 & 65.18 & 96.66 & 83.86 \\
StyleGAN-2 & \textcolor{myred}{98.33} & 63.75 & 82.94 & 66.06 & 97.80 & 70.26 & 84.40 & 74.26 & 74.15 & 61.09 & 95.15 & 66.24 & 78.73 & 72.80 & 84.49 & 69.15 & 77.84 & 74.92 & 90.67 & \textcolor{myblue}{82.45} \\
WFIR & 91.04 & 55.30 & 45.09 & 46.85 & \textcolor{myred}{94.02} & \textcolor{myblue}{77.80} & 78.56 & 68.25 & 65.87 & 58.30 & 93.41 & 70.75 & 62.70 & 60.40 & 50.33 & 48.55 & 54.70 & 51.45 & 82.34 & 73.05 \\
ADM & 64.23 & 50.54 & 58.53 & 58.31 & 60.12 & 51.17 & 82.19 & 72.81 & 50.25 & 51.72 & \textcolor{myred}{88.50} & 64.74 & 70.00 & 64.80 & 77.29 & 67.30 & 72.16 & 67.28 & 87.54 & \textcolor{myblue}{77.66} \\
Glide & 71.26 & 51.53 & 64.96 & 59.28 & 62.39 & 51.82 & \textcolor{myred}{87.51} & \textcolor{myblue}{77.57} & 61.58 & 59.07 & 87.43 & 62.04 & 57.52 & 57.50 & 72.84 & 66.32 & 76.42 & 71.59 & 79.31 & 70.64 \\
Midjourney & 53.76 & 50.51 & 61.07 & 59.70 & 50.81 & 50.62 & 74.20 & 66.54 & 63.57 & 59.31 & 49.05 & 49.83 & 54.62 & 51.30 & 74.39 & 60.85 & 69.01 & 64.19 & \textcolor{myred}{85.29} & \textcolor{myblue}{67.70} \\
SD-v1.4 & 55.62 & 50.07 & 56.19 & 56.07 & 53.09 & 50.08 & 76.69 & 67.49 & 62.13 & 60.46 & 66.43 & 51.23 & 52.66 & 50.90 & 65.30 & 58.25 & 76.25 & 70.95 & \textcolor{myred}{78.96} & \textcolor{myblue}{74.17} \\
SD-v1.5 & 55.22 & 50.06 & 55.54 & 55.17 & 52.46 & 50.12 & 75.59 & 67.00 & 60.96 & 59.30 & 65.95 & 51.23 & 53.17 & 52.00 & 65.20 & 57.88 & 76.41 & 71.14 & \textcolor{myred}{78.62} & \textcolor{myblue}{76.96} \\
VQDM & 73.28 & 51.40 & 62.45 & 58.84 & 73.03 & 53.32 & 73.42 & 64.98 & 50.46 & 52.04 & \textcolor{myred}{95.95} & 80.23 & 65.87 & 58.70 & 73.43 & 66.12 & 74.08 & 69.58 & 90.59 & \textcolor{myblue}{80.42} \\
wukong & 52.65 & 50.08 & 59.17 & 59.47 & 52.79 & 50.21 & 74.13 & 64.17 & 65.87 & 62.46 & 76.35 & 54.20 & 51.86 & 51.00 & 61.35 & 54.73 & 75.74 & 69.31 & \textcolor{myred}{77.38} & \textcolor{myblue}{74.67} \\
DALL-E 2 & 47.16 & 49.95 & 45.87 & 47.00 & 37.03 & 49.65 & \textcolor{myred}{80.95} & \textcolor{myblue}{74.15} & 57.12 & 53.90 & 64.90 & 50.60 & 52.85 & 50.50 & 55.45 & 53.75 & 76.35 & 74.10 & 72.92 & 73.00 \\
\midrule
\textbf{Average} & 77.81 & 62.04 & 65.53 & 60.52 & 76.42 & 66.69 & 80.97 & 71.79 & 66.77 & 61.16 & 85.81 & 71.71 & 68.56 & 63.19 & 73.47 & 64.43 & 73.53 & 69.53 & \textcolor{myred}{87.00} & \textcolor{myblue}{79.76} \\  
    \bottomrule
    \end{tabular}
\end{table*}

\begin{table}[t]
    \centering
    \scriptsize
    \tabcolsep=3.3pt
    \caption{\textbf{Comparison with existing baselines, trained on CNNDet~\citep{cnndet} and evaluated on AIGCDetectBenchmark~\citep{aigcdetect}.} Note that no post-processing is applied to the inputs. The results are measured in average precision (AP) and accuracy, with a decision threshold of 0.5. The highest AP scores are highlighted in \textcolor{myred}{red}, and the highest accuracy scores are highlighted in \textcolor{myblue}{blue}. Note that only \ours's results are highlighted if they match the best performance achieved by the baselines.}
    \label{tab:raw_progan}
    \begin{tabular}{lgcgcgcgcgcgcgcgcgcgc}
    \toprule
    \multirow{2}{*}{\textbf{Detector}} & \multicolumn{2}{c}{\textbf{CNNDet}} & \multicolumn{2}{c}{\textbf{FreqFD}} & \multicolumn{2}{c}{\textbf{Fusing}} & \multicolumn{2}{c}{\textbf{LNP}} & \multicolumn{2}{c}{\textbf{LGrad}} & \multicolumn{2}{c}{\textbf{UnivFD}} & \multicolumn{2}{c}{\textbf{DIRE-G}} & \multicolumn{2}{c}{\textbf{FreqNet}} & \multicolumn{2}{c}{\textbf{NPR}} & \multicolumn{2}{c}{\ours} \\
    \cmidrule(lr){2-3} \cmidrule(lr){4-5} \cmidrule(lr){6-7} \cmidrule(lr){8-9} \cmidrule(lr){10-11} \cmidrule(lr){12-13} \cmidrule(lr){14-15} \cmidrule(lr){16-17} \cmidrule(lr){18-19} \cmidrule(lr){20-21}
    ~ & \cellcolor{white}{AP} & Acc. & \cellcolor{white}{AP} & Acc. & \cellcolor{white}{AP} & Acc. & \cellcolor{white}{AP} & Acc. & \cellcolor{white}{AP} & Acc. & \cellcolor{white}{AP} & Acc. & \cellcolor{white}{AP} & Acc. & \cellcolor{white}{AP} & Acc. & \cellcolor{white}{AP} & Acc. & \cellcolor{white}{AP} & Acc. \\
    \midrule
ProGAN & 100.0 & 100.0 & 100.0 & 99.86 & 100.0 & \textcolor{myblue}{100.0} & 99.75 & 97.31 & 99.32 & 87.78 & 100.0 & 99.81 & 91.54 & 91.80 & 100.0 & 99.58 & 99.95 & 99.84 & \textcolor{myred}{100.0} & 99.86 \\
StyleGAN & 99.19 & 72.61 & 95.81 & 86.56 & 99.26 & 82.92 & 98.55 & 92.31 & 95.53 & 77.93 & 97.48 & 80.40 & 85.18 & 71.90 & 99.78 & 89.91 & 99.74 & \textcolor{myblue}{97.52} & \textcolor{myred}{99.94} & 96.29 \\
BigGAN & 90.39 & 59.45 & 70.54 & 69.77 & 95.65 & 78.47 & 94.51 & 84.95 & 80.01 & 74.85 & 99.27 & \textcolor{myblue}{95.08} & 74.52 & 69.10 & 96.05 & 90.45 & 84.39 & 83.20 & \textcolor{myred}{99.52} & 92.00 \\
CycleGAN & 97.92 & 84.63 & 88.06 & 70.82 & 98.47 & 91.11 & 97.09 & 86.00 & 96.66 & 90.12 & \textcolor{myred}{99.80} & \textcolor{myblue}{98.33} & 71.50 & 66.80 & 99.63 & 95.84 & 97.83 & 94.10 & 99.33 & 98.03 \\
StarGAN & 97.51 & 84.74 & 100.0 & 96.87 & 99.05 & 91.40 & 99.94 & 85.12 & 99.00 & 94.15 & 99.37 & 95.75 & 94.42 & 88.50 & 99.80 & 85.67 & 100.0 & \textcolor{myblue}{99.70} & \textcolor{myred}{100.0} & 96.05 \\
GauGAN & 98.77 & 82.86 & 74.42 & 65.69 & 98.60 & 86.27 & 76.51 & 71.74 & 83.45 & 72.86 & \textcolor{myred}{99.98} & \textcolor{myblue}{99.47} & 80.90 & 72.90 & 98.63 & 93.41 & 81.73 & 79.97 & 99.95 & 90.90 \\
StyleGAN-2 & 99.03 & 69.22 & 95.59 & 80.17 & 98.84 & 78.97 & 98.98 & 94.14 & 90.85 & 72.25 & 97.71 & 70.76 & 78.73 & 72.80 & 99.58 & 87.89 & \textcolor{myred}{99.97} & \textcolor{myblue}{99.34} & 99.94 & 97.89 \\
WFIR & 91.27 & 56.60 & 43.54 & 45.30 & \textcolor{myred}{95.07} & \textcolor{myblue}{81.95} & 74.03 & 61.80 & 70.26 & 57.30 & 94.22 & 72.70 & 62.70 & 60.40 & 51.06 & 49.20 & 61.55 & 59.75 & 92.12 & 71.65 \\
ADM & 64.70 & 51.04 & 60.30 & 61.82 & 60.26 & 51.68 & 80.78 & 71.94 & 51.92 & 55.18 & 89.80 & 67.46 & 70.00 & 64.80 & 92.13 & \textcolor{myblue}{84.06} & 73.22 & 68.95 & \textcolor{myred}{95.31} & 73.28 \\
Glide & 71.61 & 52.78 & 67.69 & 58.34 & 60.45 & 52.85 & 72.21 & 62.29 & 67.69 & 68.64 & 88.04 & 63.09 & 57.52 & 57.50 & 89.78 & 82.78 & 81.01 & 75.51 & \textcolor{myred}{98.87} & \textcolor{myblue}{88.82} \\
Midjourney & 53.45 & 50.60 & 48.71 & 46.93 & 48.78 & 50.79 & 79.54 & 70.12 & 62.77 & 58.83 & 49.72 & 49.87 & 54.62 & 53.10 & 80.88 & 71.02 & 80.33 & 74.57 & \textcolor{myred}{89.78} & \textcolor{myblue}{88.70} \\
SD-v1.4 & 55.77 & 50.14 & 43.55 & 45.01 & 52.27 & 50.13 & 63.97 & 59.29 & 66.47 & 67.24 & 68.63 & 51.70 & 52.66 & 52.40 & 77.10 & 65.56 & 80.44 & 75.58 & \textcolor{myred}{93.30} & \textcolor{myblue}{86.01} \\
SD-v1.5 & 55.68 & 50.07 & 43.09 & 44.27 & 51.99 & 50.07 & 64.16 & 59.26 & 65.91 & 66.40 & 68.07 & 51.59 & 53.17 & 53.00 & 77.95 & 65.84 & 81.23 & 76.36 & \textcolor{myred}{93.20} & \textcolor{myblue}{86.35} \\
VQDM & 72.62 & 52.15 & 69.45 & 65.34 & 71.55 & 53.93 & 66.82 & 62.70 & 54.30 & 56.92 & 97.53 & \textcolor{myblue}{86.01} & 65.87 & 58.70 & 90.42 & 82.29 & 74.66 & 72.98 & \textcolor{myred}{97.73} & 82.35 \\
wukong & 52.71 & 50.08 & 46.96 & 48.48 & 51.53 & 50.13 & 61.99 & 56.75 & 69.84 & 68.41 & 78.44 & 55.14 & 51.86 & 48.70 & 69.43 & 58.59 & 75.42 & 72.23 & \textcolor{myred}{92.34} & \textcolor{myblue}{78.77} \\
DALL-E 2 & 47.16 & 49.85 & 39.58 & 36.00 & 37.89 & 49.50 & 87.88 & 76.25 & 64.21 & 57.45 & 66.06 & 50.80 & 52.85 & 51.40 & 55.40 & 55.75 & 70.86 & 61.40 & \textcolor{myred}{96.23} & \textcolor{myblue}{77.05} \\
\midrule
\textbf{Average} & 77.99 & 63.55 & 67.96 & 63.83 & 76.23 & 68.76 & 82.29 & 74.50 & 76.14 & 70.39 & 87.13 & 74.25 & 68.63 & 64.61 & 86.10 & 78.61 & 83.90 & 80.69 & \textcolor{myred}{96.72} & \textcolor{myblue}{87.75} \\
    \bottomrule
    \end{tabular}
\end{table}

\section{Detection Performance using DRCT Training Set over GenImage Test Set} \label{app:eval_genimage}
The test results on GenImage~\cite{genimage} are presented in \autoref{tab:eval_genimages}, where \ours{} achieves slightly better performance compared to the latest detector, DRCT. The reason is that DRCT leverages SD-v1.4 to reconstruct real images, thereby creating challenging synthetic samples and effectively increasing the amount of training data. Despite this advantage, \ours{} still outperforms DRCT due to its comprehensive decision-making approach.

\begin{table}[t]
    \centering
    \scriptsize
    \tabcolsep=3.3pt
    \caption{\textbf{Comparison with existing baselines, trained on DRCT~\citep{drct} and evaluated on GenImage dataset~\citep{genimage}.} The results are measured in average precision (AP) and accuracy, with a decision threshold of 0.5. The highest AP scores are highlighted in \textcolor{myred}{red}, and the highest accuracy scores are highlighted in \textcolor{myblue}{blue}. Note that only \ours's results are highlighted if they match the best performance achieved by the baselines.}
    \label{tab:eval_genimages}
    \begin{tabular}{lgcgcgcgcgcgcgcgcgcgc}
    \toprule
    \multirow{2}{*}{\textbf{Detector}} & \multicolumn{2}{c}{\textbf{CNNDet}} & \multicolumn{2}{c}{\textbf{FreqFD}} & \multicolumn{2}{c}{\textbf{Fusing}} & \multicolumn{2}{c}{\textbf{LNP}} & \multicolumn{2}{c}{\textbf{UnivFD}} & \multicolumn{2}{c}{\textbf{DIRE}} & \multicolumn{2}{c}{\textbf{FreqNet}} & \multicolumn{2}{c}{\textbf{NPR}} & \multicolumn{2}{c}{\textbf{DRCT}} & \multicolumn{2}{c}{\ours} \\
    \cmidrule(lr){2-3} \cmidrule(lr){4-5} \cmidrule(lr){6-7} \cmidrule(lr){8-9} \cmidrule(lr){10-11} \cmidrule(lr){12-13} \cmidrule(lr){14-15} \cmidrule(lr){16-17} \cmidrule(lr){18-19} \cmidrule(lr){20-21}
    ~ & \cellcolor{white}{AP} & Acc. & \cellcolor{white}{AP} & Acc. & \cellcolor{white}{AP} & Acc. & \cellcolor{white}{AP} & Acc. & \cellcolor{white}{AP} & Acc. & \cellcolor{white}{AP} & Acc. & \cellcolor{white}{AP} & Acc. & \cellcolor{white}{AP} & Acc. & \cellcolor{white}{AP} & Acc. & \cellcolor{white}{AP} & Acc. \\
    \midrule
ADM & 47.93 & 50.12 & 48.68 & 50.14 & 55.17 & 50.43 & \textcolor{myred}{86.69} & 57.39 & 58.97 & 53.77 & 68.60 & 53.95 & 78.21 & 58.27 & 75.00 & 59.30 & 81.74 & \textcolor{myblue}{76.81} & 81.63 & 67.25 \\
Glide & 72.70 & 51.98 & 72.85 & 52.13 & 77.29 & 51.43 & 96.33 & 74.48 & 81.92 & 69.64 & 85.61 & 63.49 & 92.47 & 68.42 & \textcolor{myred}{99.12} & 81.12 & 92.83 & 86.60 & 95.90 & \textcolor{myblue}{93.02} \\
Midjourney & 73.53 & 53.12 & 65.11 & 50.52 & 77.94 & 52.02 & 74.55 & 54.33 & 88.35 & 78.08 & 63.94 & 53.23 & 67.55 & 51.22 & 88.50 & 58.27 & 89.38 & 82.39 & \textcolor{myred}{92.26} & \textcolor{myblue}{83.45} \\
SD-v1.4 & 99.91 & 98.28 & 94.27 & 64.83 & \textcolor{myred}{99.98} & \textcolor{myblue}{99.34} & 99.88 & 98.20 & 96.58 & 90.13 & 97.51 & 86.48 & 94.18 & 68.83 & 99.54 & 93.30 & 94.40 & 88.45 & 96.92 & 96.83 \\
SD-v1.5 & 99.87 & 98.17 & 93.85 & 64.61 & \textcolor{myred}{99.93} & \textcolor{myblue}{99.28} & 99.84 & 97.95 & 96.43 & 89.94 & 97.53 & 86.56 & 94.39 & 68.89 & 99.55 & 93.23 & 94.42 & 88.44 & 96.95 & 96.68 \\
VQDM & 53.22 & 51.08 & 61.33 & 51.58 & 61.18 & 50.75 & 88.51 & 58.46 & 65.56 & 56.14 & 64.55 & 52.48 & 75.60 & 57.25 & 66.59 & 53.40 & \textcolor{myred}{90.89} & \textcolor{myblue}{84.07} & 90.57 & 78.83 \\
wukong & 99.76 & 96.14 & 92.10 & 61.82 & \textcolor{myred}{99.92} & \textcolor{myblue}{97.66} & 99.60 & 95.51 & 95.13 & 87.38 & 95.66 & 80.34 & 91.64 & 63.62 & 98.51 & 82.68 & 93.99 & 87.75 & 96.72 & 95.93 \\
BigGAN & 41.61 & 49.50 & 70.80 & 56.17 & 45.01 & 49.78 & 42.67 & 45.15 & 67.07 & 57.25 & 44.76 & 48.62 & 40.56 & 44.20 & 35.18 & 39.45 & \textcolor{myred}{74.58} & \textcolor{myblue}{74.10} & 65.39 & 65.20 \\
\midrule
\textbf{Average} & 73.57 & 68.55 & 74.87 & 56.48 & 77.05 & 68.84 & 86.01 & 72.68 & 81.25 & 72.79 & 77.27 & 65.64 & 79.32 & 60.09 & 82.75 & 70.09 & 89.03 & 83.58 & \textcolor{myred}{89.54} & \textcolor{myblue}{84.65} \\
    \bottomrule
    \end{tabular}
\end{table}

\section{Illustration of Various Post-processing Transformations} \label{app:illu_robust}

We illustrate the effect of various post-processing transformations (evaluated in \autoref{sec:eval_robust}) in \autoref{fig:eval_demo_robust}.

\begin{figure}[t]
    \centering
    \includegraphics[width=1\linewidth]{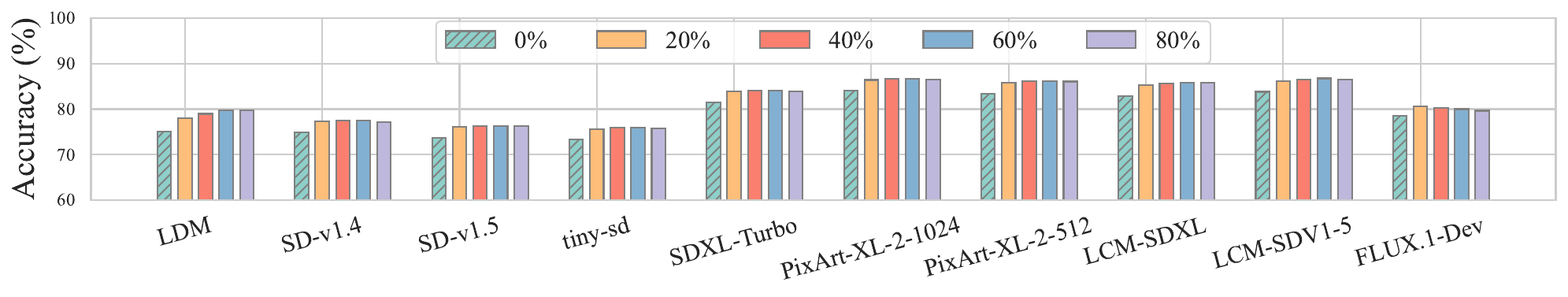}
    \caption{Ablation study on the probability of random feature interpolation}
    \label{fig:abl_feat_inter}
\end{figure}

\section{Ablation Study on the Strength of Feature Interpolation as Data Augmentation} \label{app:ablation_feat_inter}
In this section, we perform an ablation study to evaluate the impact of feature interpolation strength (as introduced in \autoref{sec:design_semantic}) on detection performance. We utilize the DRCT~\cite{drct} training set and \ourtest{} as the test set. We randomly select 0\% (no augmentation), 20\%, 40\%, 60\%, and 80\% of the data in each batch for feature interpolation. The results are presented in \autoref{fig:abl_feat_inter}. Observe that incorporating feature interpolation as an augmentation technique generally enhances \ours{}'s detection performance by approximately 3\%. Specifically, interpolation probabilities of 40\% and 60\% yield the most significant improvements. Consequently, we adopt a default probability of 50\% for performing random feature interpolation.

\begin{table*}[t]
    \centering
    \scriptsize
    \tabcolsep=7pt
    \caption{\textbf{Ablation study on \ours, trained on DRCT~\citep{drct} and evaluated on \ourtest{}.} The results are measured in average precision (AP) and accuracy, with a decision threshold of 0.5. The highest AP scores are highlighted in \textcolor{myred}{red}, and the highest accuracy scores are highlighted in \textcolor{myblue}{blue}. Note that only \ours's results are highlighted if they match the best performance achieved by the baselines.}
    \label{tab:eval_ablation}
    \begin{tabular}{lgcgcgcgcgcgcgc}
    \toprule
    \multirow{2}{*}{\textbf{Method}} & \multicolumn{2}{c}{\textbf{Only Semantic}} & \multicolumn{2}{c}{\textbf{Only Artifact}} & \multicolumn{2}{c}{\textbf{Avg}} & \multicolumn{2}{c}{\textbf{Max}} & \multicolumn{2}{c}{\textbf{Min}} & \multicolumn{2}{c}{\textbf{Simple Concat}} & \multicolumn{2}{c}{\ours} \\
    \cmidrule(lr){2-3} \cmidrule(lr){4-5} \cmidrule(lr){6-7} \cmidrule(lr){8-9} \cmidrule(lr){10-11} \cmidrule(lr){12-13}  \cmidrule(lr){14-15}
    ~ & \cellcolor{white}{AP} & Acc. & \cellcolor{white}{AP} & Acc. & \cellcolor{white}{AP} & Acc. & \cellcolor{white}{AP} & Acc. & \cellcolor{white}{AP} & Acc. & \cellcolor{white}{AP} & Acc. & \cellcolor{white}{AP} & Acc. \\
    \midrule
LDM & 79.86 & 66.07 & 67.78 & 58.19 & 77.50 & 63.00 & 74.76 & 68.16 & 78.52 & 56.10 & 85.25 & 84.51 & \textcolor{myred}{98.91} & \textcolor{myblue}{95.04} \\
SD-v1.4 & 93.46 & 86.90 & 95.10 & 87.62 & 97.16 & \textcolor{myblue}{92.54} & 95.26 & 85.92 & 96.82 & 88.60 & 87.95 & 83.54 & \textcolor{myred}{97.80} & 91.95 \\
SD-v1.5 & 93.55 & 86.85 & 95.21 & 87.78 & 97.27 & \textcolor{myblue}{92.78} & 95.34 & 85.97 & 96.94 & 88.66 & 96.84 & 87.64 & \textcolor{myred}{98.02} & 91.31 \\
tiny-sd & 87.77 & 76.84 & 83.76 & 74.82 & 90.70 & 80.94 & 86.27 & 79.84 & 91.41 & 71.82 & 85.98 & 82.35 & \textcolor{myred}{95.99} & \textcolor{myblue}{84.80} \\
SegMoE-SD & 89.70 & 80.35 & 91.26 & 83.85 & 94.22 & 88.33 & 91.68 & 84.22 & 93.99 & 79.98 & 91.26 & 68.01 & \textcolor{myred}{97.39} & \textcolor{myblue}{89.49} \\
SDXL-turbo & 95.08 & 89.15 & 95.57 & 88.03 & 97.86 & 93.42 & 96.22 & 86.01 & 97.64 & 91.17 & 97.90 & 82.89 & \textcolor{myred}{99.17} & \textcolor{myblue}{95.39} \\
SDXL & 87.79 & \textcolor{myblue}{77.22} & 75.54 & 65.25 & 86.72 & 74.66 & 83.57 & 76.45 & 86.85 & 66.02 & 77.80 & 76.27 & \textcolor{myred}{91.68} & 74.12 \\
PG-v2-512 & \textcolor{myred}{85.24} & \textcolor{myblue}{72.85} & 66.59 & 58.10 & 80.83 & 66.36 & 78.68 & 70.98 & 79.17 & 59.97 & 79.17 & 64.12 & 85.02 & 64.86 \\
PG-v2-256 & 89.08 & 79.20 & 68.30 & 59.05 & 85.21 & 70.05 & 82.89 & 75.15 & 81.35 & 63.10 & 86.83 & \textcolor{myblue}{79.82} & \textcolor{myred}{90.22} & 72.92 \\
PAXL-2-1024 & 97.14 & 92.80 & 81.85 & 72.58 & 94.61 & 90.53 & 95.04 & 85.67 & 92.66 & 79.71 & 89.93 & 87.27 & \textcolor{myred}{97.94} & \textcolor{myblue}{93.94} \\
PAXL-2-512 & 97.31 & 92.93 & 89.98 & 81.97 & 97.06 & 92.83 & 96.10 & 86.13 & 96.26 & 88.77 & 94.49 & \textcolor{myblue}{95.27} & \textcolor{myred}{98.63} & 94.96 \\
LCM-sdxl & 96.95 & 92.29 & 92.94 & 84.52 & 97.86 & 93.59 & 96.43 & 86.44 & 97.38 & 90.37 & 87.64 & 79.57 & \textcolor{myred}{98.72} & \textcolor{myblue}{96.20} \\
LCM-sdv1-5 & 96.77 & 92.45 & 97.88 & 88.58 & 98.86 & 94.37 & 98.60 & 86.45 & 98.57 & 94.58 & 91.98 & 89.87 & \textcolor{myred}{99.63} & \textcolor{myblue}{97.14} \\
FLUX.1-sch & 93.31 & 84.93 & 84.18 & 75.02 & 92.99 & \textcolor{myblue}{86.24} & 91.70 & 83.50 & 91.89 & 76.45 & 83.27 & 75.49 & \textcolor{myred}{95.52} & 85.24 \\
\midrule
\textbf{Average} & 91.64 & 83.63 & 84.71 & 76.10 & 92.06 & 84.26 & 90.18 & 81.49 & 91.39 & 78.24 & 88.31 & 81.19 & \textcolor{myred}{96.04} & \textcolor{myblue}{87.67} \\
    \bottomrule
    \end{tabular}
\end{table*}

\section{Ablation Study on Feature Fusion} \label{app:eval_ablation}
In this section, we conduct an ablation study of \ours{} to examine the integration of semantic and artifact features. The experiments are performed using the DRCT training set, and the performance is evaluated on \ourtest{}.
The results are presented in \autoref{tab:eval_ablation}, which compare the default \ours{} setting with several alternative and straightforward configurations. These alternatives include
(1) using only semantic features for detection, (2) only artifact features, (3) averaging the semantic and artifact scores from two detectors, (4) taking the maximum score between the semantic and artifact detectors, (5) outputting the minimum score, and (6) simply concatenating the semantic and artifact vectors without an adaptive regulator.
As shown in the table, the default setting of \ours{} outperforms these straightforward combinations in most cases, demonstrating the effectiveness of our design. This superior performance is attributed to the regulators that dynamically assign adaptive coefficients to the semantic and artifact features, allowing the model to handle different test cases effectively. In contrast, simple concatenation leads to overfitting on one feature, resulting in reduced effectiveness.

\begin{table}[t]
    \centering
    \scriptsize
    \tabcolsep=3.3pt
    \caption{\textbf{Comparison with existing baselines, trained on DRCT~\citep{drct} and evaluated on \ourtest{} dataset.} The results are measured in F1 score (\%) and ROC-AUC score (\%). The highest F1 scores are highlighted in \textcolor{myred}{red}, and the highest ROC-AUC scores are highlighted in \textcolor{myblue}{blue}. Note that only \ours's results are highlighted if they match the best performance achieved by the baselines.}
    \label{tab:eval_custom_f1_auc}
    \begin{tabular}{lgcgcgcgcgcgcgcgcgcgc}
    \toprule
    \multirow{2}{*}{\textbf{Detector}} & \multicolumn{2}{c}{\textbf{CNNDet}} & \multicolumn{2}{c}{\textbf{FreqFD}} & \multicolumn{2}{c}{\textbf{Fusing}} & \multicolumn{2}{c}{\textbf{LNP}} & \multicolumn{2}{c}{\textbf{UnivFD}} & \multicolumn{2}{c}{\textbf{DIRE}} & \multicolumn{2}{c}{\textbf{FreqNet}} & \multicolumn{2}{c}{\textbf{NPR}} & \multicolumn{2}{c}{\textbf{DRCT}} & \multicolumn{2}{c}{\ours} \\
    \cmidrule(lr){2-3} \cmidrule(lr){4-5} \cmidrule(lr){6-7} \cmidrule(lr){8-9} \cmidrule(lr){10-11} \cmidrule(lr){12-13} \cmidrule(lr){14-15} \cmidrule(lr){16-17} \cmidrule(lr){18-19} \cmidrule(lr){20-21}
    ~ & \cellcolor{white}{F1} & AUC & \cellcolor{white}{F1} & AUC & \cellcolor{white}{F1} & AUC & \cellcolor{white}{F1} & AUC & \cellcolor{white}{F1} & AUC & \cellcolor{white}{F1} & AUC & \cellcolor{white}{F1} & AUC & \cellcolor{white}{F1} & AUC & \cellcolor{white}{F1} & AUC & \cellcolor{white}{F1} & AUC \\
    \midrule
LDM & 72.53 & 89.65 & 16.38 & 72.46 & 79.64 & 98.16 & 83.01 & 96.08 & 77.97 & 87.97 & 50.57 & 85.53 & 67.85 & 91.89 & 82.42 & 92.05 & 81.81 & 89.29 & \textcolor{myred}{95.01} & \textcolor{myblue}{99.02} \\
SD-v1.4 & 89.32 & 97.56 & 41.64 & 92.67 & \textcolor{myred}{99.16} & \textcolor{myblue}{99.97} & 95.92 & 99.31 & 80.24 & 89.38 & 80.78 & 96.83 & 57.39 & 91.08 & 90.49 & 97.70 & 83.78 & 92.25 & 91.68 & 97.86 \\
SD-v1.5 & 89.08 & 97.52 & 40.76 & 92.58 & \textcolor{myred}{99.11} & \textcolor{myblue}{99.97} & 96.23 & 99.35 & 80.25 & 89.46 & 81.08 & 97.01 & 56.71 & 91.08 & 90.94 & 97.77 & 83.66 & 91.86 & 90.93 & 97.94 \\
SSD-1B & 52.68 & 88.67 & 0.16 & 48.90 & 15.15 & 81.79 & 74.94 & 94.16 & 74.54 & 86.29 & 26.08 & 73.16 & 2.76 & 46.31 & 0.95 & 52.55 & 78.16 & 82.88 & \textcolor{myred}{80.81} & \textcolor{myblue}{95.33} \\
tiny-sd & 52.30 & 89.02 & 9.50 & 82.20 & 70.45 & \textcolor{myblue}{98.02} & 78.38 & 95.35 & 75.20 & 86.34 & 44.66 & 88.48 & 45.32 & 89.03 & \textcolor{myred}{87.57} & 96.92 & 82.57 & 89.19 & 82.98 & 95.87 \\
SegMoE-SD & 67.42 & 92.21 & 7.87 & 83.12 & 64.24 & 97.18 & 85.27 & 96.96 & 82.90 & 90.70 & 49.99 & 89.68 & 47.24 & 89.65 & \textcolor{myred}{93.70} & \textcolor{myblue}{98.15} & 77.37 & 81.40 & 88.77 & 97.46 \\
small-sd & 59.82 & 91.10 & 10.86 & 84.01 & 79.08 & \textcolor{myblue}{99.11} & 77.34 & 95.28 & 75.85 & 87.16 & 55.10 & 91.90 & 49.52 & 90.88 & \textcolor{myred}{88.43} & 97.12 & 83.80 & 91.42 & 84.27 & 96.18 \\
SD-2-1 & 55.92 & 86.84 & 0.99 & 52.98 & 31.77 & 92.62 & 30.27 & 81.47 & 81.31 & 89.75 & 48.35 & 88.55 & 11.55 & 62.35 & 13.16 & 71.39 & 78.48 & 83.09 & \textcolor{myred}{87.67} & \textcolor{myblue}{96.99} \\
SD-3-medium & 39.34 & 79.70 & 1.19 & 60.64 & 9.93 & 81.00 & 20.00 & 75.69 & 77.13 & 87.34 & 26.42 & 76.76 & 4.48 & 55.06 & 8.69 & 72.18 & 77.18 & 81.35 & \textcolor{myred}{80.43} & \textcolor{myblue}{95.25} \\
SDXL-turbo & 87.79 & 95.94 & 37.63 & 93.89 & 32.81 & 95.94 & 81.15 & 95.78 & 84.35 & 91.16 & 63.96 & 89.74 & 52.11 & 88.53 & 81.40 & 96.10 & 82.95 & 91.83 & \textcolor{myred}{95.37} & \textcolor{myblue}{99.07} \\
SD-2 & 50.95 & 87.13 & 0.63 & 50.16 & 21.59 & 88.34 & 22.67 & 77.24 & 70.78 & 83.96 & 35.70 & 84.32 & 8.12 & 56.52 & 12.76 & 73.43 & 77.40 & 81.69 & \textcolor{myred}{81.42} & \textcolor{myblue}{95.32} \\
SDXL & 42.04 & 86.04 & 0.08 & 42.62 & 4.52 & 74.18 & 76.73 & \textcolor{myblue}{94.18} & 60.40 & 73.01 & 10.99 & 64.66 & 1.46 & 44.72 & 0.50 & 45.32 & \textcolor{myred}{77.44} & 81.80 & 67.12 & 91.76 \\
PG-v2.5-1024 & 19.80 & 63.76 & 0.08 & 48.50 & 2.21 & 78.65 & 74.86 & 94.40 & 78.68 & 83.54 & 12.27 & 60.18 & 0.81 & 55.59 & 0.34 & 50.43 & 73.00 & 78.14 & \textcolor{myred}{87.78} & \textcolor{myblue}{96.91} \\
PG-v2-1024 & 45.99 & 84.76 & 0.08 & 50.77 & 8.51 & 86.42 & 16.16 & 75.61 & 79.02 & 84.02 & 26.08 & 77.17 & 1.27 & 54.11 & 2.93 & 64.55 & 67.10 & 71.82 & \textcolor{myred}{88.38} & \textcolor{myblue}{97.13} \\
PG-v2-512 & 32.25 & 79.04 & 0.75 & 57.24 & 6.67 & 69.89 & 5.60 & 57.22 & 52.88 & 70.16 & 16.71 & 72.12 & 2.42 & 38.39 & 5.46 & 64.46 & \textcolor{myred}{80.09} & \textcolor{myblue}{85.27} & 49.40 & 83.82 \\
PG-v2-256 & 45.58 & 81.43 & 2.01 & 59.07 & 4.86 & 73.09 & 23.30 & 70.63 & 59.42 & 73.86 & 36.35 & 80.12 & 3.84 & 41.80 & 7.78 & 56.13 & \textcolor{myred}{75.60} & 79.57 & 64.89 & \textcolor{myblue}{88.92} \\
PAXL-2-1024 & 27.67 & 70.12 & 0.32 & 56.44 & 13.91 & 88.13 & 19.13 & 75.46 & 80.59 & 85.51 & 20.36 & 69.07 & 3.58 & 66.05 & 12.66 & 73.57 & 73.56 & 77.56 & \textcolor{myred}{93.83} & \textcolor{myblue}{98.47} \\
PAXL-2-512 & 50.33 & 82.76 & 9.54 & 82.89 & 54.79 & 96.50 & 67.18 & 92.67 & 80.83 & 85.73 & 41.11 & 79.03 & 29.08 & 84.80 & 78.20 & 95.68 & 77.83 & 82.04 & \textcolor{myred}{94.93} & \textcolor{myblue}{98.90} \\
LCM-sdxl & 78.47 & 91.85 & 10.40 & 82.84 & 58.83 & 98.16 & 83.78 & 95.92 & 78.48 & 82.70 & 55.48 & 88.48 & 42.31 & 86.53 & 11.68 & 70.57 & 83.65 & 92.78 & \textcolor{myred}{96.22} & \textcolor{myblue}{99.11} \\
LCM-sdv1-5 & 92.00 & 97.17 & 53.74 & 94.92 & 76.96 & 98.93 & 90.40 & 97.80 & 80.17 & 84.41 & 74.11 & 92.71 & 69.98 & 93.32 & 93.61 & 98.57 & 82.16 & 88.75 & \textcolor{myred}{97.19} & \textcolor{myblue}{99.67} \\
FLUX.1-sch & 26.95 & 71.29 & 1.34 & 57.68 & 4.60 & 77.11 & 22.12 & 74.25 & 72.70 & 79.85 & 24.99 & 72.22 & 7.22 & 63.70 & 19.07 & 79.39 & 69.38 & 74.26 & \textcolor{myred}{83.53} & \textcolor{myblue}{96.08} \\
FLUX.1-dev & 30.89 & 69.85 & 1.42 & 53.77 & 11.54 & 83.43 & 21.36 & 72.30 & 76.22 & 82.31 & 25.14 & 72.24 & 2.30 & 52.32 & 11.24 & 71.46 & 72.24 & 77.68 & \textcolor{myred}{84.56} & \textcolor{myblue}{96.51} \\
\midrule
\textbf{Average} & 54.96 & 85.16 & 11.24 & 68.20 & 38.65 & 88.94 & 56.63 & 86.69 & 75.45 & 84.30 & 41.19 & 81.36 & 25.79 & 69.71 & 40.64 & 77.98 & 78.15 & 83.91 & \textcolor{myred}{84.87} & \textcolor{myblue}{96.07} \\
    \bottomrule
    \end{tabular}
\end{table}

\begin{table}[t]
    \centering
    \scriptsize
    \tabcolsep=3.3pt
    \caption{\textbf{Comparison with existing baselines, trained on DRCT~\citep{drct} and evaluated on \ourtest{} dataset.} The results are measured in TPR at 10\% FPR and 1\% FPR. The highest T-10 (TPR at 10\% FPR) are highlighted in \textcolor{myred}{red}, and the highest T-1 (TPR at 1\% FPR) are highlighted in \textcolor{myblue}{blue}. Note that only \ours's results are highlighted if they match the best performance achieved by the baselines.}
    \label{tab:eval_custom_tpr_fpr}
    \begin{tabular}{lgcgcgcgcgcgcgcgcgcgc}
    \toprule
    \multirow{2}{*}{\textbf{Detector}} & \multicolumn{2}{c}{\textbf{CNNDet}} & \multicolumn{2}{c}{\textbf{FreqFD}} & \multicolumn{2}{c}{\textbf{Fusing}} & \multicolumn{2}{c}{\textbf{LNP}} & \multicolumn{2}{c}{\textbf{UnivFD}} & \multicolumn{2}{c}{\textbf{DIRE}} & \multicolumn{2}{c}{\textbf{FreqNet}} & \multicolumn{2}{c}{\textbf{NPR}} & \multicolumn{2}{c}{\textbf{DRCT}} & \multicolumn{2}{c}{\ours} \\
    \cmidrule(lr){2-3} \cmidrule(lr){4-5} \cmidrule(lr){6-7} \cmidrule(lr){8-9} \cmidrule(lr){10-11} \cmidrule(lr){12-13} \cmidrule(lr){14-15} \cmidrule(lr){16-17} \cmidrule(lr){18-19} \cmidrule(lr){20-21}
    ~ & \cellcolor{white}{T-10} & T-1 & \cellcolor{white}{T-10} & T-1 & \cellcolor{white}{T-10} & T-1 & \cellcolor{white}{T-10} & T-1 & \cellcolor{white}{T-10} & T-1 & \cellcolor{white}{T-10} & T-1 & \cellcolor{white}{T-10} & T-1 & \cellcolor{white}{T-10} & T-1 & \cellcolor{white}{T-10} & T-1 & \cellcolor{white}{T-10} & T-1 \\
    \midrule
LDM & 75.54 & 41.38 & 37.72 & 13.22 & 95.70 & 78.20 & 89.10 & 49.66 & 75.88 & 21.56 & 64.20 & 25.47 & 74.84 & 33.66 & 84.76 & 38.02 & 65.28 & 19.32 & \textcolor{myred}{98.32} & \textcolor{myblue}{78.26} \\
SD-v1.4 & 99.66 & 93.86 & 77.28 & 35.30 & \textcolor{myred}{99.98} & \textcolor{myblue}{99.58} & 99.22 & 85.40 & 74.10 & 21.54 & 91.43 & 57.33 & 70.66 & 20.74 & 97.48 & 43.58 & 73.28 & 24.50 & 95.42 & 61.52 \\
SD-v1.5 & 99.84 & 93.72 & 75.84 & 34.10 & \textcolor{myred}{99.98} & \textcolor{myblue}{99.40} & 99.36 & 86.36 & 74.22 & 21.70 & 92.53 & 57.43 & 69.90 & 20.56 & 97.68 & 44.24 & 70.94 & 23.36 & 95.00 & 63.38 \\
SSD-1B & 69.80 & 20.48 & 6.92 & 0.18 & 61.34 & 18.76 & 82.38 & 36.28 & 57.38 & 13.16 & 38.07 & 9.73 & 5.20 & 0.26 & 2.64 & 0.00 & 43.40 & 6.92 & \textcolor{myred}{85.62} & \textcolor{myblue}{38.44} \\
tiny-sd & 84.48 & 32.74 & 44.38 & 8.64 & 95.92 & \textcolor{myblue}{73.82} & 86.82 & 38.10 & 56.78 & 8.08 & 64.57 & 19.77 & 60.50 & 12.62 & \textcolor{myred}{96.54} & 25.04 & 61.62 & 13.80 & 88.10 & 40.84 \\
SegMoE-SD & 79.26 & 28.94 & 45.22 & 7.16 & 94.02 & \textcolor{myblue}{66.76} & 92.94 & 48.18 & 66.74 & 15.44 & 67.80 & 25.07 & 63.20 & 12.96 & \textcolor{myred}{99.36} & 43.44 & 36.94 & 4.86 & 94.08 & 53.18 \\
small-sd & 93.22 & 49.52 & 47.98 & 9.48 & \textcolor{myred}{98.88} & \textcolor{myblue}{83.98} & 86.78 & 36.70 & 61.84 & 11.26 & 73.77 & 27.57 & 67.40 & 13.44 & 97.52 & 25.62 & 68.86 & 19.04 & 89.74 & 40.16 \\
SD-2-1 & 75.64 & 28.70 & 10.08 & 0.92 & 79.40 & 33.52 & 38.88 & 6.50 & 69.80 & 18.60 & 64.40 & 23.20 & 17.34 & 1.80 & 21.78 & 0.90 & 42.32 & 7.24 & \textcolor{myred}{92.88} & \textcolor{myblue}{51.92} \\
SD-3-medium & 54.08 & 13.96 & 13.26 & 1.22 & 53.02 & 13.18 & 26.68 & 3.08 & 60.98 & 16.44 & 40.77 & 9.97 & 8.40 & 0.82 & 18.04 & 0.56 & 38.08 & 5.50 & \textcolor{myred}{87.10} & \textcolor{myblue}{35.62} \\
SDXL-turbo & 92.08 & 65.70 & 82.08 & 33.48 & 90.68 & 43.06 & 88.76 & 38.66 & 77.52 & 21.38 & 72.83 & 37.70 & 64.70 & 12.68 & 90.90 & 28.22 & 74.46 & 25.90 & \textcolor{myred}{98.54} & \textcolor{myblue}{80.52} \\
SD-2 & 69.54 & 21.38 & 9.10 & 0.72 & 70.12 & 24.66 & 29.46 & 4.16 & 53.78 & 10.90 & 53.97 & 13.77 & 12.08 & 1.14 & 21.58 & 0.84 & 39.56 & 5.72 & \textcolor{myred}{86.90} & \textcolor{myblue}{34.50} \\
SDXL & 61.38 & 13.00 & 2.90 & 0.08 & 46.88 & 7.18 & \textcolor{myred}{83.18} & \textcolor{myblue}{38.00} & 46.52 & 8.76 & 23.10 & 3.70 & 3.50 & 0.18 & 1.04 & 0.04 & 39.36 & 5.44 & 73.12 & 23.60 \\
PG-v2.5-1024 & 47.58 & 8.72 & 4.32 & 0.10 & 34.62 & 3.72 & 83.92 & 32.80 & 79.20 & 23.62 & 21.67 & 3.83 & 4.10 & 0.04 & 1.66 & 0.02 & 34.12 & 8.04 & \textcolor{myred}{92.46} & \textcolor{myblue}{45.76} \\
PG-v2-1024 & 71.64 & 20.80 & 4.94 & 0.10 & 62.52 & 12.48 & 25.62 & 2.24 & 80.50 & 26.56 & 41.00 & 9.13 & 4.02 & 0.12 & 7.74 & 0.04 & 19.14 & 1.42 & \textcolor{myred}{93.54} & \textcolor{myblue}{48.24} \\
PG-v2-512 & 52.40 & 10.54 & 11.22 & 0.76 & 43.76 & 9.54 & 9.34 & 0.72 & 39.92 & 5.74 & 30.67 & 5.83 & 3.78 & 0.34 & 13.28 & 0.32 & 49.22 & 8.56 & \textcolor{myred}{52.80} & \textcolor{myblue}{11.76} \\
PG-v2-256 & 60.88 & 17.80 & 13.36 & 1.88 & 40.56 & 7.22 & 27.42 & 4.60 & 50.76 & 7.34 & 49.23 & 14.97 & 6.80 & 0.66 & 13.30 & 0.90 & 33.90 & 4.68 & \textcolor{myred}{68.88} & \textcolor{myblue}{21.86} \\
PAXL-2-1024 & 56.92 & 17.40 & 9.46 & 0.36 & 62.08 & 16.38 & 26.80 & 3.44 & 91.60 & 43.48 & 32.50 & 7.50 & 12.10 & 0.18 & 21.02 & 0.78 & 30.92 & 3.84 & \textcolor{myred}{97.28} & \textcolor{myblue}{64.74} \\
PAXL-2-512 & 73.96 & 35.34 & 45.52 & 8.36 & 89.88 & 55.02 & 76.98 & 24.08 & 93.94 & 48.38 & 52.53 & 18.53 & 46.22 & 5.62 & 90.54 & 21.58 & 39.86 & 6.42 & \textcolor{myred}{98.34} & \textcolor{myblue}{75.92} \\
LCM-sdxl & 87.82 & 49.48 & 48.90 & 9.18 & 97.12 & 64.46 & 88.18 & 55.42 & 91.36 & 44.90 & 69.03 & 28.30 & 58.02 & 9.30 & 18.28 & 0.90 & 76.68 & 27.44 & \textcolor{myred}{99.32} & \textcolor{myblue}{77.02} \\
LCM-sdv1-5 & 94.16 & 74.98 & 84.30 & 46.26 & 98.40 & 78.82 & 94.56 & 68.20 & 93.22 & 46.48 & 82.37 & 48.57 & 78.12 & 31.86 & 98.10 & 67.46 & 61.98 & 11.88 & \textcolor{myred}{99.68} & \textcolor{myblue}{92.68} \\
FLUX.1-sch & 40.54 & 7.14 & 11.48 & 1.02 & 43.24 & 6.96 & 27.70 & 4.40 & 66.06 & 20.14 & 37.83 & 9.10 & 13.28 & 1.22 & 29.80 & 2.42 & 28.92 & 5.62 & \textcolor{myred}{89.50} & \textcolor{myblue}{39.12} \\
FLUX.1-dev & 46.86 & 12.36 & 10.30 & 1.24 & 54.26 & 14.22 & 24.92 & 5.24 & 76.98 & 30.20 & 39.23 & 9.57 & 5.06 & 0.34 & 18.12 & 1.28 & 35.50 & 7.12 & \textcolor{myred}{90.96} & \textcolor{myblue}{48.60} \\
\midrule
\textbf{Average} & 72.15 & 34.45 & 31.66 & 9.72 & 73.29 & 41.41 & 63.14 & 30.56 & 69.96 & 22.08 & 54.70 & 21.18 & 34.06 & 8.21 & 47.33 & 15.74 & 48.38 & 11.21 & \textcolor{myred}{89.44} & \textcolor{myblue}{51.26} \\
    \bottomrule
    \end{tabular}
\end{table}

\begin{table}[t]
    \centering
    \footnotesize
    \tabcolsep=6.pt
    \caption{Configuration of \ourtest{}.}
    \label{tab:data_custom}
    \begin{tabular}{lcrr}
    \toprule
    \textbf{Abbreviation} & \textbf{Model Name (on Huggingface)} & \textbf{Release Date} & \textbf{Image Count}\\
    \midrule
LDM & CompVis/ldm-text2im-large-256 & Jul. 2022 & 25,000 \\
SD-v1.4 & CompVis/stable-diffusion-v1-4 & Aug. 2022 & 25,000 \\
SD-v1.5 & runwayml/stable-diffusion-v1-5 & Oct. 2022 & 25,000 \\
SSD-1B & segmind/SSD-1B & Jul. 2023 & 25,000 \\
tiny-sd & segmind/tiny-sd & Jun. 2023 & 25,000 \\
SegMoE-SD & segmind/SegMoE-SD-4x2-v0 & \underline{Jan. 2024} & 25,000 \\
small-sd & segmind/small-sd & Jul. 2023 & 25,000 \\
SD-2-1 & stabilityai/stable-diffusion-2-1 & Dec. 2022 & 25,000 \\
SD-3-medium & stabilityai/stable-diffusion-3-medium-diffusers & \underline{Jun. 2024} & 25,000 \\
SDXL-turbo & stabilityai/sdxl-turbo & Nov. 2023 & 25,000 \\
SD-2 & stabilityai/stable-diffusion-2 & Nov. 2022 & 25,000 \\
SDXL & stabilityai/stable-diffusion-xl-base-1.0 & Jul. 2023 & 25,000 \\
PG-v2.5-1024 & playgroundai/playground-v2.5-1024px-aesthetic & \underline{Feb. 2024} & 25,000 \\
PG-v2-1024 & playgroundai/playground-v2-1024px-aesthetic & Dec. 2023 & 25,000 \\
PG-v2-512 & playgroundai/playground-v2-512px-base & Dec. 2023 & 25,000 \\
PG-v2-256 & playgroundai/playground-v2-256px-base & Dec. 2023 & 25,000 \\
PAXL-2-1024 & PixArt-alpha/PixArt-XL-2-1024-MS & Nov. 2023 & 25,000 \\
PAXL-2-512 & PixArt-alpha/PixArt-XL-2-512x512 & Nov. 2023 & 25,000 \\
LCM-sdxl & latent-consistency/lcm-lora-sdxl & Nov. 2023 & 25,000 \\
LCM-sdv1-5 & latent-consistency/lcm-lora-sdv1-5 & Nov. 2023 & 25,000 \\
FLUX.1-sch & black-forest-labs/FLUX.1-schnell & \underline{Aug. 2024} & 25,000 \\
FLUX.1-dev & black-forest-labs/FLUX.1-dev & \underline{Aug. 2024} & 25,000 \\
    \bottomrule
    \end{tabular}
\end{table}

\begin{table}[t]
    \centering
    \footnotesize
    \tabcolsep=7.pt
    \caption{Configuration of \ourtest{}/in-the-wild.}
    \label{tab:data_wild}
    \begin{tabular}{lccc}
    \toprule
    \textbf{Abbreviation} & \textbf{Source Website} & \textbf{Generative Model} & \textbf{Image Count} \\
    \midrule
Civitai & \href{https://civitai.com/}{https://civitai.com/} & Stable Diffusion~\cite{stablediffusion} & 10,000 \\
DALL-E 3 & \href{https://huggingface.co/datasets/ProGamerGov/synthetic-dataset-1m-dalle3-high-quality-captions}{huggingface} & DALL-E 3~\cite{dalle3} & Around 1M \\
instavibe.ai & \href{https://www.instavibe.ai/discover}{https://www.instavibe.ai/discover} & FLUX~\cite{flux} & 30,000 \\
Lexica & \href{https://lexica.art/}{https://lexica.art/} & Lexica Aperture~\cite{lexica} & 9,000 \\
Midjourney-v6 & \href{https://huggingface.co/datasets/terminusresearch/midjourney-v6-520k-raw}{huggingface} & Midjourney-v6~\cite{midjourney} & 520,000 \\
    \bottomrule
    \end{tabular}
\end{table}

\section{Detection Performance on More Evaluation Metrics} \label{app:more_metrics}
In addition to AP and accuracy, we consider F1 and ROC-AUC scores in \autoref{tab:eval_custom_f1_auc}, and TPRs at low FPRs in \autoref{tab:eval_custom_tpr_fpr}. The experiment is conducted on DRCT training set and \ourtest{} evaluation set. Observe that \ours{} consistently outperforms the existing baselines regarding the four new metrics, demonstrating its general high effectiveness.

\section{Details of \ourtest{} and \ourtest{}/in-the-wild} \label{app:detail_data}

\begin{figure}[t]
    \begin{minipage}[t]{0.65\linewidth}
        \centering
        \includegraphics[width=0.5\linewidth]{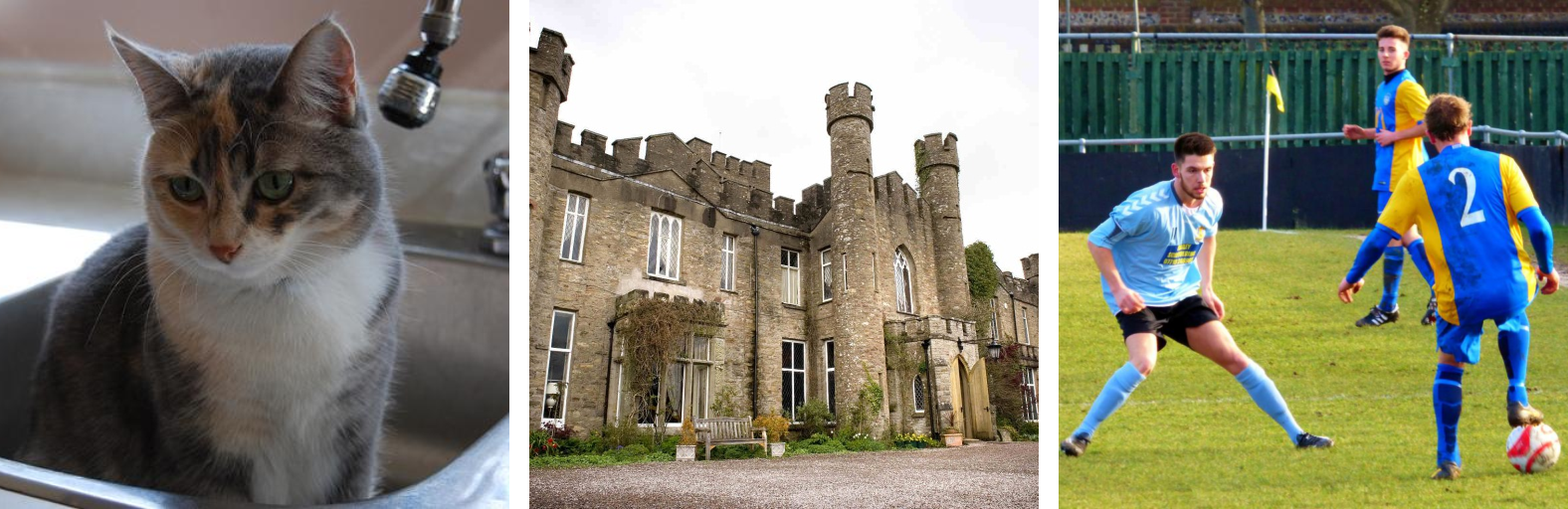}
        \subcaption{\underline{\textbf{Real images}}}
    \end{minipage}
    \hfill
    \begin{minipage}[t]{0.325\linewidth}
        \centering
        \includegraphics[width=1\linewidth]{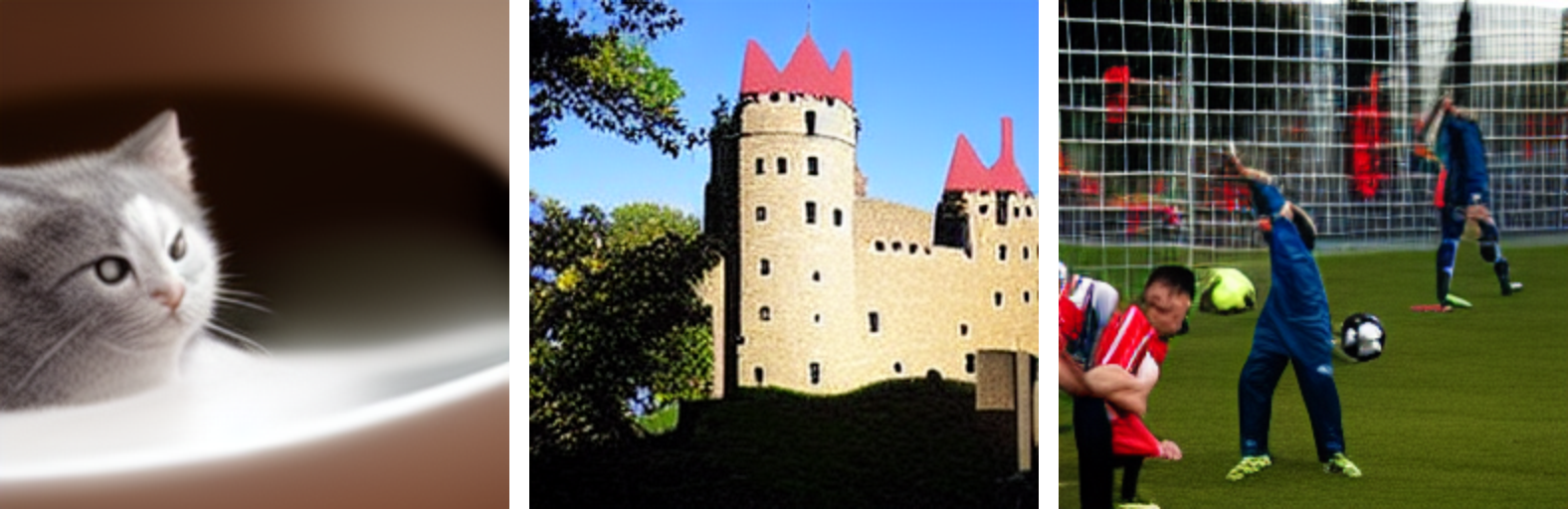}
        \subcaption{ldm-text2im-large-256}
    \end{minipage}
    \hfill
    \begin{minipage}[t]{0.325\linewidth}
        \centering
        \includegraphics[width=1\linewidth]{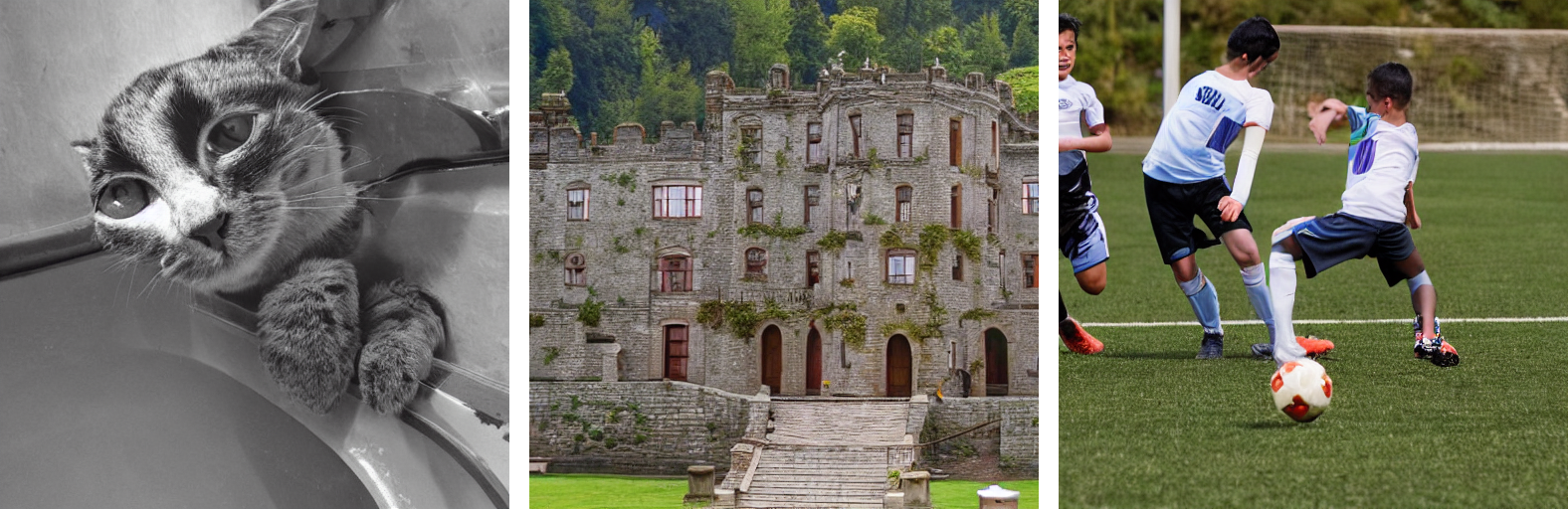}
        \subcaption{stable-diffusion-v1-4}
    \end{minipage}
    \hfill
    \begin{minipage}[t]{0.325\linewidth}
        \centering
        \includegraphics[width=1\linewidth]{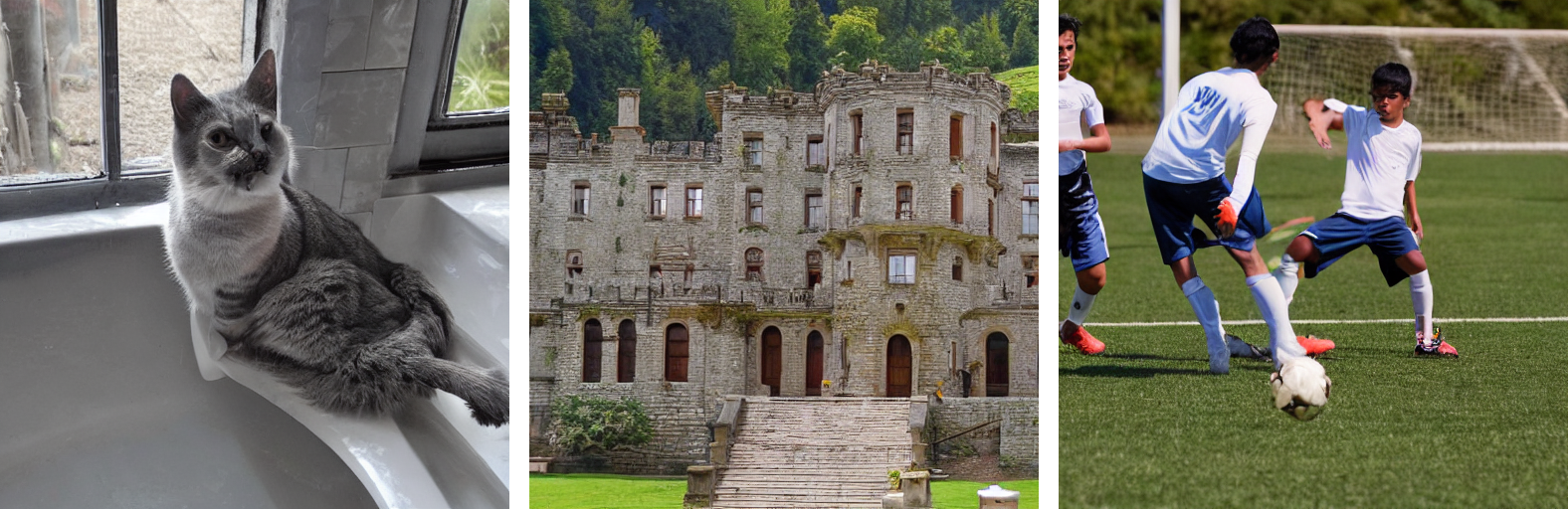}
        \subcaption{stable-diffusion-v1-5}
    \end{minipage}
    \hfill
    \begin{minipage}[t]{0.325\linewidth}
        \centering
        \includegraphics[width=1\linewidth]{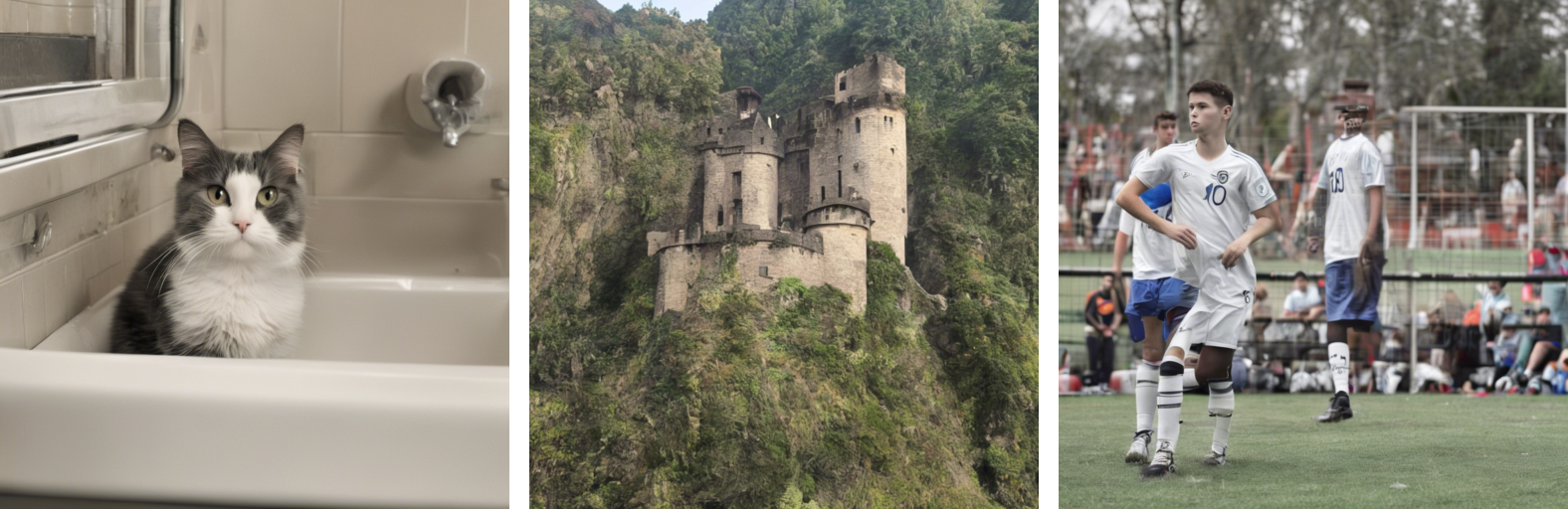}
        \subcaption{SSD-1B}
    \end{minipage}
    \hfill
    \begin{minipage}[t]{0.325\linewidth}
        \centering
        \includegraphics[width=1\linewidth]{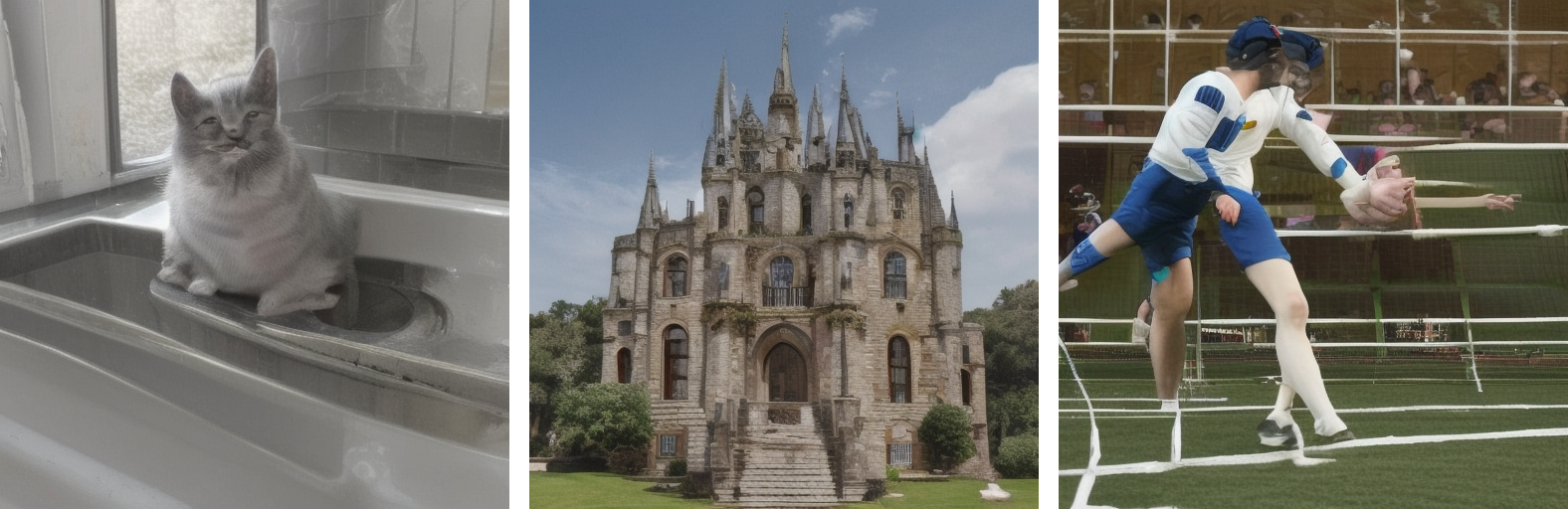}
        \subcaption{tiny-sd}
    \end{minipage}
    \hfill
    \begin{minipage}[t]{0.325\linewidth}
        \centering
        \includegraphics[width=1\linewidth]{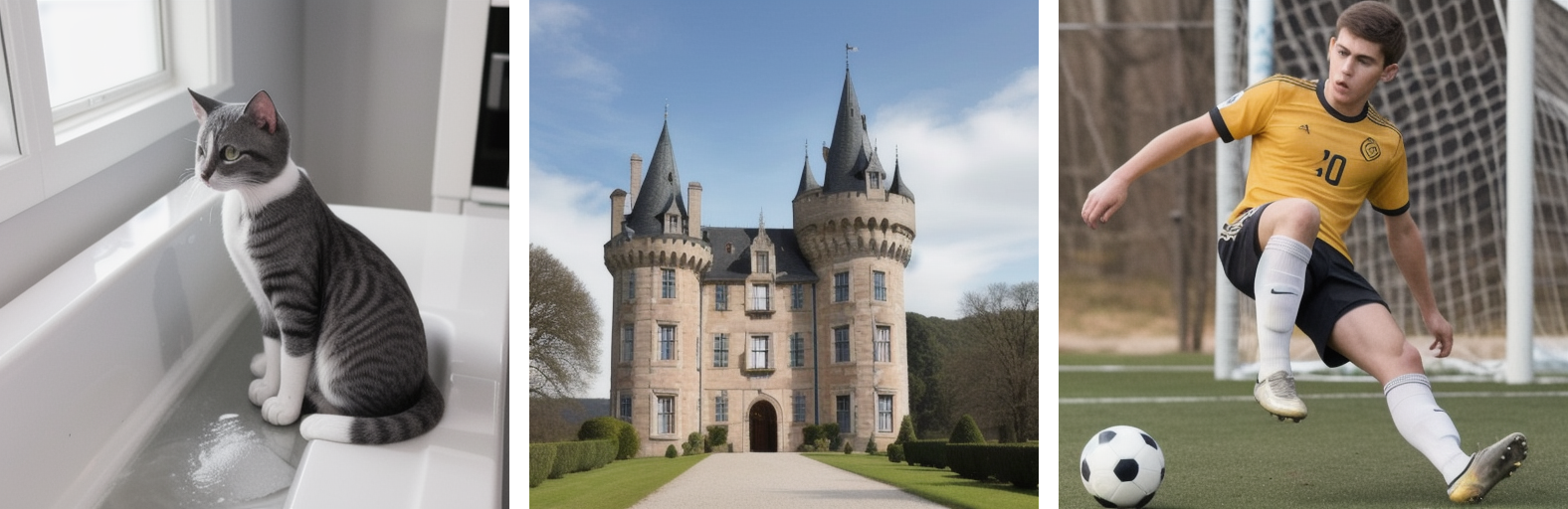}
        \subcaption{SegMoE-SD-4x2-v0}
    \end{minipage}
    \hfill
    \begin{minipage}[t]{0.325\linewidth}
        \centering
        \includegraphics[width=1\linewidth]{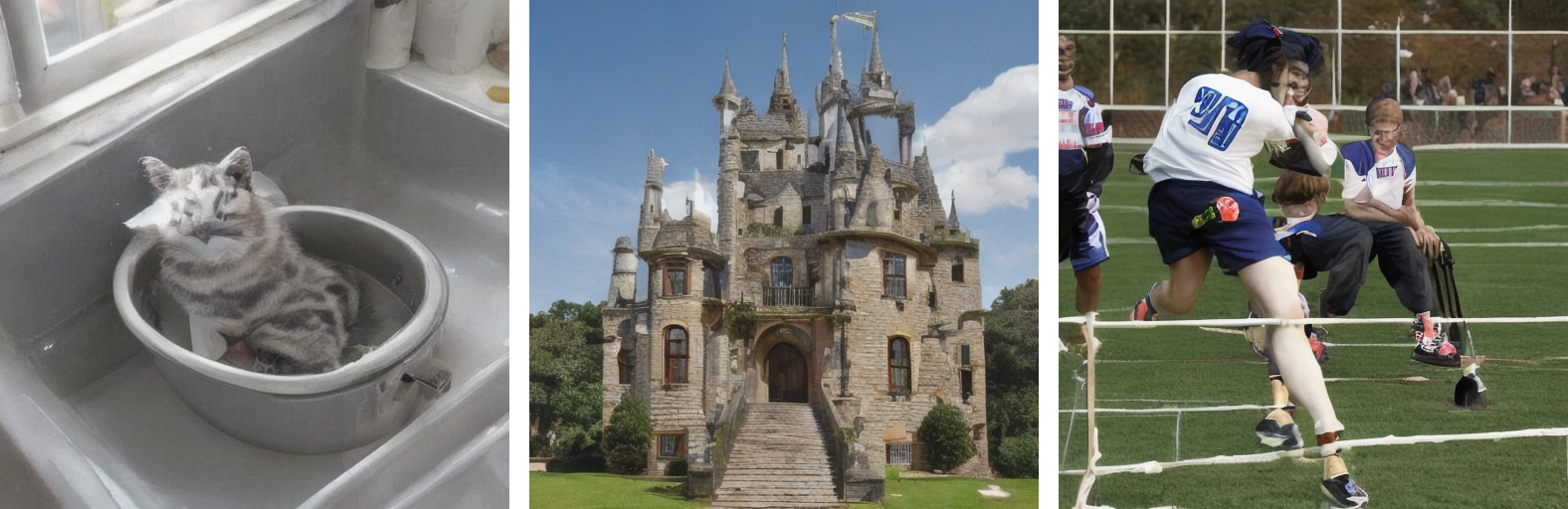}
        \subcaption{small-sd}
    \end{minipage}
    \hfill
    \begin{minipage}[t]{0.325\linewidth}
        \centering
        \includegraphics[width=1\linewidth]{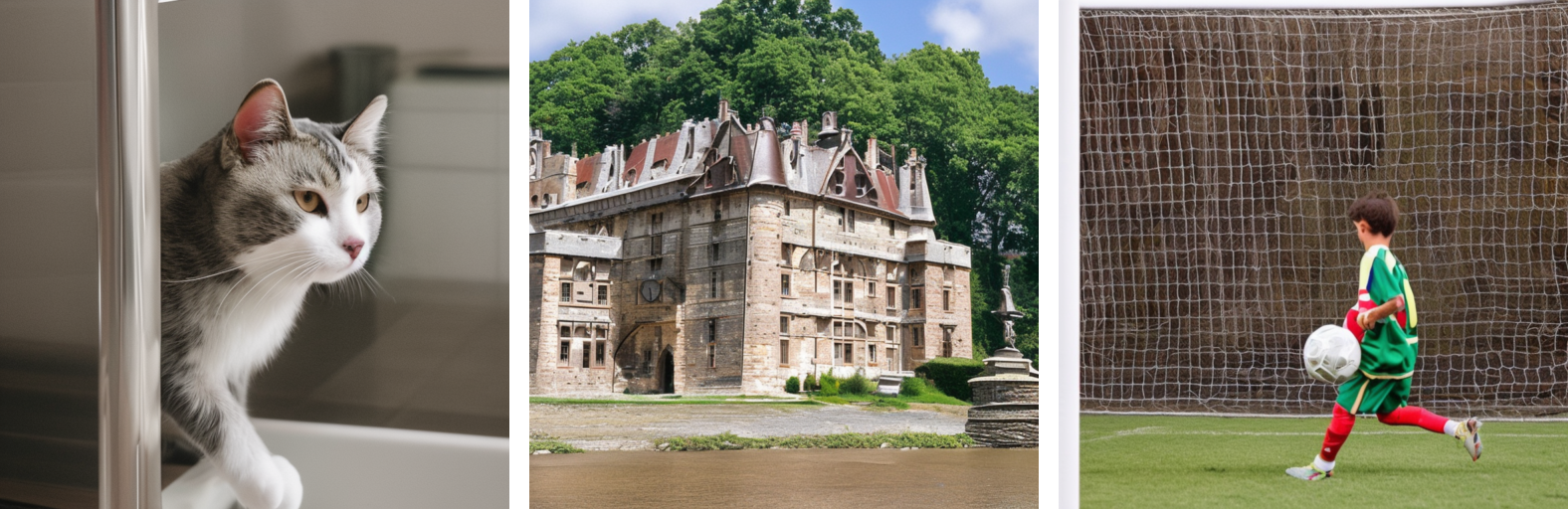}
        \subcaption{stable-diffusion-2-1}
    \end{minipage}
    \hfill
    \begin{minipage}[t]{0.325\linewidth}
        \centering
        \includegraphics[width=1\linewidth]{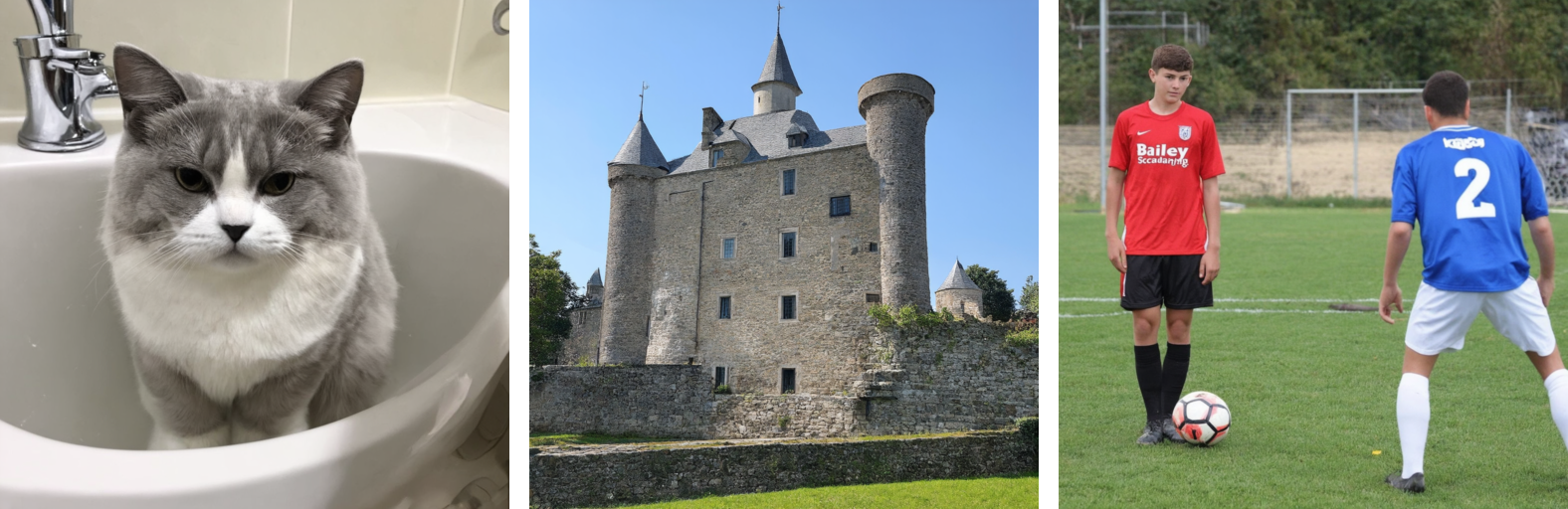}
        \subcaption{stable-diffusion-3-medium}
    \end{minipage}
    \hfill
    \begin{minipage}[t]{0.325\linewidth}
        \centering
        \includegraphics[width=1\linewidth]{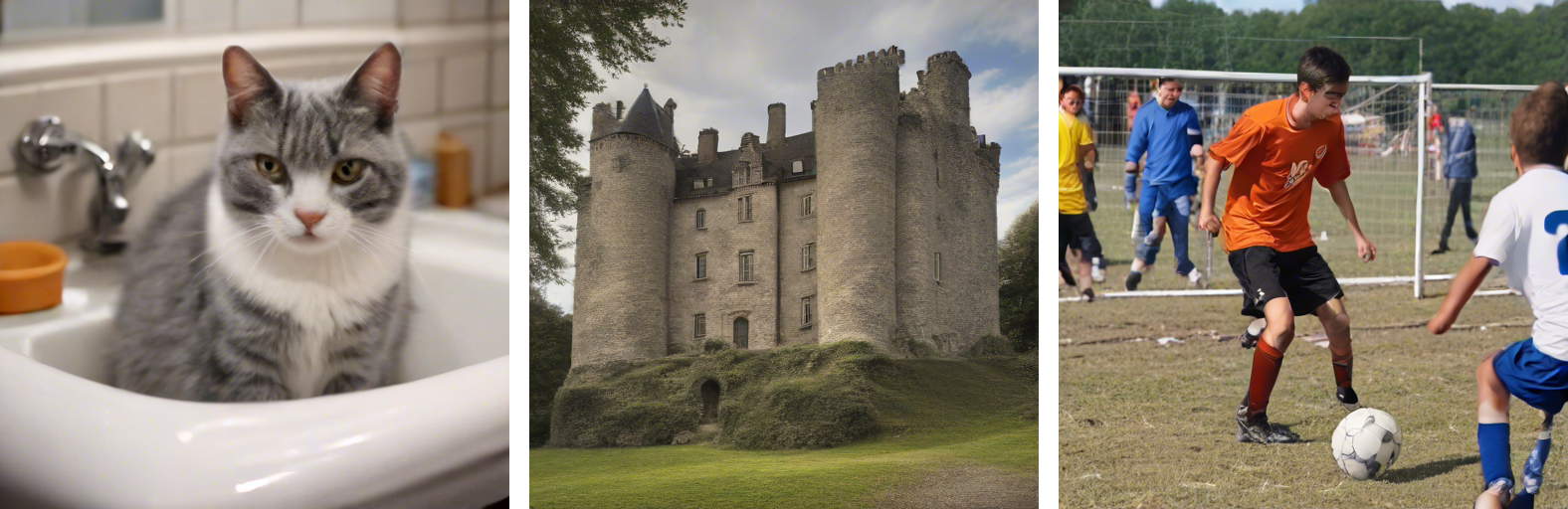}
        \subcaption{sdxl-turbo}
    \end{minipage}
    \hfill
    \begin{minipage}[t]{0.325\linewidth}
        \centering
        \includegraphics[width=1\linewidth]{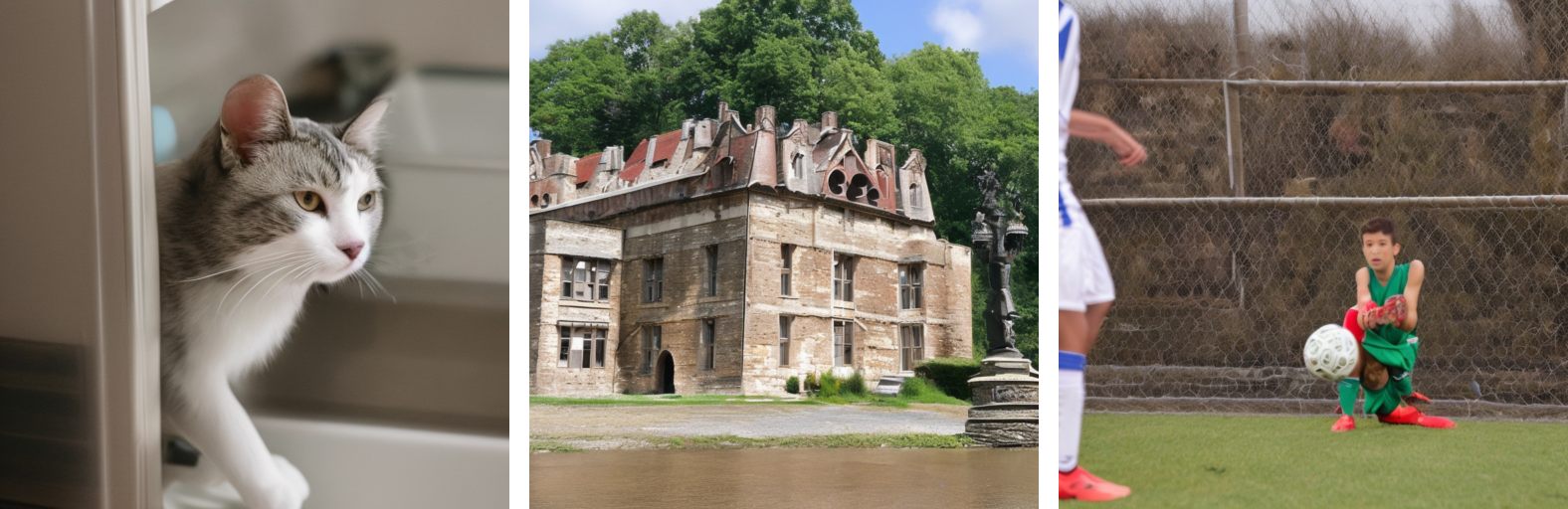}
        \subcaption{stable-diffusion-2}
    \end{minipage}
    \hfill
    \begin{minipage}[t]{0.325\linewidth}
        \centering
        \includegraphics[width=1\linewidth]{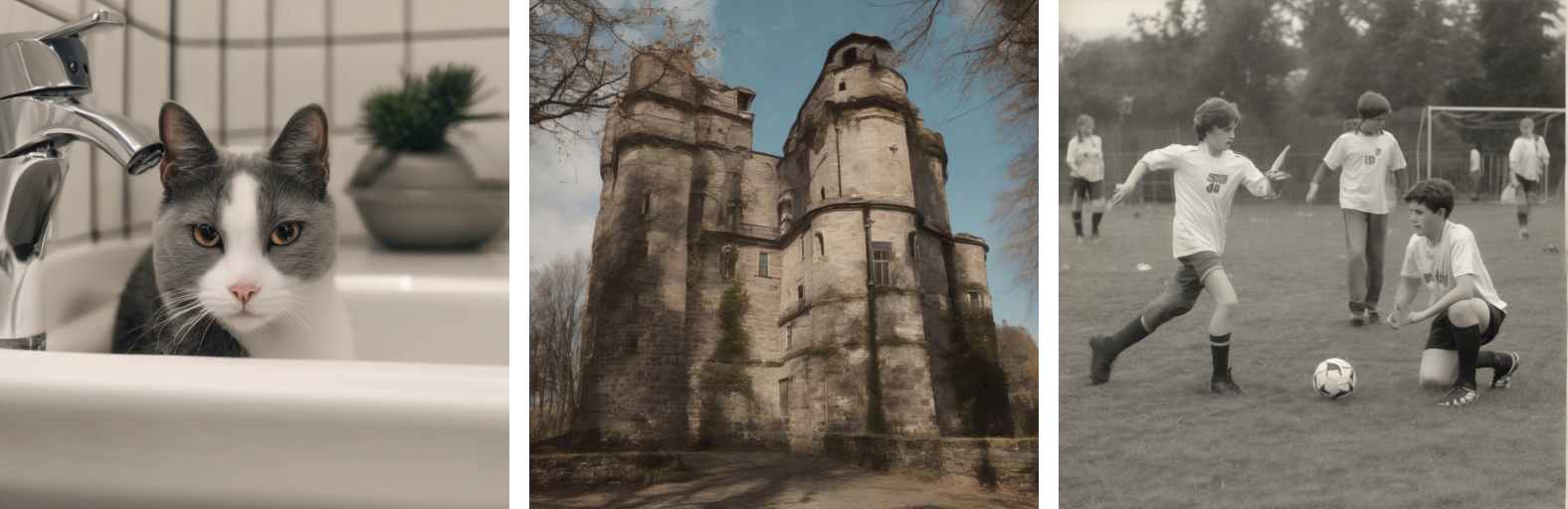}
        \subcaption{stable-diffusion-xl-base-1.0}
    \end{minipage}
    \hfill
    \begin{minipage}[t]{0.325\linewidth}
        \centering
        \includegraphics[width=1\linewidth]{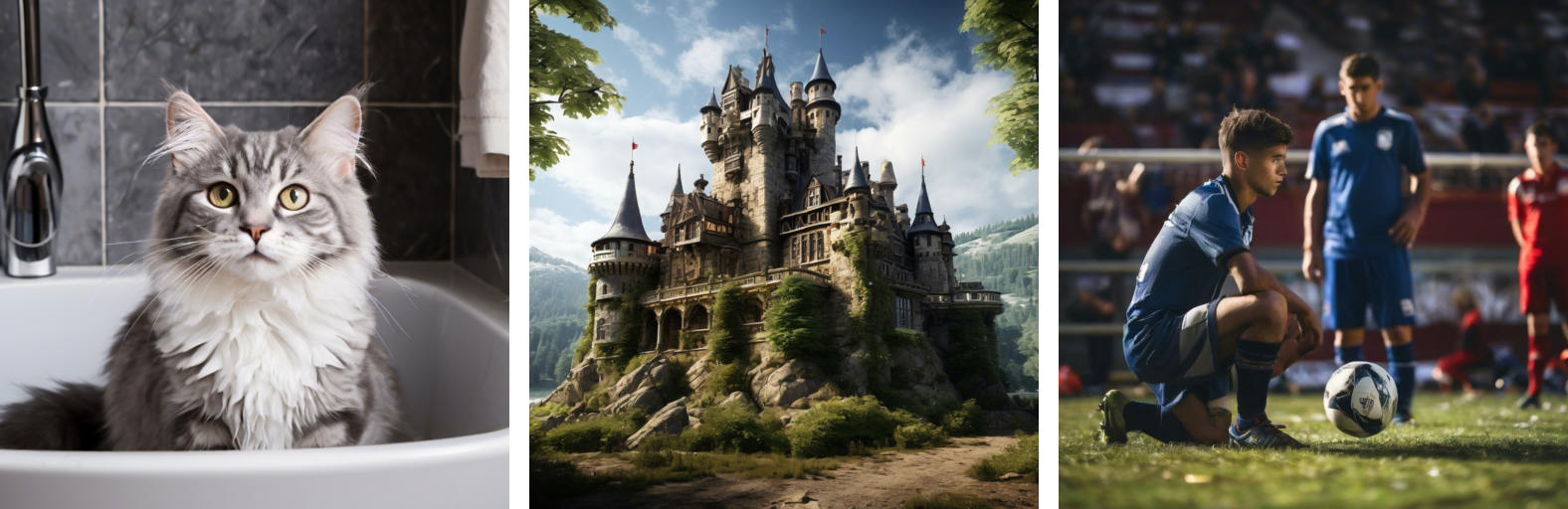}
        \subcaption{playground-v2.5-1024px-aesth.}
    \end{minipage}
    \hfill
    \begin{minipage}[t]{0.325\linewidth}
        \centering
        \includegraphics[width=1\linewidth]{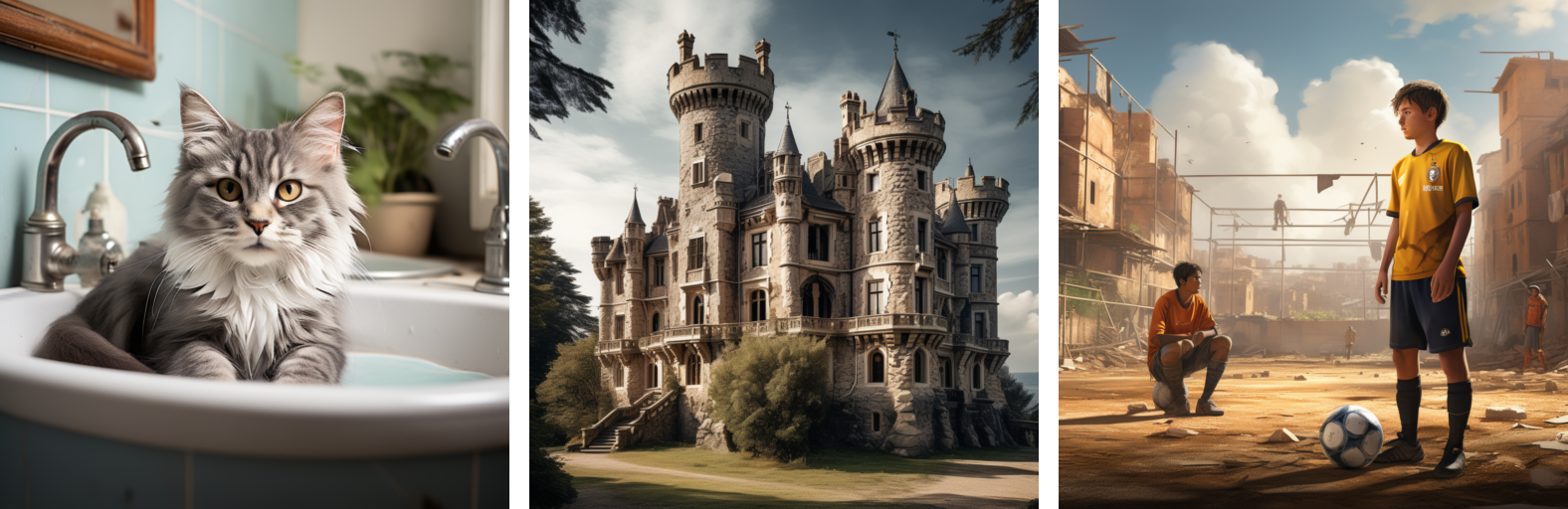}
        \subcaption{playground-v2-1024px-aesthetic}
    \end{minipage}
    \hfill
    \begin{minipage}[t]{0.325\linewidth}
        \centering
        \includegraphics[width=1\linewidth]{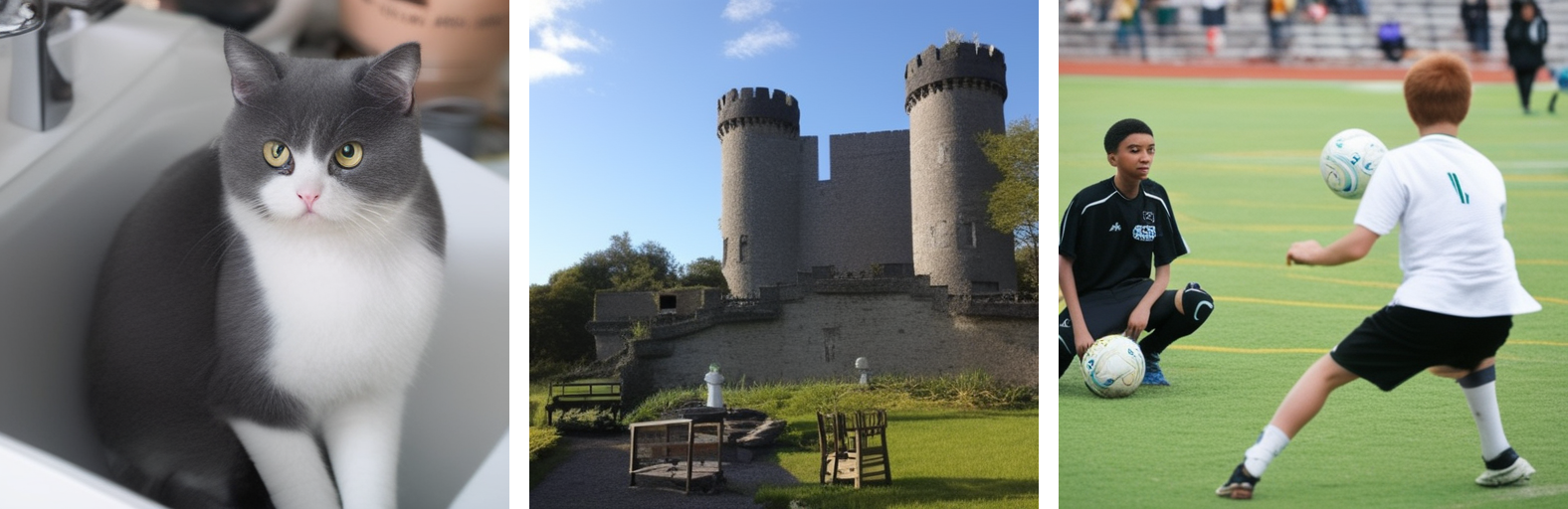}
        \subcaption{playground-v2-512px-base}
    \end{minipage}
    \hfill
    \begin{minipage}[t]{0.325\linewidth}
        \centering
        \includegraphics[width=1\linewidth]{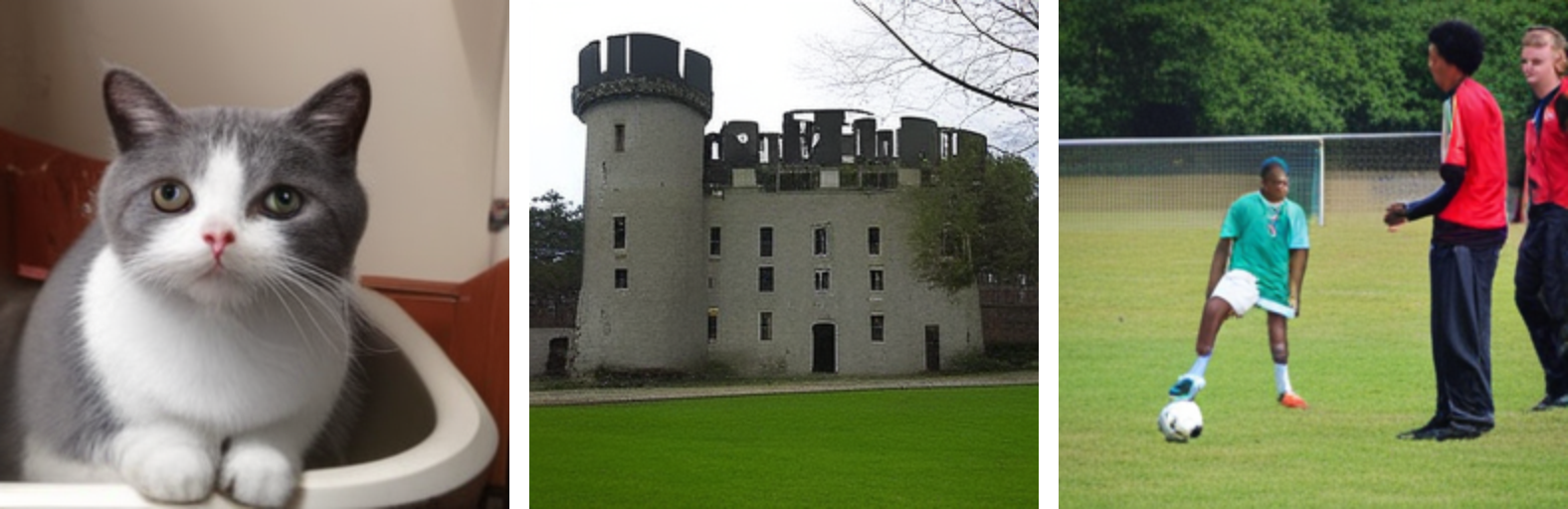}
        \subcaption{playground-v2-256px-base}
    \end{minipage}
    \hfill
    \begin{minipage}[t]{0.325\linewidth}
        \centering
        \includegraphics[width=1\linewidth]{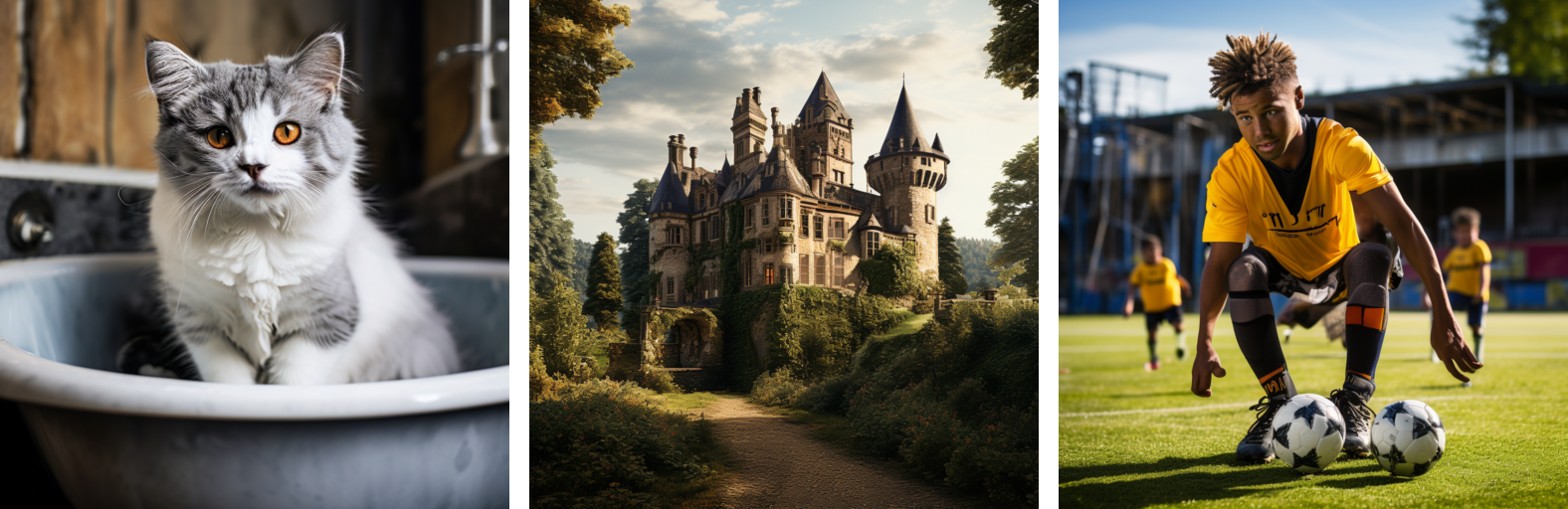}
        \subcaption{PixArt-XL-2-1024-MS}
    \end{minipage}
    \hfill
    \begin{minipage}[t]{0.325\linewidth}
        \centering
        \includegraphics[width=1\linewidth]{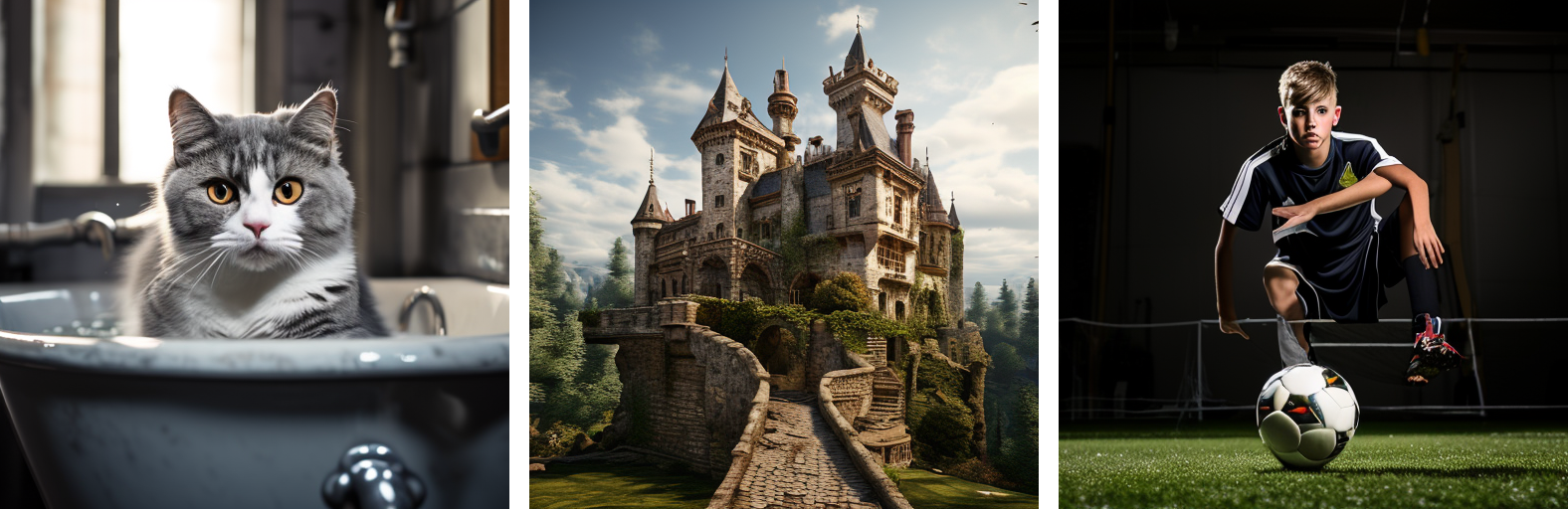}
        \subcaption{PixArt-XL-2-512x512}
    \end{minipage}
    \hfill
    \begin{minipage}[t]{0.325\linewidth}
        \centering
        \includegraphics[width=1\linewidth]{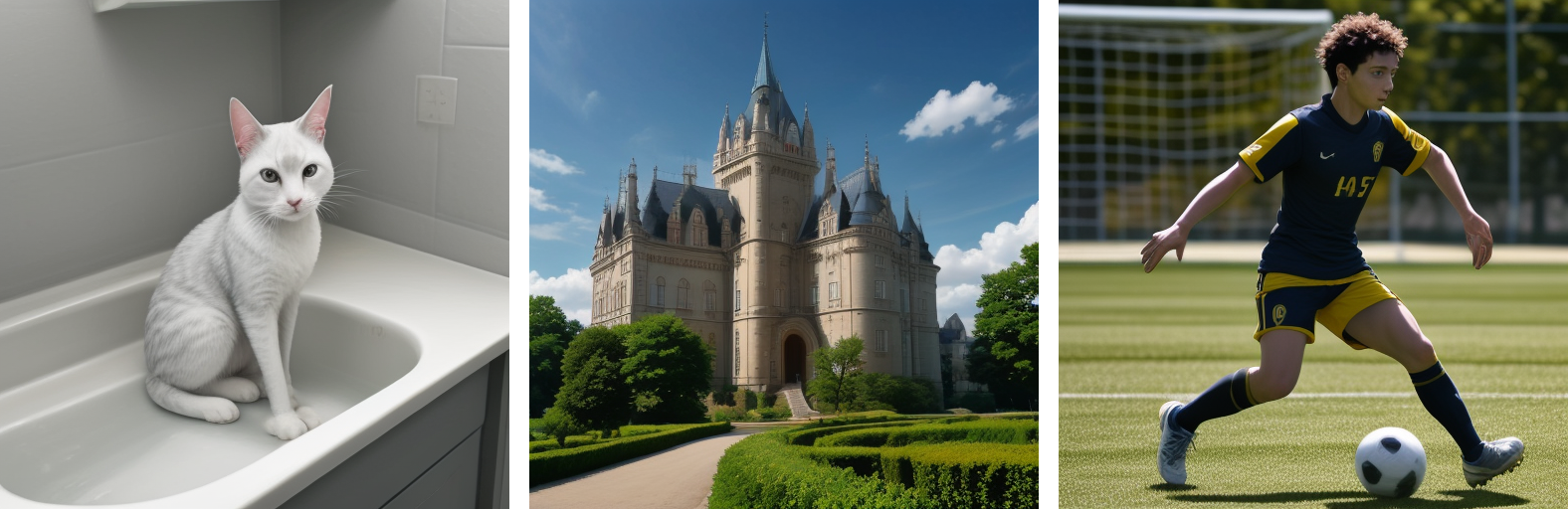}
        \subcaption{lcm-lora-sdv1-5}
    \end{minipage}
    \hfill
    \begin{minipage}[t]{0.325\linewidth}
        \centering
        \includegraphics[width=1\linewidth]{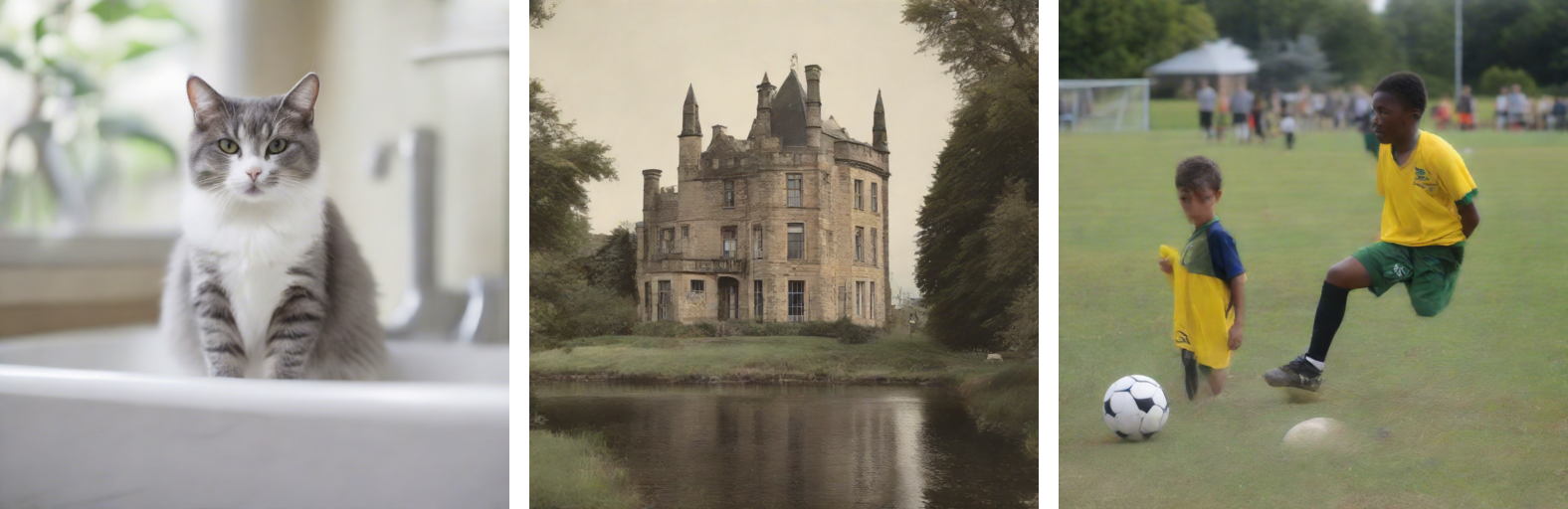}
        \subcaption{lcm-lora-sdxl}
    \end{minipage}
    \hfill
    \begin{minipage}[t]{0.325\linewidth}
        \centering
        \includegraphics[width=1\linewidth]{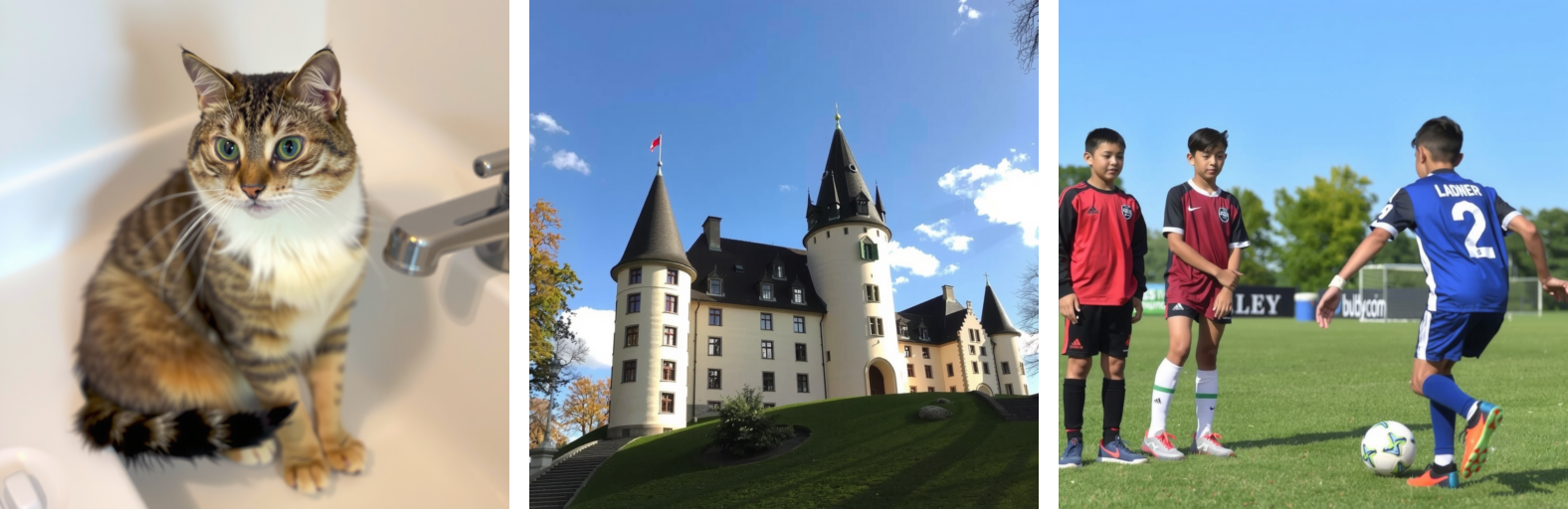}
        \subcaption{FLUX.1-schnell}
    \end{minipage}
    \hfill\begin{minipage}[t]{0.325\linewidth}
        \centering
        \includegraphics[width=1\linewidth]{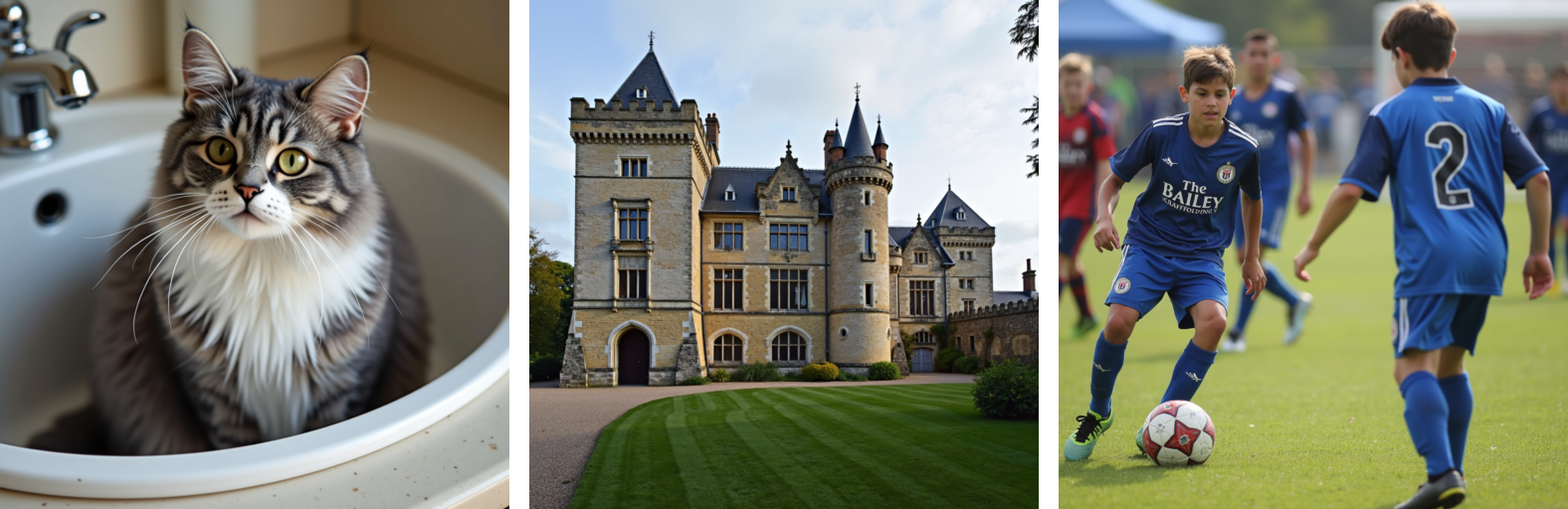}
        \subcaption{FLUX.1-dev}
    \end{minipage}
    \caption{\textbf{Demonstrations of \ourtest{} dataset.} Sub-figure (a) shows three real images from MSCOCO-2017, CC3M and TextCaps, respectively. The subsequent figures are generated using a series of state-of-the-art text-to-image generative models. The text captions used for generation are derived from the descriptions of the three real images: (Left) \href{http://images.cocodataset.org/train2017/000000410533.jpg}{``A grey and white cat sitting in a sink.''}; (Middle) \href{http://media.cntraveller.in/wp-content/uploads/2014/08/UK_castle.jpg}{This castle dates back to the 19th century.}; and (Right) \href{https://c5.staticflickr.com/8/7413/16590922592_41dcfda42c_o.jpg}{The young man on the soccer team sponsored by Bailey Scaffolding waits carefully as the number 2 player on the opposing team prepares to kick the ball.}. The url of each real image is provided using hyper-reference.}
    \label{fig:demo_custom}
\end{figure}

The first component of \ourtest{} focuses on generating synthetic images using state-of-the-art open-source models. We emphasize text-to-image generation due to its simplicity and widespread adoption. To ensure diversity within the dataset, we incorporate several key variations:
\begin{enumerate}
    \item \textbf{Different Generative Models}: We include 22 diffusion models, such as the latest FLUX~\cite{flux}, to cover a wide range of generative architectures.
    \item \textbf{Diverse Caption Inputs}: Captions are collected from five well-known image-text datasets, including MSCOCO~\cite{mscoco}, CC3M~\cite{cc3m}, Flickr~\cite{flickr}, TextCaps~\cite{textcaps}, and SBU~\cite{sbu}. Captions are randomly selected to generate a variety of image descriptions.
    \item \textbf{Varied Generation Configurations}: We randomize generation parameters by setting the number of inference steps between 10 and 50 and adjusting the guidance scales from 3.0 to 7.0. These settings are chosen to reflect typical and reasonable values used in image generation.
    \item \textbf{Broad Range of Real Images}: To maintain a balanced and comprehensive benchmark, we include an equal number of real images from the aforementioned caption datasets. This ensures that the benchmark provides a robust comparison between real and synthetic images.
\end{enumerate}
Details of the used models are provided in \autoref{tab:data_custom}. Additionally, \autoref{fig:demo_custom} presents illustrations of various synthetic images.

\begin{figure}[t]
    \begin{minipage}[t]{0.18\linewidth}
        \centering
        \includegraphics[width=1\linewidth]{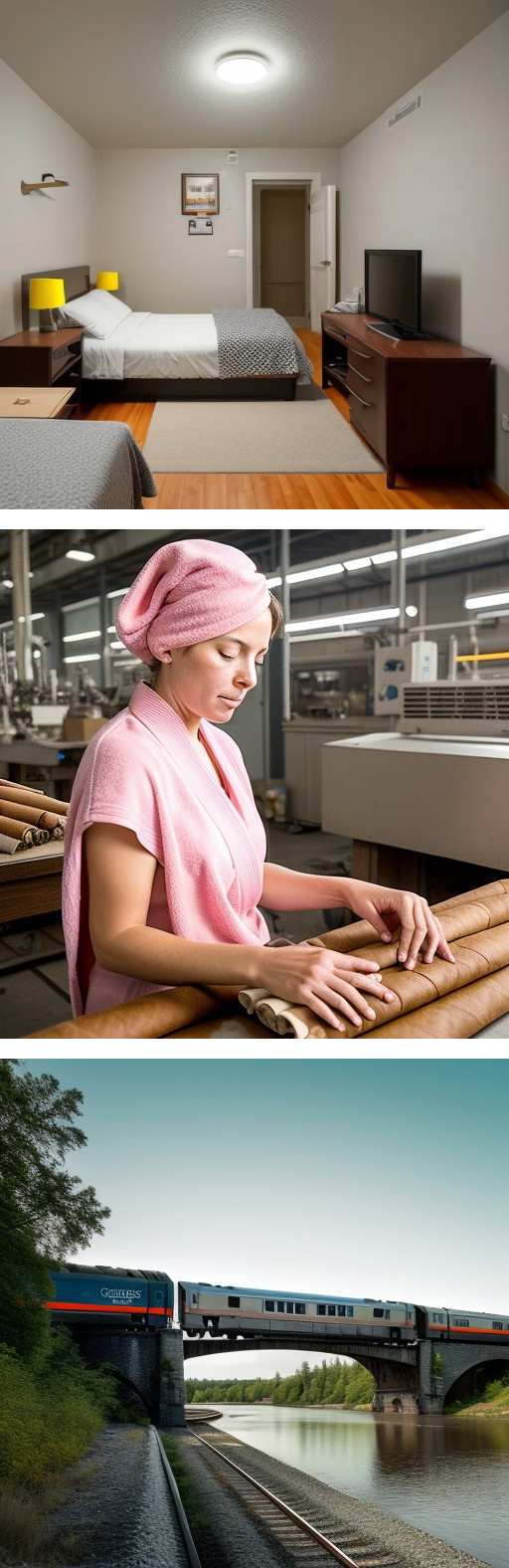}
        \subcaption{Civitai}
    \end{minipage}
    \hfill
    \begin{minipage}[t]{0.18\linewidth}
        \centering
        \includegraphics[width=1\linewidth]{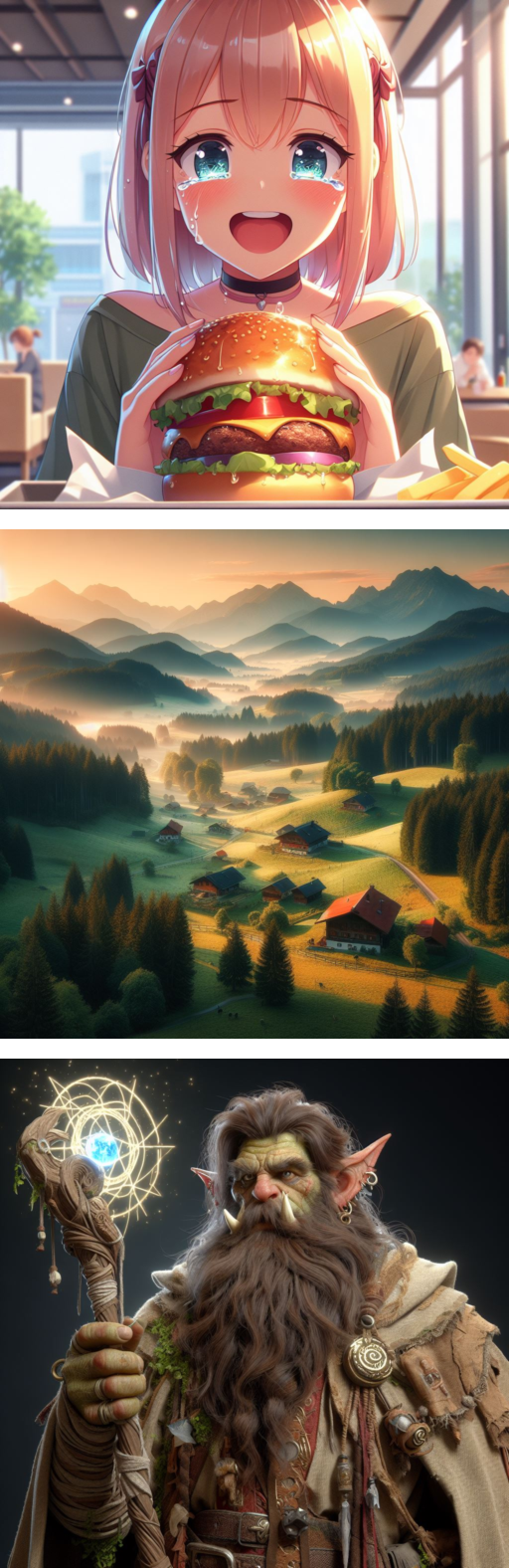}
        \subcaption{DALLE-3}
    \end{minipage}
    \hfill
    \begin{minipage}[t]{0.18\linewidth}
        \centering
        \includegraphics[width=1\linewidth]{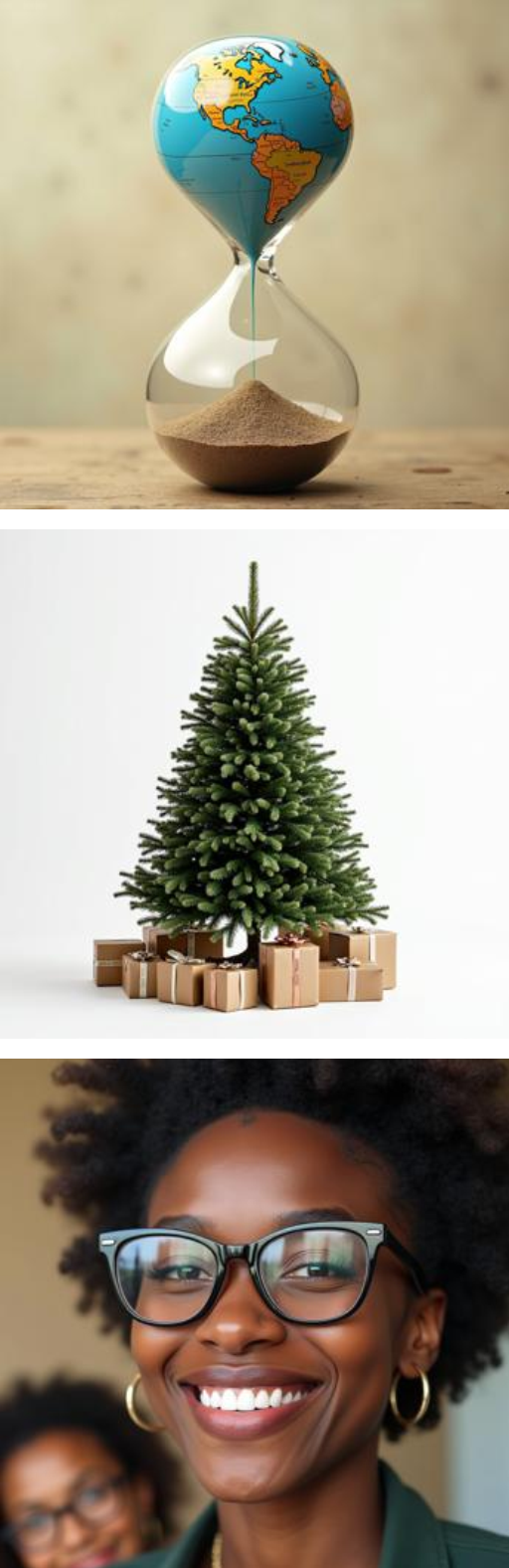}
        \subcaption{instavibe.ai}
    \end{minipage}
    \hfill
    \begin{minipage}[t]{0.18\linewidth}
        \centering
        \includegraphics[width=1\linewidth]{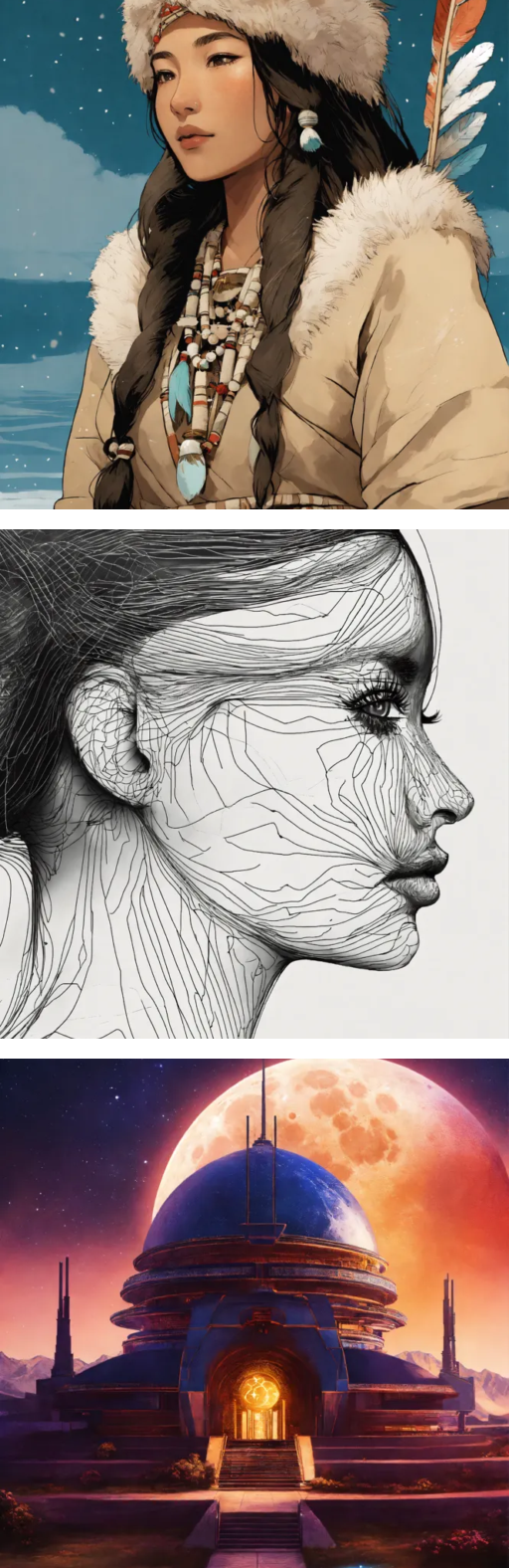}
        \subcaption{Lexica}
    \end{minipage}
    \hfill
    \begin{minipage}[t]{0.18\linewidth}
        \centering
        \includegraphics[width=1\linewidth]{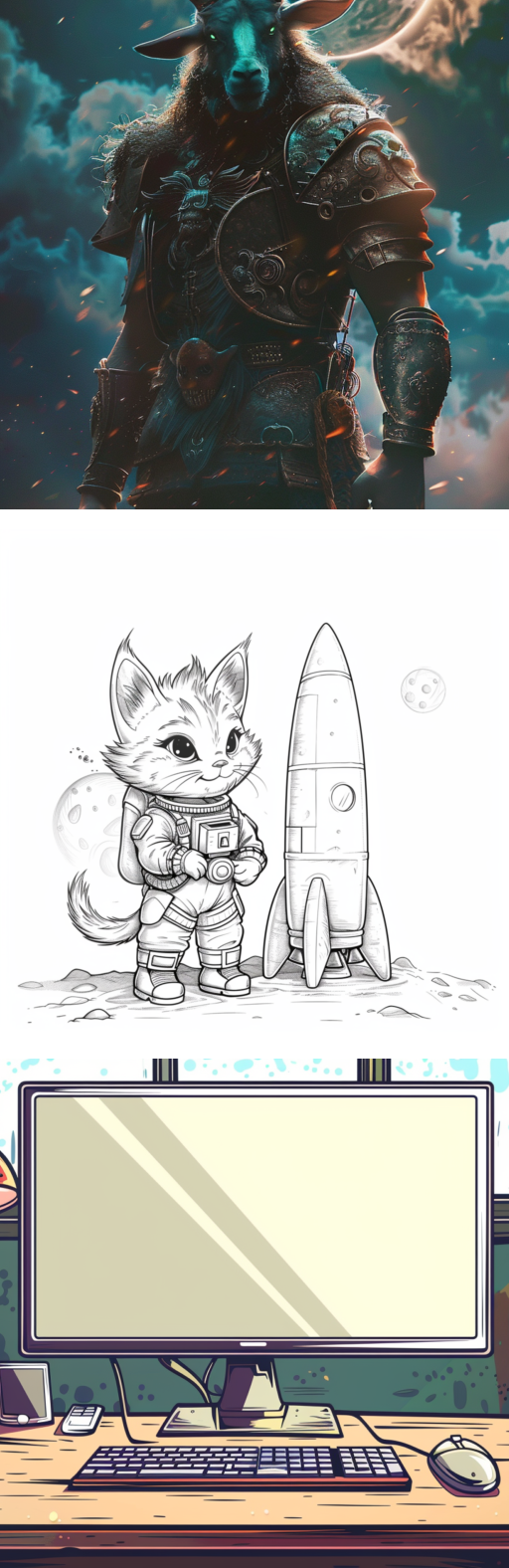}
        \subcaption{Midjourney-v6}
    \end{minipage}
    \caption{\textbf{Demonstration of \ourtest{}/in-the-wild dataset.}}
    \label{fig:demo_in_the_wild}
\end{figure}

\smallskip \noindent
\textbf{\ourtest{}/in-the-wild.}
The second component of \ourtest{} comprises synthetic images sourced from the Internet, providing a realistic evaluation environment for detection methods. Specifically, we collect images from five popular platforms, such as Civitai~\cite{civitai} and Lexica~\cite{lexica}. These images are generated and post-processed by developers or users, without our control over their creation processes. Consequently, this dataset closely mirrors practical, real-world scenarios, offering a robust evaluation of detection methods.
We provide descriptions of each source in \autoref{tab:data_wild} and illustrations in \autoref{fig:demo_in_the_wild}.

\end{document}